\documentclass{article}

\usepackage{microtype}
\usepackage{graphicx}
\usepackage{subcaption}
\usepackage{booktabs}

\usepackage{hyperref}

\usepackage[preprint]{styles/icml2026}

\usepackage[utf8]{inputenc}
\usepackage[T1]{fontenc}
\PassOptionsToPackage{hyphens}{url}  
\usepackage{url}
\usepackage{amsfonts}
\usepackage{amsmath}
\usepackage{amsthm}
\usepackage{nicefrac}
\usepackage{xcolor}
\usepackage{listings}
\usepackage{algorithm}
\usepackage{algorithmic}
\usepackage{enumitem}
\usepackage{multirow}
\usepackage{stfloats}      
\usepackage{placeins}       
\usepackage{tikz}
\usetikzlibrary{shapes,arrows,positioning,fit,backgrounds}
\usepackage{helvet}  
\usepackage{tcolorbox}  
\usepackage{pifont}  
\newcommand{\cmark}{\textcolor{green!60!black}{\ding{51}}}
\newcommand{\xmark}{\textcolor{red!70!black}{\ding{55}}}

\definecolor{abstractbg}{HTML}{EEF3F9}  
\newtcolorbox{headerpanel}{%
  colback=abstractbg,
  colframe=abstractbg,
  arc=8pt,
  boxrule=0pt,
  left=18pt, right=18pt, top=16pt, bottom=14pt,
  width=\textwidth,
  enlarge left by=0pt,
  enlarge right by=0pt,
}

\theoremstyle{definition}

\newcommand{\ara}{\textsc{Ara}}

\icmltitlerunning{The Last Human-Written Paper: Agent-Native Research Artifacts}

\begin{document}

\twocolumn[
  \thispagestyle{plain}
  \begin{headerpanel}
    \raggedright
    {\sffamily\bfseries\fontsize{19}{23}\selectfont
      The Last Human-Written Paper:\\
      Agent-Native Research Artifacts\par
    }
    \vskip 12pt
    {\sffamily\bfseries\normalsize
      Jiachen Liu\textsuperscript{1,\,*},\ %
      Jiaxin Pei\textsuperscript{2},\ %
      Jintao Huang\textsuperscript{3},\ %
      Chenglei Si\textsuperscript{2},\ %
      Ao Qu\textsuperscript{4},\ %
      Xiangru Tang\textsuperscript{5},\ %
      Runyu Lu\textsuperscript{1},\ %
      Lichang Chen\textsuperscript{6},\ %
      Xiaoyan Bai\textsuperscript{7},\ %
      Haizhong Zheng\textsuperscript{8},\ %
      Carl Chen\textsuperscript{9},\ %
      Zhiyang Chen\textsuperscript{10},\ %
      Haojie Ye\textsuperscript{11},\ %
      Yujuan Fu\textsuperscript{12},\ %
      Zexue He\textsuperscript{2},\ %
      Zijian Jin\textsuperscript{13},\ %
      Zhenyu Zhang\textsuperscript{2},\ %
      Shangquan Sun\textsuperscript{14},\ %
      Maestro Harmon\textsuperscript{15},\ %
      Dianzhuo Wang\textsuperscript{16},\ %
      Qian-ze Zhu\textsuperscript{16},\ %
      Jianqiao Zeng,\ %
      Jiachen Sun\textsuperscript{17},\ %
      Mingyuan Wu\textsuperscript{18},\ %
      Baoyu Zhou\textsuperscript{19},\ %
      Chenyu You\textsuperscript{20},\ %
      Shijian Lu\textsuperscript{14},\ %
      Yiming Qiu\textsuperscript{21},\ %
      Fan Lai\textsuperscript{18},\ %
      Yuan Yuan\textsuperscript{22},\ %
      Yao Li\textsuperscript{23},\ %
      Junyuan Hong\textsuperscript{24},\ %
      Ruihao Zhu\textsuperscript{25},\ %
      Beidi Chen\textsuperscript{8},\ %
      Alex Pentland\textsuperscript{2},\ %
      Ang Chen\textsuperscript{1},\ %
      Mosharaf Chowdhury\textsuperscript{1},\ %
      Zechen Zhang\textsuperscript{16,\,15}\par
    }
    \vskip 6pt
    {\small
      \textsuperscript{1}University of Michigan,\ %
      \textsuperscript{2}Stanford University,\ %
      \textsuperscript{3}Ohio State University,\ %
      \textsuperscript{4}MIT,\ %
      \textsuperscript{5}Yale University,\ %
      \textsuperscript{6}Meta Superintelligence Labs,\ %
      \textsuperscript{7}University of Chicago,\ %
      \textsuperscript{8}Carnegie Mellon University,\ %
      \textsuperscript{9}University of Washington,\ %
      \textsuperscript{10}University of Toronto,\ %
      \textsuperscript{11}NVIDIA,\ %
      \textsuperscript{12}Meta,\ %
      \textsuperscript{13}New York University,\ %
      \textsuperscript{14}Nanyang Technological University,\ %
      \textsuperscript{15}Orchestra Research,\ %
      \textsuperscript{16}Harvard University,\ %
      \textsuperscript{17}LinkedIn,\ %
      \textsuperscript{18}UIUC,\ %
      \textsuperscript{19}Arizona State University,\ %
      \textsuperscript{20}Stony Brook University,\ %
      \textsuperscript{21}University of Hong Kong,\ %
      \textsuperscript{22}Boston College,\ %
      \textsuperscript{23}Portland State University,\ %
      \textsuperscript{24}National University of Singapore,\ %
      \textsuperscript{25}Cornell University\par
    }
    \vskip 14pt
Scientific publication compresses a branching, iterative research process into a linear narrative, discarding the majority of what was discovered along the way. This compilation imposes two structural costs: a \textbf{Storytelling Tax}, where failed experiments, rejected hypotheses, and the branching exploration process are discarded to fit a linear narrative; and an \textbf{Engineering Tax}, where the gap between reviewer-sufficient prose and agent-sufficient specification leaves critical implementation details unwritten. Tolerable for human readers, these costs become critical when AI agents must understand, reproduce, and extend published work. We introduce the \textbf{Agent-Native Research Artifact (\ara{})}, a protocol that replaces the narrative paper with an agent-executable research package structured around four layers: scientific logic, executable code with full specifications, an exploration graph that preserves the failures compilation discards, and evidence grounding every claim in raw outputs. Three mechanisms support the ecosystem: a \textbf{Live Research Manager} that captures decisions and dead ends during ordinary development; an \textbf{\ara{} Compiler} that translates legacy PDFs and repos into \ara{}s; and an \textbf{\ara{}-native review system} that automates objective checks (analogous to a grammar checker for prose) so human reviewers can focus on significance, novelty, and taste. On PaperBench and RE-Bench, \ara{} raises question-answering accuracy from 72.4\% to 93.7\% and reproduction success from 57.4\% to 64.4\%. On RE-Bench's five open-ended extension tasks, preserved failure traces in \ara{} accelerate progress, but can also constrain a capable agent from stepping outside the prior-run box depending on the agent's capabilities.

    \vskip 14pt
    \noindent
    \parbox{0.70\linewidth}{%
      \noindent\textbf{Correspondence:}\ Jiachen Liu (\href{mailto:amberljc@umich.edu}{amberljc@umich.edu})\par
      \vskip 2pt
      \noindent\textbf{Code:}\ \href{https://github.com/AmberLJC/Agent-Native-Research-Artifact}{github.com/AmberLJC/Agent-Native-Research-Artifact}%
    }\hfill
    $\vcenter{\hbox{\sffamily\bfseries\small ARA Commons}}\,\vcenter{\hbox{\includegraphics[height=30pt]{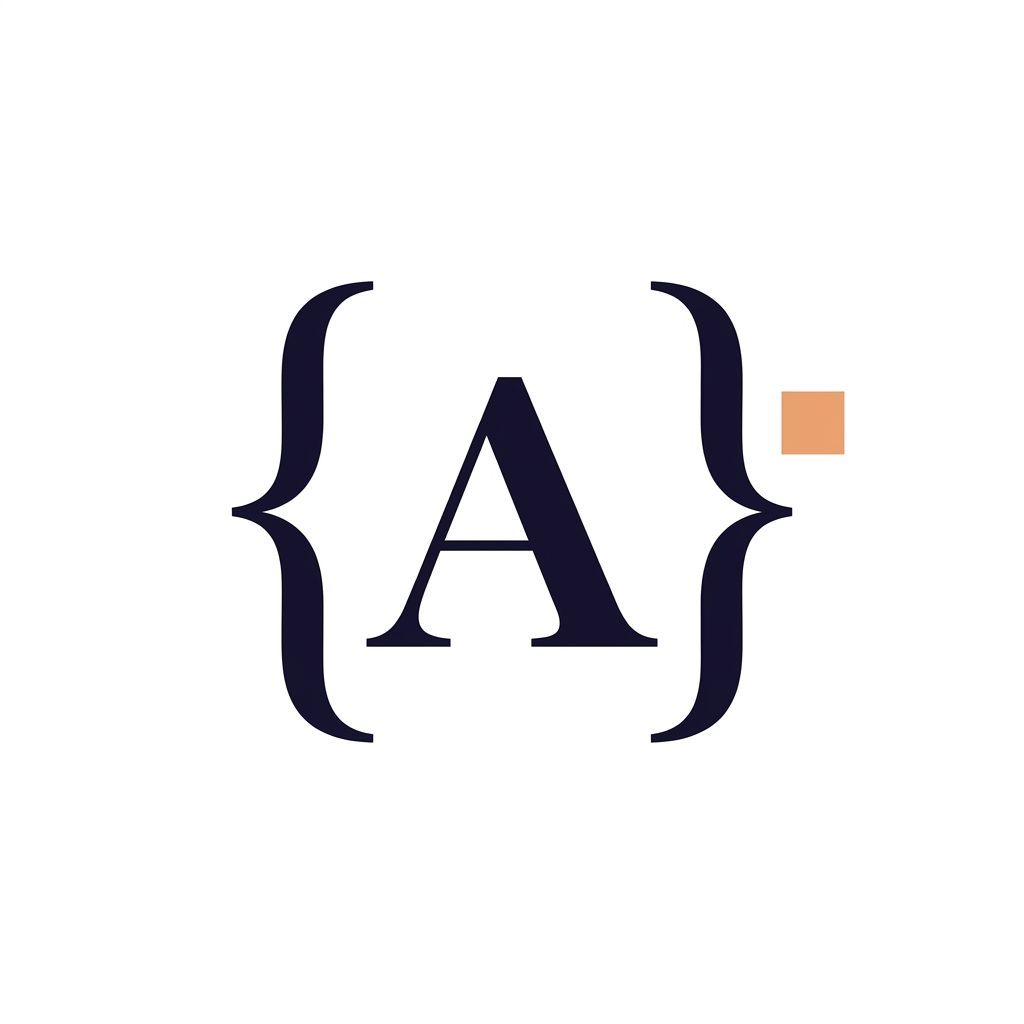}}}$
  \end{headerpanel}
]

\ifdefined\hypersetup
  \hypersetup{pdftitle={The Last Human-Written Paper: Agent-Native Research Artifacts}}
\fi

\makeatletter
\global\icml@noticeprintedtrue
\makeatother

\section{Introduction}
\label{sec:intro}

\begin{figure}[t]
    \centering
    \includegraphics[width=\linewidth]{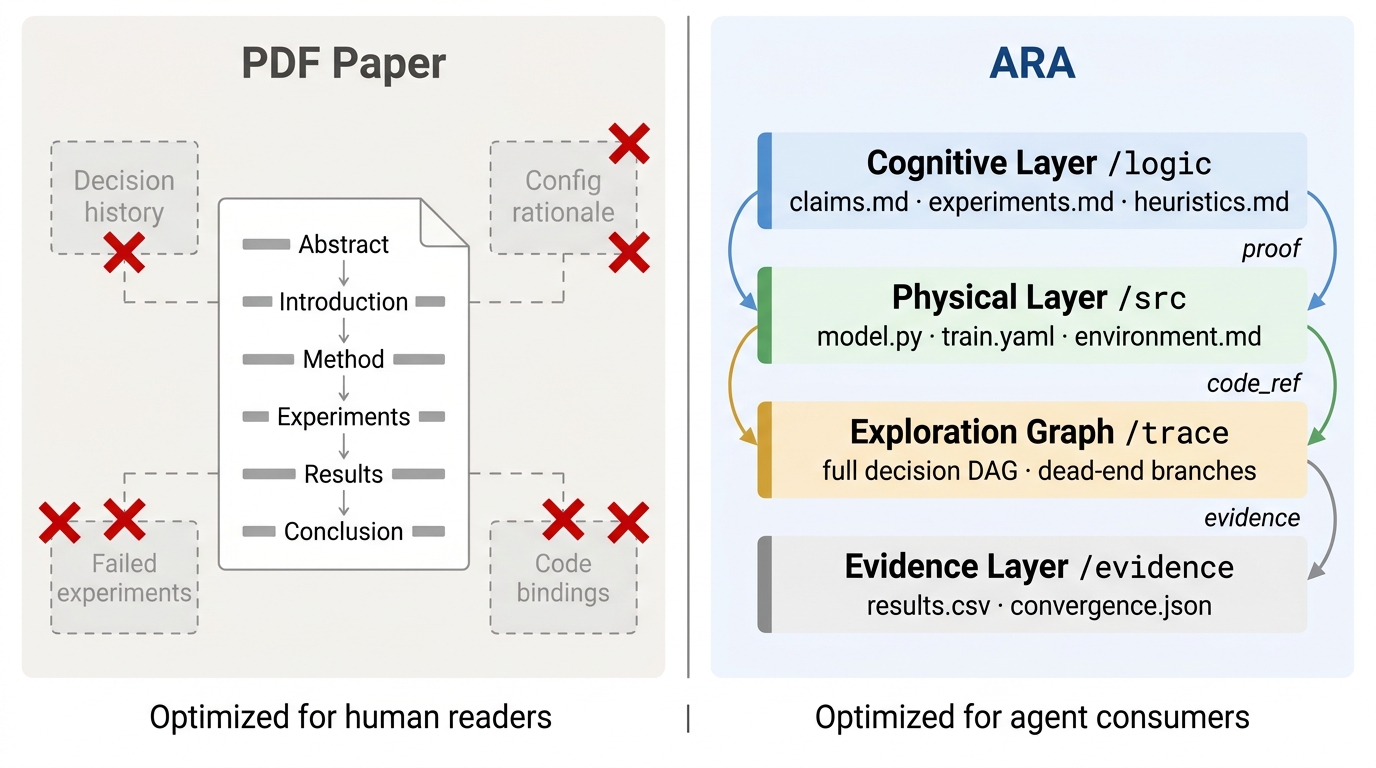}
    \caption{Publishing compiles a rich research object into a lossy narrative (left); \ara{} preserves the original as a high-fidelity, agent-executable knowledge package (right).}
    \label{fig:overview}
\end{figure}

Research produces a rich, branching knowledge object: months of hypotheses tested and rejected, implementation tricks discovered through trial and error, design alternatives weighed against each other, and the full exploration trajectory that explains why the final approach was chosen.
Publishing compiles this object into a linear narrative~\citep{medawar1963fraud, canini2026stopwriting}, discarding failed experiments, tacit engineering knowledge, and the branching process to satisfy the conventions of human-readable storytelling~\citep{rosenthal1979filedrawer, franco2014publication}.
This compilation cost, a consequence of the documentation convention rather than any particular file format, was tolerable when every consumer of a paper was human.
It is not when AI agents routinely read papers to understand a field, reproduce experiments to validate findings, and extend published methods to new settings~\citep{lu2024aiscientist, liu2026ainative}: each task requires precisely the knowledge that compilation discards (Figure~\ref{fig:overview}).
More specifically, the compilation incurs two structural costs.

\begin{figure}[t]
    \centering
    \includegraphics[width=0.95\linewidth]{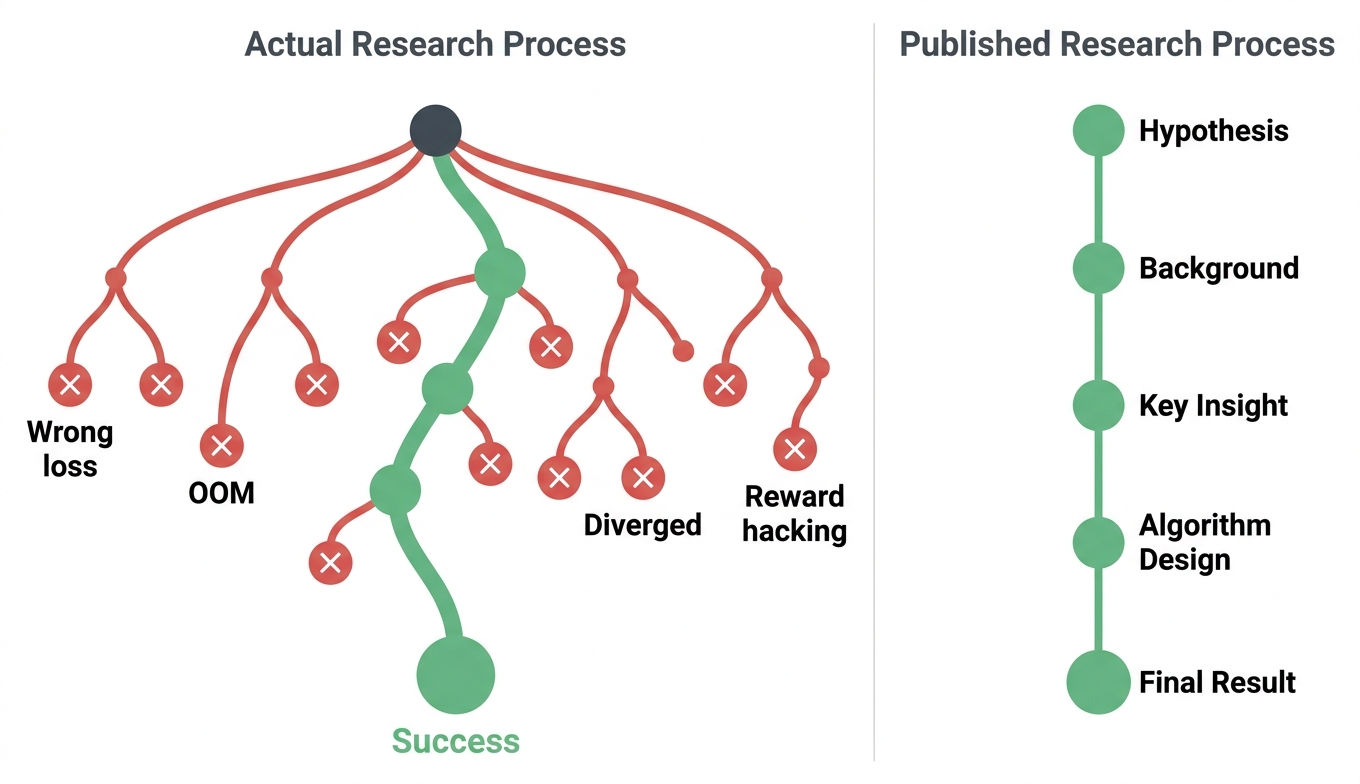}
    \caption{The Storytelling Tax: research proceeds as a branching tree with dead ends (left), but publication compiles it into a linear narrative (right), discarding all failure knowledge.}
    \label{fig:storytelling-tax}
\end{figure}

The first is the \textbf{Storytelling Tax}: the systematic erasure of research process knowledge imposed by compilation into narrative (Figure~\ref{fig:storytelling-tax}).
Research does not proceed linearly---it branches, backtracks, and accumulates hard-won failure knowledge before converging on a publishable result~\citep{kuhn1962structure,medawar1963fraud}.
Narrative compilation flattens this process into a polished linear story, discarding every failed experiment, rejected hypothesis, and abandoned approach.
This emphasis on success leaves failures undocumented; although modern platforms archive final artifacts, the branching research process remains unrecorded, causing independent rediscovery of the same dead ends across groups~\citep{rosenthal1979filedrawer, franco2014publication}.
Our own analysis of the METR eval-analysis-public dataset~\citep{wijk2025rebench}, covering 24{,}008 agent runs across 21 frontier models on RE-Bench, quantifies the cost (per-task breakdown in Appendix~\ref{app:exploration-tax}): failed runs account for \textbf{90.2\%} of total dollar cost (and 59.2\% of tokens), with a median failed-to-success token ratio of \textbf{113$\times$}, agents without access to prior failure records must independently rediscover every dead end.
Equally lost is the record of human \emph{judgment} along the trajectory: every rejection, revision, and endorsement is a preference signal over what constitutes good research, the scarce resource that binds once agents shoulder the grunt work. Narrative compilation discards this signal; a preserved trajectory renders it as structured supervision that compounds across projects.

\begin{figure}[t]
    \centering
    \includegraphics[width=\linewidth]{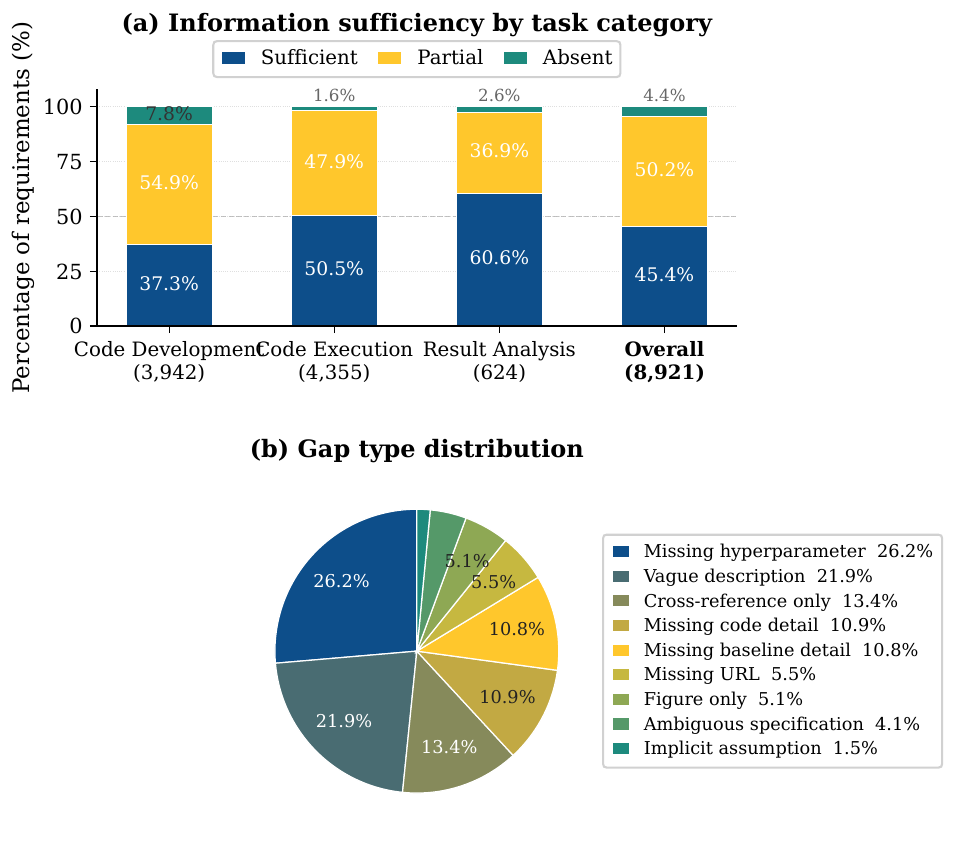}
    \caption{The reproduction information gap across 8,921 PaperBench requirements. \textbf{(a)}~PDFs systematically under-specify code development tasks. \textbf{(b)}~The three largest gap types are precisely the categories \ara{}'s structured layers address.}
    \label{fig:info-gap}
\end{figure}

The second is the \textbf{Engineering Tax}: the gap between \emph{reviewer-sufficient} and \emph{agent-sufficient} documentation (Figure~\ref{fig:info-gap}).
The paper communicates its contribution at the level of detail needed to convince a human reviewer; the codebase provides an implementation but not the operational specification needed to execute it.
Between the two lies tacit knowledge~\citep{polanyi1966tacit}---algorithmic tricks, implementation decisions, and configuration choices---knowledge that exists in no written document and is transmitted only through direct lab contact or painstaking reverse-engineering.
We quantify this void by classifying each of PaperBench's 8{,}921 expert-annotated reproduction requirements across 23 ICML~2024 papers~\citep{starace2025paperbench} against its source PDF (per-category breakdown and gap-type taxonomy in Appendix~\ref{app:gap-types}): despite widespread artifact sharing, only \textbf{45.4\%} are fully specified.
Code development is the most underspecified category (37.3\% sufficient), and missing hyperparameters alone account for 26.2\% of all gaps (full breakdown in Appendix~\ref{app:gap-types}): a fundamental mismatch between the precision at which papers are written (sufficient to produce belief) and the precision at which agents must operate (sufficient to produce correct execution)~\citep{stodden2016enhancing, baker2016reproducibility}.

Both taxes have persisted throughout the history of research because the human reader has always been the bandwidth-limited layer processing a vast, non-linear research trajectory.
Capable AI agents now offer a more efficient, human-proximal proxy that processes the trajectory at machine bandwidth on the reader's behalf, and three trends suggest what such an artifact should look like.
\textbf{First}, AI agents have become indispensable research companions, co-authoring code, running experiments, and iterating on hypotheses alongside humans~\citep{lu2024aiscientist}, with LLM adoption associated with paper-production increases of 23.7\%--89.3\% across scientific fields~\citep{kusumegi2025scientific}; the full research trajectory (every failure, implementation trick, configuration choice, design pivot) is now captured as machine-readable text in researcher-agent sessions, yet no protocol preserves it as a first-class output.
\textbf{Second}, humans and AI agents have divergent information needs: humans skim abstracts and figures~\citep{renear2009strategic}, while agents benefit from exhaustive detail that strictly improves reproduction, verification, and extension, so a single artifact optimized for human narrative can no longer serve both.
\textbf{Third}, research is scaling into a massively parallel enterprise where agents fork, extend, and merge each other's work at machine speed, shifting the bottleneck from individual productivity to artifact \emph{operability}: narrative PDFs, compiled for sequential human reading, cannot be forked, diffed, or merged, but a structured, lossless artifact can, letting research compound like software.

Existing efforts address fragments of this problem.
The FAIR principles~\citep{wilkinson2016fair} mandate findable, accessible data but say nothing about the structure of research \emph{arguments}.
RO-Crate~\citep{soilandreyes2022rocrate} packages research artifacts as archival bundles, not executable objects.
Nanopublications~\citep{groth2010nanopub} formalize atomic claims but lack the execution layer needed for reproduction.
The emerging \texttt{AGENTS.md} standard~\citep{openai2025agentsmd} provides agent-oriented documentation for code repositories but does not address the epistemic structure of research itself.
None of these efforts jointly structure scientific logic, executable code, and exploration history into a single operable object (\S\ref{sec:related}).

We propose the \textbf{Agent-Native Research Artifact (\ara{})}, a protocol that recasts the primary research object from narrative document to agent-executable knowledge package, with papers serving as compiled views of the underlying artifact (Figure~\ref{fig:overview}).
\ara{} organizes research into four interlocking layers: \emph{structured scientific logic} that distills the paper's conceptual abstractions into queryable claims and dependency graphs; \emph{executable code} with full operational specifications; an \emph{exploration graph} preserving the branching research process---failed experiments, rejected hypotheses, and design pivots---that narrative compilation discards; and \emph{grounded evidence} binding every claim to its raw empirical outputs.
Instead of parsing prose, reverse-engineering repositories, and rediscovering dead ends, an agent operating on an \ara{} artifact can query structured claims, execute declarative specifications, and build on the full decision history directly---a research object designed not to be read, but to be \emph{operated}.

To build the ecosystem around \ara{}, we develop three enabling mechanisms.
The \textbf{Live Research Manager} (\S\ref{sec:live-pm}) captures research decisions and dead ends as natural side-effects of everyday development, producing conforming artifacts without additional documentation burden.
The \textbf{\ara{} Compiler} (\S\ref{sec:ingestor}) translates legacy PDFs, repositories, and supplementary materials into \ara{} format, providing backward compatibility with the existing publication ecosystem.
The \textbf{\ara{}-Native Review System} (\S\ref{sec:review-system}) automates structural verification and budget-aware reproduction (analogous to a grammar checker for prose), redirecting expert attention from mechanical checking to judgment---significance, novelty, and taste~\citep{aczel2021billiondollar}.

\paragraph{Contributions.}
\begin{itemize}[leftmargin=*, itemsep=2pt]
    \item We identify two structural costs of compiling research into narrative---the \textbf{Storytelling Tax} and the \textbf{Engineering Tax}---and introduce the \textbf{Agent-Native Research Artifact (\ara{})}: a protocol that recasts the primary research object from narrative document to agent-executable knowledge package organized into four interlocking layers (\S\ref{sec:protocol}).
    \item We develop three enabling mechanisms: a \textbf{Live Research Manager} (\S\ref{sec:live-pm}) that captures research decisions during ordinary development; an \textbf{\ara{} Compiler} (\S\ref{sec:ingestor}) that translates legacy PDFs and repositories into \ara{} format; and an \textbf{\ara{}-Native Review System} (\S\ref{sec:review-system}) that automates objective verification (analogous to a grammar checker for prose) so human reviewers can focus on judgment.
    \item We evaluate \ara{} across three layers of research utility (\S\ref{sec:eval}): \emph{understanding} (what an agent can extract from the artifact), \emph{reproduction} (whether the agent can re-execute the paper's experiments), and \emph{extension} (whether the agent can build beyond the documented results and discover genuinely new findings, the defining goal of research). Across all three layers, agents operating on an \ara{} consistently outperform those reading the paper PDF and its associated code repository.
\end{itemize}

\FloatBarrier
\section{The \ara{} Protocol}
\label{sec:protocol}

The Agent-Native Research Artifact (\ara{}) protocol defines a file-system ontology that transforms CS research from a narrative document into a machine-executable knowledge package.
We describe the design philosophy (\S\ref{sec:philosophy}) and the layered architecture (\S\ref{sec:layers}).
Figure~\ref{fig:lrm-overview} (\S\ref{sec:live-pm}) illustrates how the Live Research Manager mediates between the human--AI research process and the living artifact across a long time horizon.

\begin{figure}[!t]
    \centering
    \includegraphics[width=\linewidth]{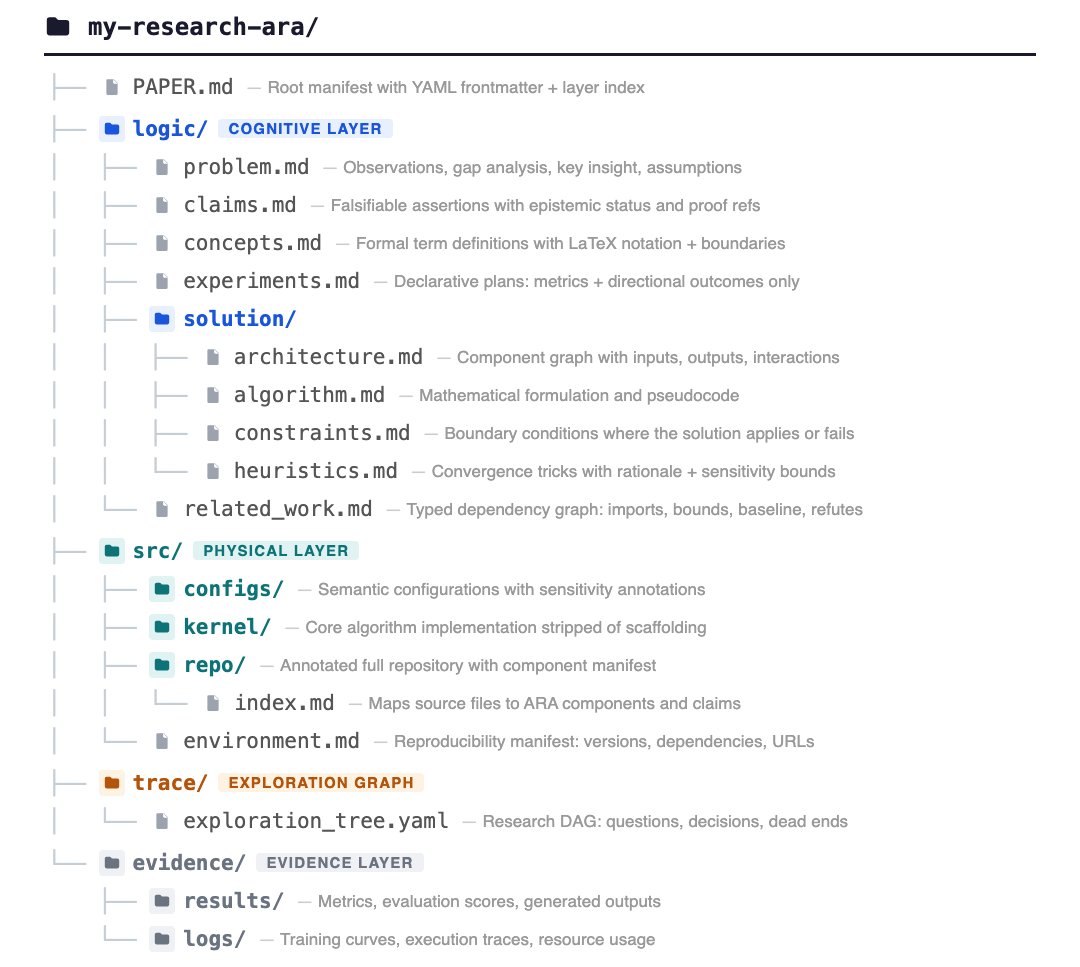}
    \caption{The \ara{} directory structure. Each file's role is annotated inline; layer badges mark the four top-level divisions. Cross-layer \emph{forensic bindings} link claims in \texttt{/logic} to evidence in \texttt{/evidence} and code in \texttt{/src}. Full schema in Appendix~\ref{app:schema}; design rationale in Appendix~\ref{app:protocol-details}.}
    \label{fig:ara-structure}
\end{figure}

\subsection{Design Philosophy}
\label{sec:philosophy}

The \ara{} protocol is grounded in a single principle: \textbf{Knowledge over Narrative}---the organized, evolving knowledge produced during research is the primary scientific object; the narrative paper is a compiled view.

\paragraph{A structured knowledge vessel.}
An agent engaging with a research project asks four structurally distinct questions: \emph{why} does this work (scientific reasoning), \emph{how} is it implemented (executable code), \emph{what was tried} along the way (exploration trajectory), and \emph{what are the numbers} (raw empirical evidence).
A narrative paper forces the agent to extract all four answers from the same linear prose, yet these knowledge types conflict in structure: reasoning demands stable, citable units while code iterates continuously; the exploration history is inherently branching while narrative enforces linearity; evidence requires machine-precise values while prose rounds and paraphrases.
Compressing them into a single document is not merely suboptimal but \emph{lossy}: once flattened into narrative, the original structure cannot be recovered.
\ara{} eliminates this loss by materializing each knowledge type as its own layer within an agent-native file-system structure (Figure~\ref{fig:ara-structure}): plain-text, independently queryable directories that agents navigate, read, and act on with standard tool calls, without parsing prose or reverse-engineering repositories.
Because agent context windows are a shared, finite resource, the structure further supports \emph{progressive disclosure}: agents load only the layers and files relevant to their current task, avoiding unnecessary context pollution.

The four layers below are the structural response to the Storytelling and Engineering Taxes introduced in \S\ref{sec:intro}: the Exploration Graph recovers the branching trajectory collapsed by narrative; the Cognitive and Physical Layers, bound by forensic links, close the gap between conceptual description and executable specification.
Within each layer, artifact text maximizes information per token: subjective qualifiers, hedges, and narrative connectives are stripped, and statements requiring judgment carry provenance rather than rhetoric.

\begin{figure}[!t]
    \centering
    \includegraphics[width=\linewidth]{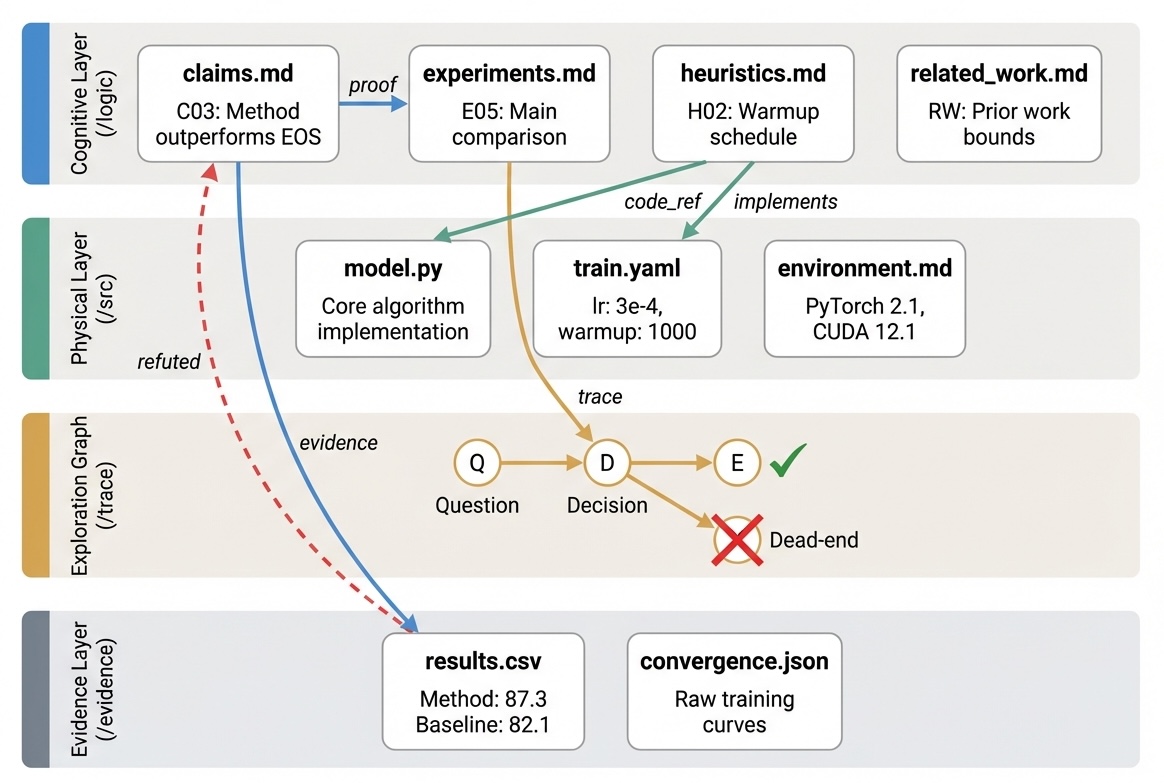}
    \caption{Cross-layer structure of a real \ara{}. Claims in \texttt{/logic} link to code in \texttt{/src} and evidence in \texttt{/evidence} via forensic bindings. The Exploration Graph (bottom center) captures the research DAG with dead-end nodes (marked $\times$) that preserve failure modes and lessons.}
    \label{fig:exploration-graph}
    \end{figure}
    
\subsection{\ara{} Architecture}
\label{sec:layers}

An agent engaging with a research artifact has four fundamental needs: \emph{understand} the contribution, \emph{reproduce} it, \emph{learn from the process}, and \emph{verify} claims against raw evidence.
\ara{} materializes each need as a dedicated layer (Figure~\ref{fig:ara-structure}), rooted in a manifest \texttt{PAPER.md} whose YAML frontmatter and layer index enable an agent to triage relevance in ${\sim}500$ tokens (Appendix~\ref{app:schema}).
The decomposition is informed by a taxonomy of ten reproduction-critical information categories derived from PaperBench rubrics (Appendix~\ref{app:taxonomy}).

\paragraph{The Cognitive Layer (\texttt{/logic}).}
An agent reads \texttt{/logic} \emph{first} to understand \emph{what was done and why}: \texttt{problem.md} defines the gap and key insight, \texttt{solution/} specifies the architecture, algorithm, and convergence-critical heuristics, \texttt{claims.md} distills falsifiable assertions with explicit proof pointers, and \texttt{experiments.md} declares the verification plan.
\texttt{related\_work.md} replaces passive citations with \emph{typed dependencies} that agents can act on: \texttt{imports} inject prior definitions, \texttt{bounds} propagate constraints to hyperparameter search, and \texttt{baseline} entries enable automatic regression detection, transforming literature review into a machine-executable dependency graph.

\paragraph{The Physical Layer (\texttt{/src}).}
Contains the \emph{how}---executable code calibrated to the contribution type.
Algorithmic contributions use \emph{kernel mode}: only the core modules with typed I/O signatures, often one to two orders of magnitude smaller than the full repository, so that a coding agent can regenerate environment-native boilerplate on demand.
Systemic contributions (CUDA kernels, distributed training, systems architectures) use \emph{repository mode}: the full implementation is retained but annotated via an \texttt{index.md} manifest that maps each source file to the \ara{} component it implements.
\texttt{configs/} annotates every hyperparameter with rationale and search range; \texttt{environment.md} pins dependencies, hardware, and seeds (mode specification, forensic bindings, and detailed rationale in Appendix~\ref{app:src-modes}).

\paragraph{The Exploration Graph (\texttt{/trace}).}
\texttt{exploration\_tree.yaml} stores the complete research directed acyclic graph (DAG) as a nested YAML tree with five typed node kinds---\texttt{question}, \texttt{decision}, \texttt{experiment}, \texttt{dead\_end}, \texttt{pivot}---where nesting encodes parent$\to$child edges and an \texttt{also\_depends\_on} field captures convergence points.
The format functions as a ``git log for research'': agents traverse branches directly, and dead-end nodes preserve the hypothesis, failure mode, and lesson that narrative papers discard (Figure~\ref{fig:exploration-graph}).

\paragraph{The Evidence Layer (\texttt{/evidence}).}
The raw outputs that ground every claim: \texttt{results/} contains machine-readable metric tables and generated data with exact values and source annotations; \texttt{logs/} captures training curves, resource usage, and diagnostics.
It holds \emph{outputs only}, so that every claim's proof chain flows \texttt{claims.md}\,$\to$\,\texttt{experiments.md}\,$\to$\,\texttt{/evidence/}.

\paragraph{Withholding ground-truth enables layered access control.}
Experiment \emph{logic} (what to verify) lives in \texttt{/logic}; experiment \emph{data} (exact results) lives exclusively in \texttt{/evidence}.
A verification agent can be granted the code kernel and algorithm descriptions while the evidence layer is withheld, preventing fabrication by copying expected values.
This separation also makes every \ara{} a ready-to-use training environment: the task lives in \texttt{/logic/experiments.md}, the reward in \texttt{/evidence/}, and preference signals in every accept, reject, and revision logged in \texttt{/trace/}.

We consider an \ara{} sufficient when a \emph{sufficiently capable} coding agent can reproduce the core claim \emph{zero-shot} from it, without human intervention or external context beyond the artifact itself. Sufficiency is therefore a capability-relative criterion: it measures whether the artifact \emph{contains} the information required to reproduce its claim, not whether any present-day agent can fully exploit it. At the limit of agent capability, a complete \ara{} is reproducible by definition, so artifacts authored today remain valid as agents advance.

\FloatBarrier
\section{The Live Research Manager}
\label{sec:live-pm}

The previous section defines the \ara{} protocol: a structured target format for research knowledge.
A natural question follows: \emph{who populates it?}
Requiring researchers to hand-author structured files would simply replace one documentation burden with another, negating the very tax the protocol was designed to eliminate.

\begin{figure}[t!]
\centering
\includegraphics[width=\linewidth]{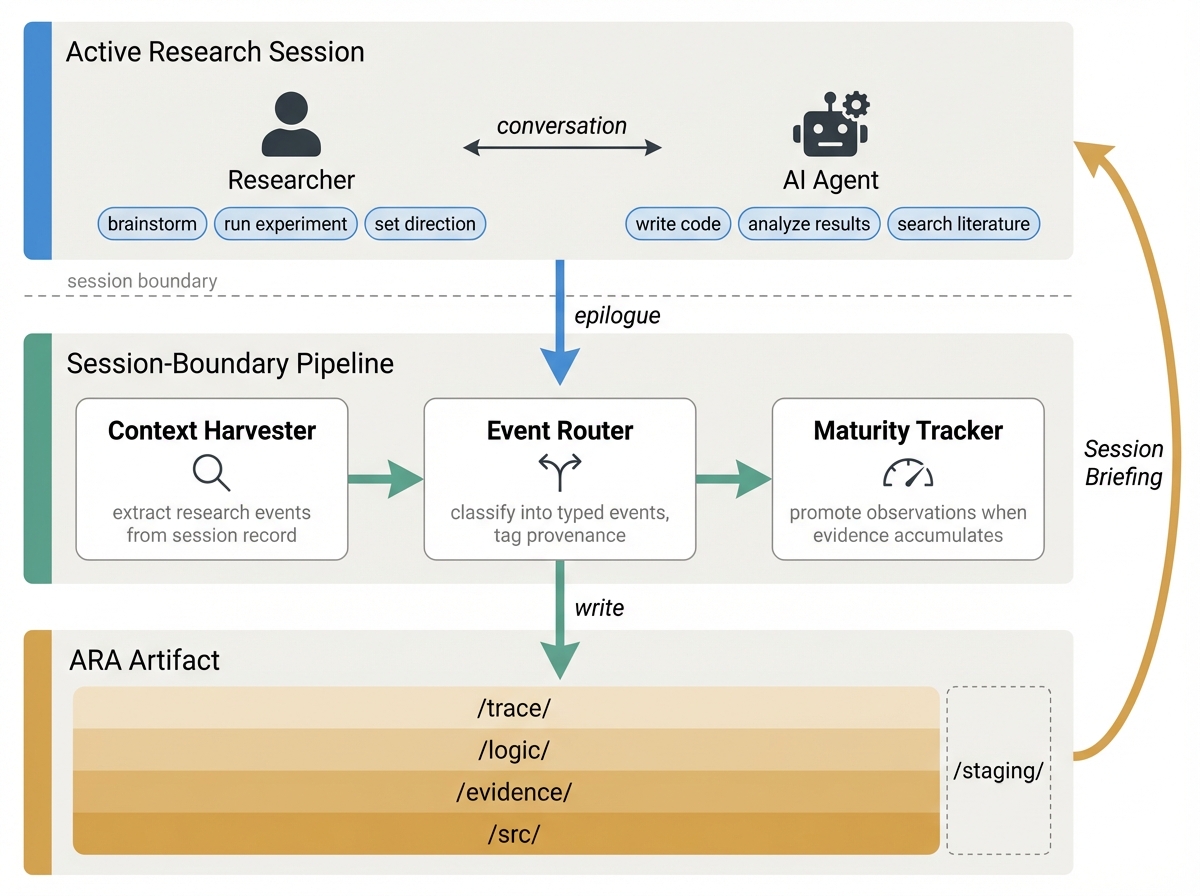}
\caption{The Live Research Manager operates at session boundaries: a three-stage pipeline (Context Harvester $\to$ Event Router $\to$ Maturity Tracker) distills each researcher--agent conversation into typed events that accumulate across \ara{} layers over time.}
\label{fig:lrm-overview}
\end{figure}

The key observation is that in AI-native research, the full research trajectory---every design choice, failed experiment, and hard-won heuristic---already exists as machine-readable text in the conversation between researcher and agent.
The Live Research Manager crystallizes this latent signal into a living \ara{} artifact that accumulates organically in the background, with zero documentation overhead for the researcher.\footnote{\url{github.com/AmberLJC/Agent-Native-Research-Artifact}}
We derive design principles from the AI-native paradigm (\S\ref{sec:design-principles}) and present the system that realizes them (\S\ref{sec:lrm-design}; Figure~\ref{fig:lrm-overview}).

\subsection{Design Principles}
\label{sec:design-principles}

\paragraph{AI-native research.}
Computer science research is increasingly \emph{AI-native}: the researcher collaborates with a general-purpose coding agent across the full research lifecycle.
A single conversational loop may span brainstorming a hypothesis, surveying related work, writing and debugging experiment code, analyzing results, and drafting the paper---the researcher provides direction, the agent executes, and the cycle repeats across days or weeks.
This paradigm shift has a structural consequence: for the first time, the research process is \emph{born digital and born textual}.
Every instruction, intermediate result, design choice, and abandoned direction already exists as machine-readable text---the raw material for a complete research record, lacking only a system to let it crystallize naturally.
Prior efforts to preserve process knowledge---negative-result journals~\citep{matosin2014negativity}, registered reports~\citep{chambers2013registered}---foundered because documentation remained a separate, unrewarded burden on the researcher.
AI-native research dissolves this barrier: the process trace is not an additional deliverable but a \emph{byproduct} of the research itself, generated automatically in every researcher--agent session.
We are, in other words, at the first moment in scientific history where comprehensive research-process capture is feasible at negligible marginal cost.

This observation motivates the Live Research Manager: if the research trajectory is already captured in conversation, a background system can distill it into a structured \ara{} without asking the researcher to document anything.
We derive three design principles from this paradigm.

\begin{enumerate}[label=\textbf{P\arabic*.}, leftmargin=2em, itemsep=1pt]
    \item \textbf{Silent, framework-independent integration}: the manager is a natural-language specification loadable by any coding agent, requiring no custom SDKs. It runs as a background process that silently collects research traces without researcher intervention.
    \item \textbf{Faithful epistemic provenance}: every event is tagged with provenance (\texttt{user}, \texttt{ai-suggested}, \texttt{ai-executed}, \texttt{user-revised}) to preserve epistemic origin. Raw observations are staged rather than forced into categories, maturing progressively into typed events as evidence accumulates, so downstream consumers can reconstruct not just \emph{what} was concluded but \emph{why}.
    \item \textbf{Comprehensive trajectory capture}: the full branching research process, including dead ends and pivots, is recorded with cross-layer bindings established at capture time. The artifact is version-controlled, so each milestone produces a navigable snapshot and retroactive revisions are first-class operations rather than destructive overwrites.
\end{enumerate}
\noindent Full rationale for each principle is in Appendix~\ref{app:lrm-principles}.

\subsection{System Design}
\label{sec:lrm-design}

\paragraph{Overview.}
We implement the Live Research Manager as an \emph{agent skill}~\cite{agentskills2025}: a lightweight, natural-language specification that, when loaded into a general-purpose coding agent's context window, turns it into a domain-specialized agent (\textbf{P1}).
A skill requires no custom SDKs or infrastructure; it simply composes tools the agent already possesses (file read, write, edit, shell execution) with domain knowledge of the \ara{} schema, and artifact quality improves automatically as the underlying language model advances.
The skill is open-source and available at \url{https://github.com/AmberLJC/Agent-Native-Research-Artifact}.

The manager remains silent during active research, then runs a three-stage retrospective pipeline at the end of each conversation (\textbf{P1}; Figure~\ref{fig:lrm-overview}).
The \textbf{Context Harvester} scans the session record---conversational history, tool outputs, experiment results, and code diffs---to extract research-significant events.
The \textbf{Event Router} classifies each event, tags it with provenance, and writes it to the appropriate \ara{} layer.
The \textbf{Maturity Tracker} reviews the staging area, promoting observations that have accumulated sufficient evidence into formal entries and flagging stale or conflicting items.
At the start of the next session, the manager reads the artifact back to surface a structured briefing, closing the loop.
The remainder of this subsection details each stage and the cross-session mechanisms that bind them together.
\paragraph{Context harvesting and event routing.}
The Context Harvester scans the full session record and identifies two categories of research-significant activity: actions the AI agent has already performed (e.g., experiment runs, code changes, literature searches) and directions the researcher has expressed or confirmed (e.g., hypotheses to test, design choices, abandoned approaches).
The Event Router then classifies each extracted event into one of seven types (Table~\ref{tab:event-types}) and tags it with provenance (\textbf{P2})---\texttt{user}, \texttt{ai-suggested}, \texttt{ai-executed}, or \texttt{user-revised}.
An \texttt{ai-suggested} event never auto-upgrades until the researcher explicitly confirms it, preserving the epistemic integrity of the artifact.
Event payloads follow the protocol's factual-density requirement (\S\ref{sec:philosophy}): conversational prose is distilled into telegraphic, quantitative language before committing to the artifact.
Each typed event is written to the appropriate \ara{} layer: trace events (decisions, experiments, dead ends, pivots) append to the Exploration Graph (\texttt{/trace/}), building a faithful, chronological record of the research trajectory including the directions that were tried and abandoned (\textbf{P3}); claims and heuristics enter the Cognitive Layer (\texttt{/logic/}); and events that resist classification are staged in \texttt{/staging/} for later promotion.

\begin{table*}[!t]
\centering
\small
\begin{tabular}{@{}ll@{\hskip 1.5em}ll@{}}
\toprule
\textbf{Event Type} & \textbf{Structured Payload} & \textbf{Event Type} & \textbf{Structured Payload} \\
\midrule
\texttt{decision}    & Choice, alternatives, evidence        &
\texttt{claim}       & Statement, falsification criteria     \\
\texttt{experiment}  & Metrics, claim linkage                &
\texttt{heuristic}   & Trick, sensitivity, bounds            \\
\texttt{dead\_end}   & Hypothesis, failure mode, lesson      &
\texttt{observation} & Raw finding, awaiting classification  \\
\texttt{pivot}       & Trigger, rationale                    &
& \\
\bottomrule
\end{tabular}
\caption{Research event types and structured payloads.}
\label{tab:event-types}
\vspace{-4pt}
\end{table*}

\paragraph{Progressive crystallization.}
Artifact construction proceeds at two timescales (\textbf{P2}).
\emph{Continuously}, at every session boundary, the manager appends trace events to the Exploration Graph, recording how the researcher navigates the open-ended research landscape.
\emph{Periodically}, when a major milestone is reached---a hypothesis confirmed or refuted, a working prototype completed, a key design choice finalized---the Maturity Tracker \emph{crystallizes} the accumulated observations into the more structured layers of the artifact.
Raw observations mature into formal claims in the Cognitive Layer (\texttt{/logic/}), working code is documented in the Physical Layer (\texttt{/src/}), and concepts are added to the knowledge index.
This two-phase rhythm mirrors how research understanding actually develops: insights begin as scattered observations, and forcing premature structure would distort the record.
The manager stages what is not yet classifiable and crystallizes only when \emph{closure signals} in the session record---topic abandonment, explicit researcher affirmation, empirical resolution of linked experiments, or artifact-level commitment---indicate the observation has been treated as settled (Appendix~\ref{app:lrm-maturity}).
When a pivot invalidates an earlier design choice, the manager propagates the change: it updates the affected artifact entries (claims, heuristics, or configuration) to reflect the new direction while the Exploration Graph retains the original rationale and the reason for revision.
Because the artifact is version-controlled, each milestone crystallization produces a commit, giving the project a navigable history of its intellectual evolution, much as GitHub provides for code.

\paragraph{Cross-session continuity.}
The manager is stateless; the artifact itself carries memory across sessions. At session close, it writes a short session record (events captured, claims touched, open threads) and appends to a running session index. The next session reads this index alongside the current claims and staged observations, and raises the relevant pieces only when they bear on the task at hand, rather than leading with a formal briefing the researcher did not ask for (Appendix~\ref{app:lrm-continuity}).

\FloatBarrier
\section{The \ara{} Compiler}
\label{sec:ingestor}

The born-agent pathway (\S\ref{sec:live-pm}) produces the highest-fidelity \ara{}s by capturing knowledge as it emerges during the research process.
However, the scientific record contains millions of legacy PDFs that were never structured, and the born-agent pathway cannot reach backward.
Beyond the PDF itself, research knowledge is distributed across complementary sources: GitHub repositories encode implementation decisions that prose omits, expert-curated evaluation rubrics encode what domain practitioners consider the core claims, and recorded experimental trajectories preserve the failure modes that papers systematically suppress.
To bridge this gap, we introduce the \textbf{\ara{} Compiler}, an agent skill that translates any combination of legacy research sources into a complete \ara{}.\footnote{\url{github.com/AmberLJC/Agent-Native-Research-Artifact}}
Quality is enforced in two stages. During compilation, the Compiler uses only ARA Seal Level~1 as an in-loop validation signal. After compilation, the finished artifact enters the downstream Seal pipeline, where Levels~2--3 evaluate its argumentative support and empirical reproducibility (\S\ref{sec:liveness}).

\begin{figure}[!t]
    \centering
    \includegraphics[width=\linewidth]{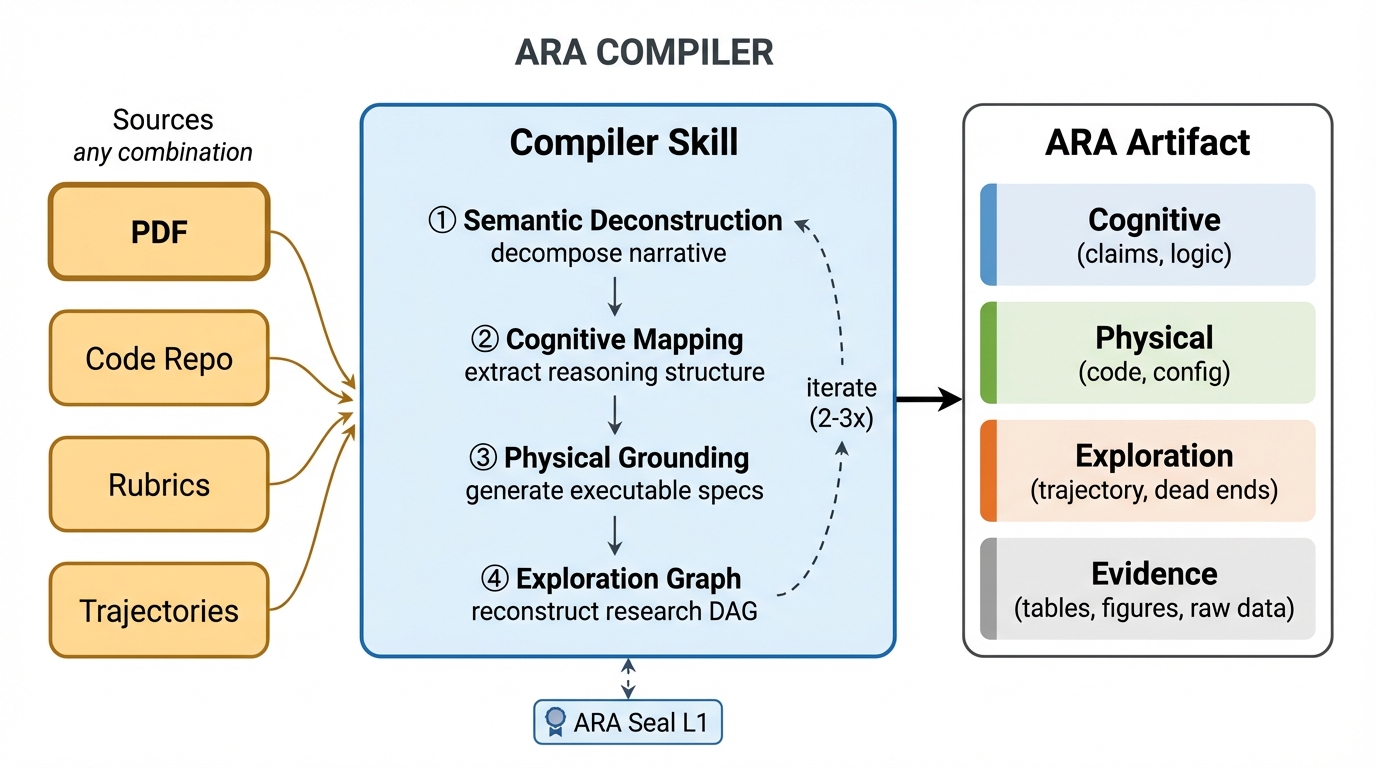}
    \caption{The \ara{} Compiler accepts any combination of research sources and guides a coding agent through four stages of top-down artifact compilation, iterating 2--3$\times$ with in-loop ARA Seal Level~1 validation until the output conforms to the protocol.}
    \label{fig:ingestor-pipeline}
    \end{figure}
    
\subsection{Design Principles}
\label{sec:ingestor-principles}

\paragraph{Universal input, canonical output.}
The Compiler is many-to-one: it accepts any combination of PDFs, code repositories, datasets, human-curated rubrics (e.g., PaperBench~\citep{starace2025paperbench} claim decompositions), and experimental trajectory logs (e.g., RE-Bench~\citep{wijk2025rebench} traces that record failures the paper omits), and always produces a single \ara{} conforming to \S\ref{sec:protocol}.
Degradation is graceful: a PDF alone yields a valid artifact with stub-level physical layers; richer inputs populate progressively more complete layers.

\paragraph{High-fidelity preservation.}
A narrative paper compresses and selects; the Compiler decompresses and restores.
Every numerical result, hyperparameter, architectural detail, and negative finding in the sources must appear \emph{somewhere} in the artifact, and any PDF-accessible information missing from the \ara{} constitutes a compilation failure (evaluated in \S\ref{sec:eval-understanding}).
Preservation is faithful to sources; \emph{enrichment} (\S\ref{sec:collective}) separately surfaces cross-artifact patterns no single source expresses.

\paragraph{Knowledge lineage, not flat extraction.}
Narrative compilation destroys the \emph{provenance chains} connecting claims to experiments, experiments to evidence, and evidence to code.
Plain-text extraction recovers content but not these connections: parsing a PDF into Markdown populates four directories yet leaves them structurally isolated.
The Compiler performs \emph{forensic reconstruction}, recovering the cross-layer forensic bindings (\S\ref{sec:layers}) from sources where lineage exists only implicitly---scattered across prose, figure captions, appendix tables, and code comments---so an agent can trace any claim downstream to code or any number upstream to its hypothesis.
Recovering this lineage, not populating layers, is the core compilation problem.

\subsection{Compiler Implementation}
\label{sec:ingestion-skill}

Realizing these principles against narrative PDFs requires concrete mechanisms that address the specific ways prose obscures structure, lineage, and source diversity.
The Compiler is implemented as a single agent skill (Figure~\ref{fig:ingestor-pipeline}) that guides a coding agent through the translation from legacy sources to \ara{} format.
The skill prompt encodes the full \ara{} schema, field-level requirements, and a structured generation protocol.
We describe the key elements below; the complete skill specification is reproduced in Appendix~\ref{app:ingestor-prompt}.

\paragraph{Top-down generation.}
The skill instructs the agent to construct the artifact \emph{top-down}, mirroring how a researcher explains their work to a new collaborator: high-level concepts first, details next, implementation last.
Concretely, the agent proceeds through four stages.
\emph{Semantic Deconstruction} strips narrative framing to expose raw research content (formulations, configurations, results, dependencies, failed approaches), rewritten in fact-dense telegraphic form that eliminates the Storytelling Tax at the source level.
\emph{Cognitive Mapping} populates \texttt{/logic}: motivation chain (observations $\to$ gaps $\to$ insight), falsifiable claims with proof pointers, formal concepts, and the solution structure, with every claim linked to the experiment that verifies it.
\emph{Physical Grounding} generates \texttt{/src}: annotated configurations, typed code stubs, and the environment manifest; when a code repository is available, stubs are replaced with actual implementations and the agent performs \emph{code-paper reconciliation}, cross-referencing the codebase against claims to surface tacit knowledge---implicit assumptions, undocumented tricks, extra parameters~\citep{li2026tacit}---that is written back to \texttt{/logic} as provenance-tagged heuristics.
\emph{Exploration Graph Extraction} reconstructs the research DAG as a nested YAML tree with dead-end leaf nodes documenting hypothesis, failure mode, and lesson.
This ordering ensures each layer is informed by the one above: cognitive structure guides physical generation, and the exploration graph contextualizes both.

\paragraph{Iterative refinement via validation feedback.}
After initial generation, ARA Seal Level~1 checks (\S\ref{sec:liveness})---schema conformance, cross-layer reference resolution, required field completeness---run within the same agent conversation, returning failures as structured diagnostics that drive targeted fixes.
This generate\,$\to$\,validate\,$\to$\,fix loop typically converges in 2--3 rounds, turning Level~1 into actionable feedback rather than a post-hoc report.

\paragraph{Source-aware enrichment.}
\label{sec:collective}
Auxiliary sources are routed to the layer they most directly populate: code repositories replace stubs with verified implementations in \texttt{/src}; evaluation rubrics anchor \texttt{/logic} with expert-verified claim decompositions; and trajectory logs seed \texttt{/trace} with dead-end nodes the PDF omits.
When a library of previously compiled \ara{}s is available, the Compiler further performs \emph{collective inference}: it retrieves heuristics and configurations from same-domain artifacts, flags common patterns the current paper omits, and adds them as candidate heuristics tagged \texttt{collective\_inference} so downstream agents can distinguish stated from inferred knowledge.

\FloatBarrier
\section{\ara{} Verification and Review}
\label{sec:review-system}

Expert human attention is the scarcest resource in scientific evaluation.
Review loads at top venues have grown faster than the reviewer pool~\citep{aczel2021billiondollar}, and reviewer bandwidth is increasingly consumed by mechanical verification (``does the code run?'', ``does Table~3 support Claim~2?'') rather than the substantive judgment that only domain experts can provide.

Because an \ara{} submission \emph{is} a structured, machine-executable artifact, structural properties that PDF review checks subjectively become objectively verifiable: schema conformance and cross-layer reference integrity either pass or fail deterministically.
Higher-order properties, information completeness and directional reproducibility of central claims, become \emph{machine-assessable}: automated checks provide probabilistic evidence that informs human reviewers rather than replacing them.
This separation redirects expert attention from mechanical checking to the substantive judgment that only humans can provide: significance, novelty, and taste.
We define the \textbf{ARA Seal} as a machine-verifiable credential with three escalating levels (\S\ref{sec:liveness}, Figure~\ref{fig:ara-seal}) and describe a three-stage review pipeline (Figure~\ref{fig:review-pipeline}) that consumes those levels to separate automated verification from human judgment.

\subsection{Design Principles}
\label{sec:review-principles}

We derive two design principles below; cross-artifact review via agent-to-agent ARA comparison, which becomes meaningful only as the corpus reaches critical mass, is discussed in \S\ref{sec:discussion}.

\paragraph{P1. Automate the mechanical; reserve humans for judgment.}
Structural validity, internal consistency, and claim reproducibility are objective properties that either pass or fail, whereas significance, novelty, and taste require domain expertise.
The review system should never ask a human to verify that Experiment~E03 matches Claim~C02 when a machine can do so in seconds; resolving all machine-checkable issues \emph{before} the artifact reaches a human reviewer ensures that expert attention is spent exclusively on questions that genuinely require it.

\paragraph{P2. Reproducibility as a foundational requirement.}
``Code available upon request'' nominally satisfies today's reproducibility bar; in an \ara{}-native system, reproducibility is a \emph{machine-verified property} of the artifact itself.
Passing ARA Seal Level~1 (structural integrity) is a submission requirement, and Level~2 (argumentative rigor) produces a structured critique before the venue spends compute on Level~3 (execution reproducibility), whose results are then attached to every review.
Artifacts that fail structural checks, or whose claims remain obviously under-supported after Level~2 critique, do not advance to human review.

\subsection{The ARA Seal: Machine-Verifiable Research Credentials}
\label{sec:liveness}

\begin{figure}[!t]
    \centering
    \includegraphics[width=\linewidth]{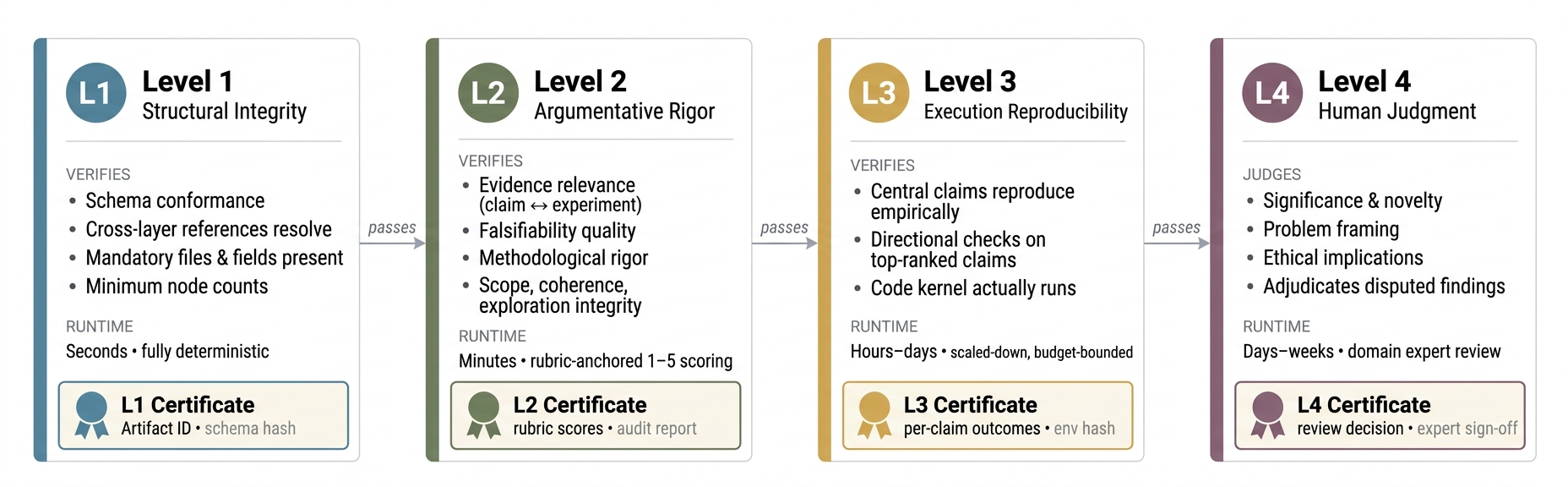}
    \caption{The ARA Seal is a three-level verification credential. Each level tests a progressively stronger property of the artifact, escalating in cost and breadth: structural integrity (seconds, deterministic), argumentative rigor (minutes, rubric-anchored agent), and execution reproducibility (hours to days, sandboxed coding agent). Passing the applicable levels issues a Seal Certificate that downstream agents check before investing compute.}
    \label{fig:ara-seal}
\end{figure}

A PDF paper earns trust indirectly: venue prestige, citation counts, and author reputation serve as proxies for quality, but none verify the work itself.
Because an \ara{} encodes research as typed, machine-traversable data with explicit claim--evidence bindings, its quality becomes \emph{directly verifiable}.
The ARA Seal operationalizes this as a three-level verification protocol (Figure~\ref{fig:ara-seal}), where each level tests a progressively stronger property (implementation details in Appendix~\ref{app:seal-details}).

\begin{description}[leftmargin=1.5em, itemsep=4pt, topsep=2pt]
    \item[Level 1 -- Structural Integrity]
    verifies that the artifact is well-formed and internally consistent:
    the directory ontology exists, all structured files conform to their schema (each claim carries \texttt{Statement}, \texttt{Status}, \texttt{Falsification criteria}, and \texttt{Proof}; each heuristic carries \texttt{Rationale}, \texttt{Sensitivity}, and \texttt{Bounds}),
    and all cross-layer references resolve (experiment IDs in \texttt{claims.md} point to valid entries in \texttt{experiments.md}, code references trace to implementations in \texttt{/src}).

    \item[Level 2 -- Argumentative Rigor]
    without executing any code or consulting external sources, a \emph{Rigor Auditor} agent evaluates whether the content of a Level-1-valid artifact is epistemically sound along six objective dimensions, each scored on an anchored 1 to 5 rubric.
    The three load-bearing dimensions are:
    \emph{evidence relevance}, checking that every claim's cited experiments substantively address what the claim asserts under type-aware entailment (causal claims require isolating ablations, generalization claims require heterogeneous test conditions, improvement claims require baseline comparisons);
    \emph{falsifiability quality}, checking that criteria are actionable, non-tautological, scope-matched, and independently testable without access to proprietary data;
    and \emph{methodological rigor}, covering baseline adequacy, ablation coverage, statistical reporting, and metric--claim alignment.
    Three further dimensions---scope calibration, argument coherence, and exploration integrity---are defined in Appendix~\ref{app:seal-eval-l2}.
    Findings are collected with severity (\emph{critical}, \emph{major}, \emph{minor}, \emph{suggestion}), verbatim evidence spans, and actionable suggestions; the overall grade is derived from the mean score and per-dimension floors.
    Every check reduces to a rubric-anchored property of the artifact's content, so Level~2 remains objective; judgments of significance, novelty, and taste are reserved for human reviewers in Stage~3.
    The reference Rigor Auditor is released as an agent skill~\citep{agentskills2025};\footnote{\url{github.com/AmberLJC/Agent-Native-Research-Artifact}} full protocol, dimension rubrics, and grade thresholds appear in Appendix~\ref{app:seal-eval-l2}.

    \item[Level 3 -- Execution Reproducibility]
    verifies that the artifact's central claims reproduce empirically.
    The system selects claims by criticality---those in the contribution list, those anchoring the most downstream dependencies, or those flagged by the authors---and runs scaled-down directional checks (small data, few epochs, toy configurations) that test whether claimed properties hold qualitatively rather than reproducing exact numbers.
    The verifying agent is isolated from the artifact's evidence layer: it receives only the code kernel and algorithm descriptions, never the reported numbers, preventing fabrication by copying expected outcomes.
    Venues set a compute budget; claims that exceed it are flagged as \emph{unverified}.
    Full-scale reproduction (original datasets, full training runs, exact metric recovery) is optional and typically post-acceptance or community-driven; results are appended to the living Seal certificate.
\end{description}

Passing the applicable levels issues a \textbf{Seal Certificate}---a signed record of artifact ID, verification level achieved, timestamp, environment hash, and per-claim reproduction outcomes.
Downstream agents check this certificate before investing compute, avoiding redundant re-verification.

\subsection{Three-Stage Review Pipeline}

\begin{figure}[!t]
    \centering
    \includegraphics[width=\linewidth]{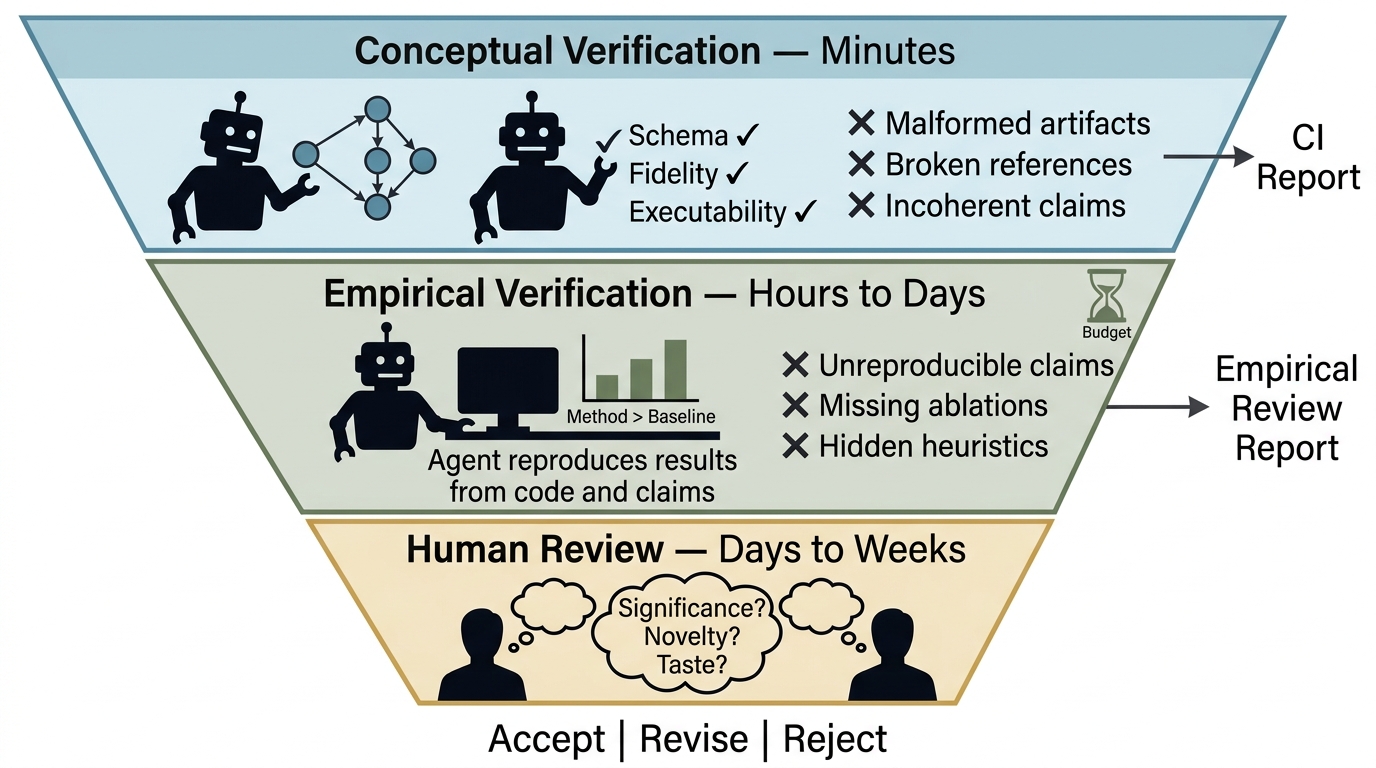}
    \caption{Three-stage \ara{}-native review pipeline. Stages~1--2 invoke the ARA Seal levels of Figure~\ref{fig:ara-seal} to resolve mechanical and rigor issues before human reviewers engage, redirecting expert attention to novelty and significance.}
    \label{fig:review-pipeline}
\end{figure}

We envision a three-stage review pipeline (Figure~\ref{fig:review-pipeline}) that mirrors CI/CD practices, where each stage gates the next: conceptual verification, empirical verification, and human judgment.

\paragraph{Stage 1: Conceptual Verification (Minutes).}
The first stage checks whether the artifact is \emph{well-formed and conceptually well-supported} without running any experiments.
Automated checks validate structural integrity (ARA Seal Level~1---schema conformance, cross-layer reference resolution, required field completeness), and the Rigor Auditor executes Level~2 argumentative rigor by assessing whether major claims are appropriately scoped, whether attached evidence and ablations justify them, and whether obvious baselines, assumptions, or limitations are missing.
Level~1 checks are mechanical and deterministic: either every claim in \texttt{claims.md} links to a valid experiment, or it does not; either heuristics carry sensitivity bounds, or they are missing.
Level~2 instead produces a structured rigor report keyed to specific claims, experiments, and omissions.
The stage also generates \emph{advisory diagnostics} surfaced to human reviewers in Stage~3: whether the exploration tree contains dead-end nodes (suggesting genuine process documentation versus a sanitized linear chain), whether the related-work graph covers obvious baselines, and whether experiment--claim coverage has gaps.
These diagnostics inform but do not gate; only the Seal checks are pass/fail requirements.
The output is a \textbf{CI Report} plus a Level~2 rigor report, both attached to the submission and visible to authors and reviewers.
Authors iterate on structural failures and rigor critiques before the artifact advances, analogous to fixing lint errors and design issues before code review.

\paragraph{Stage 2: Empirical Verification (Hours--Days).}
Once the artifact passes conceptual review, the second stage tests whether the \emph{claimed results actually reproduce}.
An AI reviewer agent executes ARA Seal Level~3: it ranks the artifact's claims by criticality---prioritizing those in the contribution list and those with the most downstream dependents in the claim graph---then runs scaled-down directional checks (small data, few epochs, toy configurations) within a venue-specified compute budget.
The agent is isolated from the artifact's evidence layer and ground-truth results: it receives only the code kernel and algorithm descriptions, never the paper's reported numbers, preventing fabrication by copying expected outcomes.
For each tested claim, it reports whether the claimed direction holds (e.g., ``method A outperforms baseline B on metric M''), wall-clock time consumed, and any divergence from expected outcomes.
Claims that exceed the remaining budget are flagged as \emph{unverified} with an estimate of the compute required.
Beyond reproduction, the agent assesses experimental comprehensiveness: Are ablations present for each design choice?
Do the experimental conditions cover the claimed generality, or are results cherry-picked from favorable settings?
Are there undocumented heuristics in the code that do not appear in the artifact's cognitive layer?
The output is an \textbf{Empirical Review Report} that records which claims were verified, which failed, which were deferred due to budget, and what experimental gaps were identified.

\paragraph{Stage 3: Human Review (Days--Weeks).}
Human reviewers receive the submission alongside the CI Report and Empirical Review Report, and their role shifts from \emph{verification}---already handled by Stages~1--2---to \emph{judgment}.
They focus on the questions that no machine can answer: Is this contribution significant---does it address a real problem that matters?
Is the core insight genuinely novel, or an incremental recombination of known ideas?
Is this the right formulation of the problem, and are there better approaches the authors should have considered?
What are the ethical implications and potential for misuse?
Where the AI reviewer's findings are contested by the authors, humans adjudicate with the full audit trail available for inspection.
Human reviewers write structured reviews in the same typed format---findings linked to specific \ara{} components---ensuring that all feedback is actionable and traceable.
The key efficiency gain: humans no longer spend time on ``your code doesn't run'' or ``Table~3 contradicts Claim~2''---these are resolved before the artifact reaches them.

\FloatBarrier
\section{The (Human+AI)$^{2}$ Research Network}
\label{sec:network}

\begin{figure}[!t]
    \centering
    \includegraphics[width=0.85\linewidth]{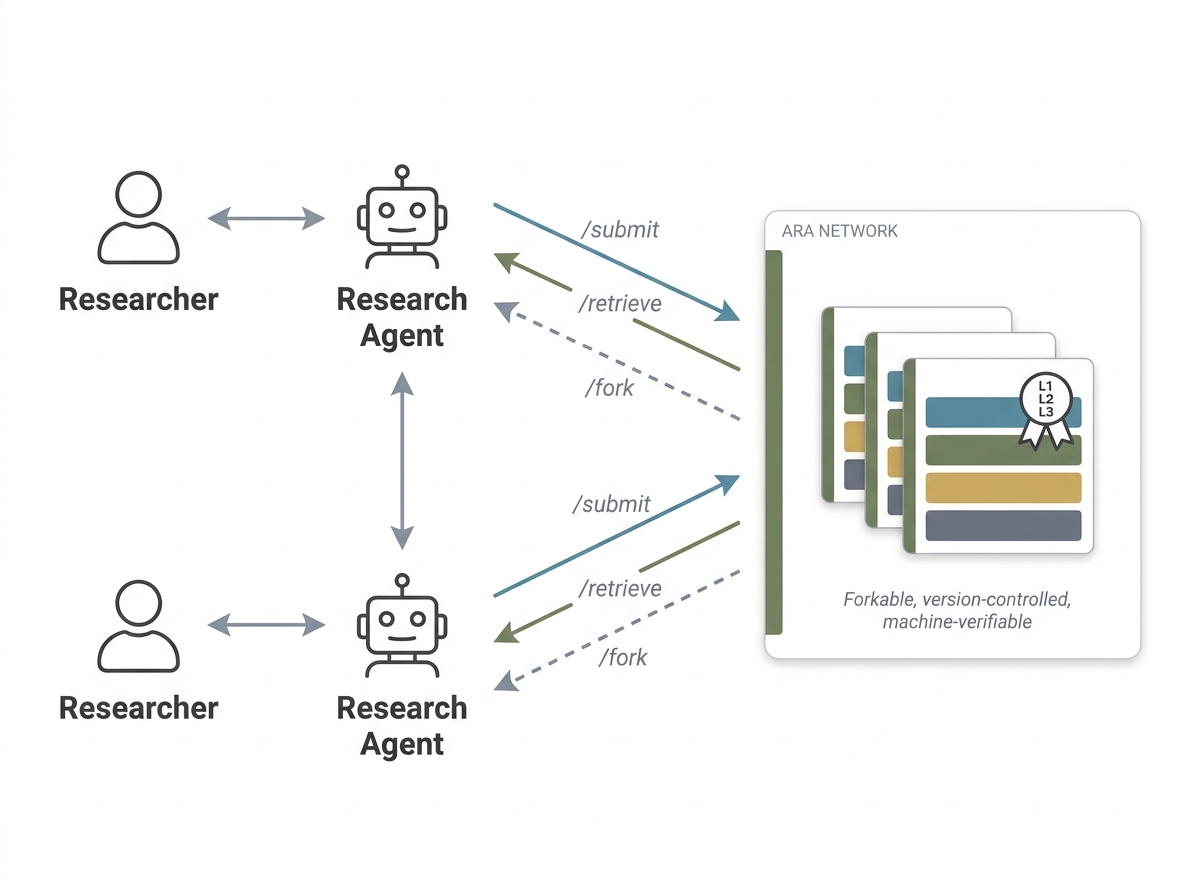}
    \caption{The (Human+AI)$^{2}$ research network. Each researcher works through a research agent that interfaces with a shared \ara{} network via \texttt{/submit}, \texttt{/retrieve}, and \texttt{/fork}; agents may also collaborate directly.}
    \label{fig:network}
\end{figure}

The previous sections describe the components of an \ara{}-native stack: the protocol (\S\ref{sec:protocol}), the Live Research Manager that captures work into the artifact during development (\S\ref{sec:live-pm}), the Compiler that imports the legacy literature (\S\ref{sec:ingestor}), and the Seal-gated review pipeline that verifies it (\S\ref{sec:review-system}). Composed, they form a scientific communication system whose primary object is no longer the static document but the \ara{} itself: a single canonical artifact that humans on each end direct agents to author, certify, render, and extend. We call the resulting structure a \textbf{(Human+AI)$^{2}$ network} (Figure~\ref{fig:network}).

On the producer side, researchers no longer work toward papers; they pursue questions, and the paper-as-output accrues automatically as an \ara{} along the way: the Live Research Manager folds each decision, dead end, and confirmed claim into the artifact during ordinary work, the Compiler imports legacy sources on demand, and at any milestone the artifact is routed through the Seal pipeline and registered publicly, where another team can fork a passing artifact, extend a claim, retain attribution to the parent, and submit the diff for re-review. 
On the consumer side, because the \ara{} is canonical, an agent renders it on demand into whatever surface the reader needs (paper, video, slides, interactive demo, or grounded dialogue), shaped by the reader's expertise, attention budget, and intent. 

With both ends agent-mediated and the artifact the only persistent state, contributions compound at the level of artifacts rather than sentences: publishing becomes a Git-like operation, reviewers consume Seal-attested artifacts through their preferred surfaces, and downstream agents read \ara{}s as structured baselines, training environments, or starting points for new questions. The result is a queryable scientific commons in which every contribution is an executable diff, and the cost of understanding, reproduction, and extension falls with each new artifact admitted rather than rising against it. \S\ref{sec:eval} asks whether today's agents can already operate this substrate; \S\ref{sec:discussion} maps near-, medium-, and long-term extensions.

\FloatBarrier
\section{Evaluation}
\label{sec:eval}

\begin{table*}[t]
  \centering
  \small
  \begin{tabular}{@{} p{0.46\linewidth} | p{0.46\linewidth} @{}}
  \toprule
  \textbf{PaperBench}~\citep{starace2025paperbench} & \textbf{RE-Bench}~\citep{wijk2025rebench} \\
  \midrule
  \textbf{Papers:} 23 peer-reviewed ML papers across diverse subfields &
  \textbf{Tasks:} 7 R\&D hill-climbing tasks on well-defined objective functions \\
  \midrule
  \textbf{Scale:} 8,921 expert-authored rubric requirements &
  \textbf{Scale:} 24,008 agent runs; 46,303 failure episodes \\
  \midrule
  \textbf{Contents:} PDF + companion GitHub repo (15/23 papers); expert rubrics encode hyperparameter values, implementation tricks, and configurations absent from the paper; failure knowledge rarely preserved in published papers &
  \textbf{Contents:} Starter codebase per task; METR MALT transcripts with full real agent successful and failure trajectories retained per run \\
  \midrule
  \textbf{Scoring:} Expert hierarchical rubric (yes\,/\,partial\,/\,no) &
  \textbf{Scoring:} Automated continuous objective; no human judgment required \\
  \midrule
  \textbf{Used in:} Understanding (\S\ref{sec:eval-understanding}, Cat.~A\,\&\,B); Reproduction (\S\ref{sec:eval-reproduction}) &
  \textbf{Used in:} Understanding (\S\ref{sec:eval-understanding}, Cat.~C); Extension (\S\ref{sec:eval-extension}) \\
  \bottomrule
  \end{tabular}
  \caption{Benchmark characteristics. PaperBench supplies configuration depth via expert rubrics; RE-Bench supplies trajectory depth via MALT failure traces. These are the two source-side enrichments \ara{} is built to exploit.}
  \label{tab:benchmarks}
  \end{table*}
We evaluate \ara{} across three layers of increasing research ambition: \emph{understanding} (can an agent extract knowledge from the artifact?), \emph{reproduction} (can an agent execute research from the artifact?), and \emph{extension} (can an agent build on prior work more efficiently with failure knowledge?).
Each layer isolates a distinct advantage of structured artifacts over the PDF-centric status quo.
We first describe the evaluation corpus and how each \ara{} is built from it (\S\ref{sec:eval-data}), then present experiments that test \ara{} at each layer (\S\ref{sec:eval-understanding}--\ref{sec:eval-extension}). Full implementation details for all experiments are in Appendix~\ref{app:eval-details}.

\subsection{Datasets and Motivation}
\label{sec:eval-data}
\label{sec:eval-setup}\label{sec:eval-gap}\label{sec:eval-t1}\label{sec:eval-negative}\label{sec:eval-t3-exec}

A single comparison underlies all three evaluation layers: holding the agent, the task, and the ground truth fixed, does an \ara{} compiled from a research project's available source materials outperform the conventional artifact most readers receive---a paper PDF plus a companion code repository when one exists?

\paragraph{Two systematic gaps in the conventional artifact.}
The comparison is interesting precisely because the conventional PDF$+$repo loses two kinds of content an \ara{} can preserve. First, \emph{configuration depth}: published papers omit hyperparameters, environment specs, and implementation details that experts deem necessary for reproduction. Across PaperBench's 8{,}921 expert reproduction requirements, only \textbf{45.4\%} are fully specified in the source PDF (Fig.~\ref{fig:info-gap}; full taxonomy in App.~\ref{app:gap-types}). Second, \emph{exploration depth}: most of the compute that produces a finished result never reaches the artifact. Across 24{,}008 RE-Bench agent runs, \textbf{90.2\%} of dollar cost is spent on failed exploration that the published artifact discards entirely (App.~\ref{app:exploration-tax}), forcing each subsequent agent to rediscover the same dead ends. We use two benchmarks chosen so that each one supplies, as a first-class source, exactly what the conventional artifact lacks: PaperBench expert rubrics close the configuration gap; RE-Bench MALT trajectories close the exploration gap (Table~\ref{tab:benchmarks}). Both benchmarks happen to be drawn from computer science and machine learning, but \ara{}'s design extends to any \emph{digitalised} research whose contribution can be expressed as code, configurations, and grounded data, including computational social science, economics, and dry-lab biology. Research whose contribution is a physical-world intervention (wet-lab biology, materials synthesis) falls outside this scope until the underlying experimental record is itself digitalised.

\paragraph{Two representations of the same research.}
For each paper or task we construct two representations of the same underlying research and hold everything else fixed.
The \emph{conventional baseline} approximates what most readers receive: for PaperBench it is the published paper PDF and, when available (15 of 23 papers), the original GitHub repository; for RE-Bench, where tasks have no published paper, it is an LLM-synthesised polished paper-style writeup of the official reference solution paired with the official source---the closest analogue to PDF$+$repo for paperless tasks.
The \ara{} is compiled from the same source bundle (PDF$+$repo$+$expert rubric for PaperBench; official solution$+$source$+$MALT trajectories for RE-Bench) via the \ara{} Compiler (\S\ref{sec:ingestor}), gated by the ARA Seal Level~1 structural-integrity loop (\S\ref{sec:liveness}); the generate--validate--fix iteration converges in 1--3 passes in practice.
For the extension experiment (\S\ref{sec:eval-extension}), RE-Bench \ara{}s are additionally produced through a specialised pipeline (App.~\ref{app:extension-pipeline}) that adds per-MALT-run extraction sub-agents and a beat-reference fairness filter; the output conforms to the same \ara{} schema.

\subsection{Knowledge Extraction from \ara{}}
\label{sec:eval-understanding}
\label{sec:eval-info-retrieval}\label{sec:eval-questions}\label{sec:eval-t2}

The first layer measures what an agent can extract from each format: does \ara{} preserve the information present in the source PDF and repo, and does its structure surface knowledge that the source leaves scattered or implicit?
A format that loses information during compilation cannot improve reproduction or extension, so this is a precondition for every downstream benefit.
The information gap quantified in \S\ref{sec:intro} (Figure~\ref{fig:info-gap}) shows that structure \emph{can} recover details PDFs under-specify; this layer tests whether the Compiler actually achieves that promise.

\paragraph{Setup.}
We probe each format with 450 questions: 15 per evaluation target across 30 targets (23 PaperBench papers and 7 RE-Bench tasks), in three categories. \emph{Category~A} (10 per target, both benchmarks) tests fidelity on surface results, methods, and conditions answerable from the PDF. \emph{Category~B} (5 per PaperBench paper) tests configuration recovery on hyperparameters, environment specs, and preprocessing details. \emph{Category~C} (5 per RE-Bench task) tests failure knowledge on dead ends, alternatives considered, and lessons. To avoid source bias in question selection, we generate two pools per target independently---one seeded by reading the PDF, one by reading the \ara{}---then merge and deduplicate them so that neither format dictates which questions get asked (full templates and the gold-reference rubric in Appendix~\ref{app:question-bank}). Each (target, format, question) triple is dispatched as an independent Claude Sonnet 4.6 sub-agent with a fresh context, and each answer is graded ternary (1.0 / 0.5 / 0.0) by a Claude Opus 4.6 judge against a gold reference.

\begin{table}[t]
\centering
\small
\begin{tabular}{lc cc cc}
\toprule
& & \multicolumn{2}{c}{\textbf{Accuracy (\%)}} & \multicolumn{2}{c}{\textbf{Tokens (K/Q)}} \\
\cmidrule(lr){3-4} \cmidrule(lr){5-6}
\textbf{Category} & $n$ & \textbf{ARA} & \textbf{BL} & \textbf{ARA} & \textbf{BL} \\
\midrule
A: Fidelity                & 300 & \textbf{95.6} & 80.8 & 84.6 & 88.5 \\
\quad PaperBench           & 230 & \textbf{96.7} & 89.8 & 86.3 & 97.7 \\
\quad RE-Bench             &  70 & \textbf{92.1} & 51.4 & 79.0 & 58.2 \\
\addlinespace
\shortstack[l]{B: Detail \\ \scriptsize(PaperBench)} & 115 & \textbf{92.6} & 67.8 & 183.0 & 178.3 \\
\addlinespace
\shortstack[l]{C: Failure \\ \scriptsize(RE-Bench)}  &  35 & \textbf{81.4} & 15.7 & 139.3 & 58.0 \\
\midrule
\textbf{Overall}           & \textbf{450} & \textbf{93.7} & 72.4 & 114.0 & 109.1 \\
\bottomrule
\end{tabular}
\caption{Understanding evaluation: accuracy and per-question token usage across 450 paired outcomes. ARA wins at every category and every benchmark; the per-category mechanism is unpacked in Appendix~\ref{app:understanding-per-category}.}
\label{tab:understanding}
\end{table}

\paragraph{Results.}
\ara{} outperforms the baseline at every category and every benchmark, with overall accuracy \textbf{93.7\% vs.\ 72.4\%} ($+21.3$\%) on 450 paired outcomes.
The aggregate gap decomposes along three distinct mechanisms, each isolated by one category.
On \textbf{Category~A}, where the answer is recoverable from the PDF, \ara{} wins $+14.8$\% \emph{while consuming 12\% fewer tokens}: PAPER.md's layer index turns linear document scanning into targeted file lookup.
On \textbf{Category~B}, where the rubric supplies configuration details papers systematically omit, \ara{} wins $+24.8$\% at comparable token usage; the baseline can mine the companion repo, but \ara{}'s \texttt{src/configs/} centralises the same knowledge in a single agent-friendly file.
On \textbf{Category~C}, where the answer lives only in the MALT trajectories, \ara{} wins $+65.7$\%; the baseline has no source for failure knowledge and abandons most queries with short empty answers (58K vs.\ 139K tokens).

\paragraph{Token usage scales with question depth on \ara{}, not on the baseline.}
On \ara{} the agent spends 61K tokens on explicit questions, 96K on scattered ones, and 153K on implicit-failure ones; baseline token usage stays roughly flat across the same tiers (83K to 118K), since linear PDF/repo scanning costs the same regardless of how deep the answer is. Per-category mechanism analysis and the full difficulty stratification are in App.~\ref{app:understanding-eval}.

\subsection{Reproduction from \ara{}}
\label{sec:eval-reproduction}
\label{sec:eval-repro-cost}\label{sec:eval-t3-ext}

The second layer tests whether structured artifacts translate understanding into action: can an agent reproduce experimental results more effectively from an \ara{} than from the traditional PDF$+$GitHub combination? Layer~1 measures what an agent knows; Layer~2 measures what it can do. The information gap from \S\ref{sec:intro} (Fig.~\ref{fig:info-gap}; 54.6\% of expert reproduction requirements partially or entirely absent from PDFs) names the concrete execution blockers \ara{}'s structured layers are designed to remove.

\paragraph{Task curation.}
We select 15 PaperBench papers with companion GitHub repositories and curate 10 reproduction tasks per paper (150 tasks total, 1{,}743 rubric requirements), stratified by difficulty (50 easy, 49 medium, 51 hard).
Tasks describe \emph{what} to reproduce, not how; within each paper, subtasks are ordered by difficulty so the agent builds cumulatively (task design details in Appendix~\ref{app:repro-scoring}).

\paragraph{Protocol.}
For each paper, two coding agents receive the same mega-task but different source materials:
\begin{itemize}[leftmargin=*, itemsep=1pt, topsep=2pt]
  \item \textbf{ARA agent}: the \ara{} artifact only (\texttt{PAPER.md}, \texttt{logic/}, \texttt{src/}, \texttt{evidence/}). No PDF or repository access.
  \item \textbf{Baseline agent}: the paper PDF and companion GitHub repository.
\end{itemize}
Both agents use Claude Sonnet 4.6 with the same system prompt (differing only in source material paths) and per-paper token budgets of 14--20M tokens scaled by task complexity (cache reads discounted to 10\%).
Expected numerical results are masked in agent prompts (replaced with \texttt{[X]\%}) to prevent parroting.
After each run, a blinded Claude Opus 4.6 judge evaluates every rubric requirement as \emph{yes} (satisfied), \emph{partial} (partially addressed), or \emph{no} (not met), without knowing which condition produced the output.

\paragraph{Scoring.}
The primary metric is the \textbf{difficulty-weighted success rate}, weighting easy, medium, and hard subtasks at $1\!:\!2\!:\!3$ to emphasize harder tasks where structured information provides the most leverage (scoring formula, statistical tests, and per-difficulty breakdowns in Appendix~\ref{app:repro-scoring}).

\paragraph{Results.}
Across all 15 papers with complete paired runs (150 subtasks, 1{,}743 rubric requirements), \ara{} achieves a difficulty-weighted success rate of \textbf{64.4\%} vs.\ \textbf{57.4\%} for the baseline; the win/tie/loss breakdown across papers is 8/5/2 (per-paper numbers in Appendix~\ref{app:repro-per-paper}).
Figure~\ref{fig:repro-difficulty} stratifies this aggregate by difficulty; the per-paper breakdown is deferred to Appendix~\ref{app:repro-per-paper} (Figure~\ref{fig:repro-difficulty-heatmap}).

\begin{figure}[t]
\centering
\includegraphics[width=0.9\linewidth]{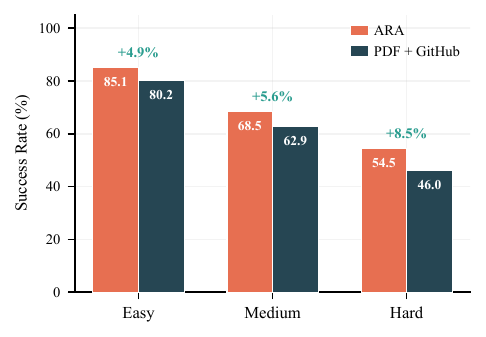}
\caption{Aggregate reproduction success rates across all 15 papers, stratified by difficulty. The \ara{} advantage widens monotonically with difficulty ($+4.9\%$ easy, $+5.6\%$ medium, $+8.5\%$ hard), tracking the tiers where reproduction depends most heavily on configuration content the PDF underspecifies.}
\label{fig:repro-difficulty}
\end{figure}

\paragraph{Analysis.}
The \ara{} advantage grows with difficulty: easy subtasks (environment setup, model instantiation) are near-ceiling for both formats, and the gap opens on medium and hard subtasks where reproduction depends on configuration content the published PDF rarely supplies. The same pattern holds paper by paper (App.~\ref{app:repro-per-paper}): the largest \ara{} advantages (\texttt{fre}, \texttt{mechanistic-understanding}, \texttt{pinn}) are on papers with multi-stage training pipelines whose hyperparameter interactions PDFs describe only at a high level, and the gain concentrates in their medium and hard columns. For example, the \texttt{fre} \ara{} agent reimplemented the original JAX codebase in PyTorch (1.8\,GB GPU vs.\ JAX's 30.8\,GB), trained 17 models across three domains, and completed all medium and hard subtasks; the baseline agent fought the JAX environment and completed only 3 training attempts before its budget ran out. The one clear baseline win is \texttt{self-expansion}, where the \ara{} agent fabricated results that the blinded judge caught; narrow ties (\texttt{adaptive-pruning}, \texttt{rice}) occur on papers whose companion repositories partially compensate for the PDF gap. Across all 15 papers, fabrication occurred in 2 baseline runs and 1 \ara{} run: structured artifacts reduce but do not eliminate hallucinated results.

\subsection{Extension from \ara{}}
\label{sec:eval-extension}
\label{sec:eval-extensibility}\label{sec:eval-pitfalls}\label{sec:eval-analysis}

The third layer tests \ara{}'s most ambitious claim: that preserving the failure trajectory of prior research lets the next agent extend it more effectively.
Section~\ref{sec:eval-data} documented that 59.2\% of agent tokens (and 90.2\% of dollar cost) on RE-Bench are spent on dead-end exploration that the published artifact discards.
This layer asks whether handing the next agent a structured record of what was already tried and abandoned closes that gap in measurable downstream gains.

\paragraph{Why RE-Bench, and which tasks.}
Testing the claim demands tasks that admit open-ended improvement \emph{and} ship a corpus of \emph{real} agent failure traces, not author-imagined ones.
RE-Bench~\citep{wijk2025rebench} satisfies both: each of its 7 tasks has a continuously valued automated scorer, and the METR MALT transcripts contain thousands of complete agent runs (with their dead ends) per task.
We use 5 of the 7 tasks: \texttt{triton\_cumsum}, \texttt{restricted\_mlm}, \texttt{fix\_embedding}, \texttt{nanogpt\_chat\_rl}, and \texttt{rust\_codecontests}.
The other two are excluded because their MALT corpora cannot supply a usable failure-trace layer: \texttt{optimize\_llm\_foundry} ships no MALT corpus at all, and \texttt{small\_scaling\_law}'s MALT runs are sparse, lack Claude-4 entries, and consist mostly of trivial parameter sweeps with no recorded strategic exploration; either case would leave \texttt{trace/} effectively empty by construction (Appendix~\ref{app:extension-tasks}).

\paragraph{\ara{} construction (specialised RE-Bench pipeline).}
RE-Bench inputs differ from a typical paper~+~repo: the source is an official reference solution plus thousands of MALT JSONL transcripts, and the failure-record layer dominates artifact value here.
We therefore wrap the standard \ara{} compiler (\S\ref{sec:ingestor}) in a RE-Bench-specific pipeline (Appendix~\ref{app:extension-pipeline}) that (i)~lifts the official solution into \texttt{src/} and the reference-derived knowledge into \texttt{logic/} via the standard compiler, then (ii)~fans out one extraction sub-agent per MALT run to populate \texttt{trace/exploration\_tree.yaml} and \texttt{evidence/tables/malt\_attempts.md} with the dead ends and partial successes those runs produced.
A direction-aware \emph{beat-reference filter} excludes any MALT scoring attempt that exceeded the reference, applied per attempt rather than per run.
This is the experiment's central fairness rule: comparing \ara{} against a polished paper writeup is only meaningful if neither side carries a worked-out beating-reference solution that the agent could simply transcribe.
Every retained node also carries a \texttt{source: official-solution} or \texttt{source: MALT~run\_id=\{id\}} provenance tag.

\paragraph{Protocol.}
Both agents start from an identical workdir: the official reference \texttt{solution.py}, a \texttt{score.sh} wrapper around the canonical RE-Bench scorer, training-data symlinks where applicable, and a \texttt{reference/} directory whose contents are the only independent variable.
\begin{itemize}[leftmargin=*, itemsep=1pt, topsep=2pt]
  \item \textbf{Paper agent}: reads \texttt{reference/paper.md}, an LLM-synthesised academic-style writeup of the official solution (abstract, method, results, dev notes; the same beat-reference filter is applied so \texttt{paper.md} never contains a worked-out beating-reference variant), plus the official \texttt{src/} tree. This emulates the artifact a conventional published paper would supply (Appendix~\ref{app:extension-paper-baseline}).
  \item \textbf{\ara{} agent}: reads the full \ara{} (\texttt{PAPER.md}, \texttt{logic/}, \texttt{src/}, \texttt{trace/}, \texttt{evidence/}); the \texttt{src/} and reference-derived \texttt{logic/} content match the paper agent's bundle, and the \texttt{trace/} and \texttt{evidence/} layers carry the failure record that the paper bundle has no analogue for.
\end{itemize}
Both agents are instructed to beat the reference score by editing \texttt{solution.py} and running \texttt{bash score.sh}; the result is the best score across all invocations during the run.
We use the Claude Agent SDK~\citep{claudecodesdk2025} with a tool surface of \{\texttt{Bash}, \texttt{Read}, \texttt{Edit}, \texttt{Write}, \texttt{Glob}, \texttt{Grep}\} and a 8\,h SLURM wall clock + \$50 API-spend cap per run.
All five tasks run on Claude Sonnet 4.6; for \texttt{triton\_cumsum} and \texttt{restricted\_mlm} we additionally ran the same comparison on the older Sonnet 4.5 base (Appendix~\ref{app:extension-cases}).
Harness engineering, score-event extraction, and reproducibility details are in Appendix~\ref{app:extension-eval}.

\begin{figure*}[t]
\centering
\includegraphics[width=\linewidth]{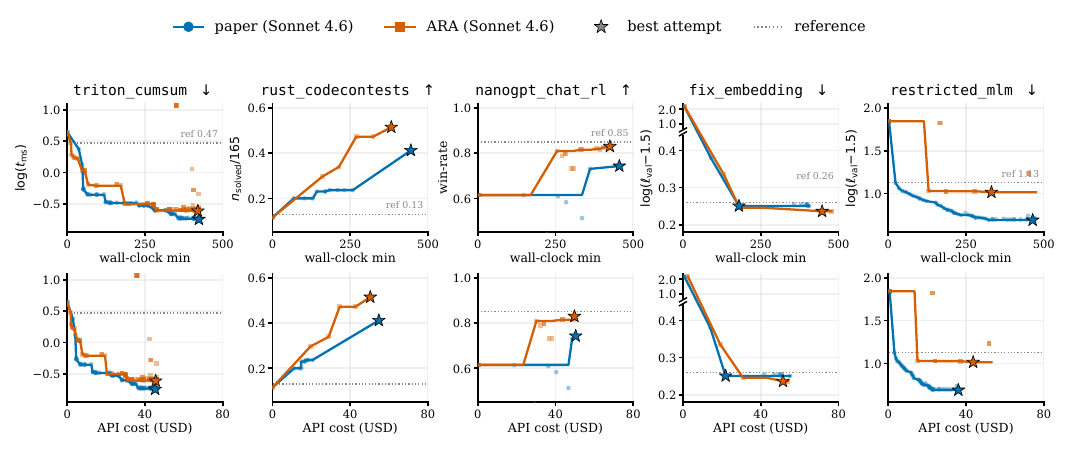}
\caption{Extension trajectories on five RE-Bench tasks under Claude Sonnet 4.6. One task per column: top row is score-vs-wall-clock-time, bottom row is score-vs-cumulative-cost; the y-axis is shared down each column. Faint markers are individual scoring attempts, solid lines are the best-so-far envelope, and stars mark the best attempt; dotted grey lines mark each task's RE-Bench reference. Arrows in the column titles indicate metric direction. \emph{Across all five tasks the \ara{} agent reaches a useful first move earlier than the paper agent, and ends with the better best score on \texttt{rust\_codecontests}, \texttt{nanogpt\_chat\_rl}, and \texttt{fix\_embedding}; on \texttt{triton\_cumsum} and \texttt{restricted\_mlm} the paper agent later overtakes via moves the trace does not name (an \texttt{int8} kernel redesign and focused depth on a single architecture, respectively). \S\ref{sec:eval-extension} unpacks why the early \ara{} lead does not hold on these two tasks and how the same comparison on the older Sonnet 4.5 base inverts the result.} Per-task case studies and Sonnet 4.5 trajectories for \texttt{triton\_cumsum} and \texttt{restricted\_mlm} are in Appendix~\ref{app:extension-cases}.}
\label{fig:extension-summary}
\end{figure*}

\paragraph{Results.}
Figure~\ref{fig:extension-summary} reports the best-so-far envelope and the underlying scoring attempts per agent on each task, against wall-clock time and API spend. On \texttt{rust\_codecontests}, \texttt{nanogpt\_chat\_rl}, and \texttt{fix\_embedding} the \ara{} agent ends with the better best score; on \texttt{triton\_cumsum} and \texttt{restricted\_mlm} under Sonnet 4.6 the paper agent ends ahead. The trajectories surface three phenomena described below; per-task case studies and trace evidence (\texttt{file reads}, \texttt{ThinkingBlock} reasoning, edit history) are in Appendix~\ref{app:extension-cases}.

\paragraph{Early acceleration across all five tasks.}
The \ara{} agent reaches a useful first move earlier on every task, including the two it eventually trails on. The clearest case is \texttt{rust\_codecontests}: it commits to a hand-coded Rust library at $t = 9$\,min after reading heuristic \texttt{H12}, while the paper agent spends six hours on prompt-engineering variants and only at $t = 395$\,min, while inspecting the workdir, notices the same idea and starts populating the existing \texttt{few\_shots/} cache. On Sonnet 4.6 \texttt{triton\_cumsum}, the \ara{} agent scores $0.47$ at $t = 11$\,min using trace-derived ideas (\texttt{decoupled lookback}, \texttt{associative\_scan}), while the paper agent does not score until $t = 37$\,min and reasons from first principles. The same head start appears on \texttt{nanogpt\_chat\_rl} (heuristic \texttt{H08} pre-names a degenerate-output filter the paper agent has to discover by debugging) and on \texttt{fix\_embedding} (heuristics \texttt{H11}/\texttt{H13} mark permutation recovery as a documented dead end, but the paper agent tries it at $t = 19$\,min, abandons it on visible failure, then re-tries it at $t = 350$\,min having forgotten its own earlier failure). Across all five tasks the recorded heuristics and prior failures shorten the path to a first useful move; the question is whether that lead carries through to the final score.

\paragraph{Late-phase reversal on Sonnet 4.6 triton and mlm.}
On these two tasks the early lead does not hold, and the reversal is itself informative. On \texttt{triton\_cumsum}, the paper agent---with no menu to commit to---keeps redesigning the kernel and at $t = 47.7$\,min introduces an \texttt{int8} input compression motivated by the scorer's 8-bit input range, a move the trace never names; it iterates from there and finishes ahead. The \ara{} agent meanwhile sticks with the trace-recommended decoupled-lookback design. On \texttt{restricted\_mlm}, the paper agent commits to a single \texttt{ConvMLMDilated} tune for the full 8\,h, while the \ara{} agent implements every heuristic-named alternative architecture (\texttt{H11} ReLU-attention, \texttt{H07} MLPMixer, etc.), finds that none beats the simpler ConvMLM under Sonnet 4.6's optimisation, and ends behind. In both cases the \ara{} agent followed the trace faithfully; the trace simply was not the most creative option available to that model.

\paragraph{A weaker base inverts the comparison.}
For both tasks, paired Sonnet 4.5 trajectories show the opposite pattern (Appendix Figs.~\ref{fig:extension-triton-45},~\ref{fig:extension-mlm-45}): the \ara{} agent reaches $0.27$ on \texttt{triton\_cumsum} vs.\ $0.64$ for the paper agent, and $0.73$ on \texttt{restricted\_mlm} vs.\ $1.03$. A weaker base lacks the bandwidth to invent moves like the \texttt{int8} compression or to commit deeply to a single architecture, and the same heuristics that constrained Sonnet 4.6 give Sonnet 4.5 a productive ranked list of strategies to try. The artifact's value appears to scale with the gap between what the trace documents and what the agent can discover on its own.

\paragraph{Outlook.}
Taken together, the trajectories suggest that an \ara{} can aid human-agent and agent-agent communication by surfacing prior pitfalls and successful strategies, but that selectively hiding or contextualising parts of the trace may matter when the agent's own bandwidth exceeds what the documented playbook records. Marking trace nodes with model-class provenance, so successors can discount claims that no longer apply, is one such mechanism; we leave the broader design space to follow-up work.

\subsection{\ara{}-Native Review Systems}
\label{sec:eval-seal}

Sections \ref{sec:eval-understanding}--\ref{sec:eval-extension} evaluate \ara{} as a \emph{format}; this subsection evaluates the \emph{review machinery} the format enables. We test whether the three-level Seal protocol (\S\ref{sec:review-system}) actually detects deficient artifacts on its three automated levels: structural integrity (Level~1), argumentative rigor (Level~2, the Rigor Auditor), and execution reproducibility (Level~3). The Stage~3 substantive-judgment layer, where human reviewers assess significance and novelty, has no automated oracle and we do not attempt to measure it.

\paragraph{Level~1 (structural integrity).}
Level~1 checks schema conformance, required-field presence, and cross-layer reference resolution. The Understanding experiment already shows the gate works in practice: every \ara{} in this paper passes Level~1 before use, and the 95.6\% Cat.~A accuracy (Table~\ref{tab:understanding}) means the gated artifacts are structurally complete enough for an agent to retrieve what the source actually contains (per-artifact iteration counts and failure-type distributions in App.~\ref{app:seal-eval-l1}).

\paragraph{Level~2 (Rigor Auditor): mutation benchmark.}
We benchmark the Rigor Auditor by injecting known errors into \ara{}s that already pass Level~1 and measuring the auditor's detection rate against the ground-truth injection manifest. Each injection is its own oracle, removing the need for human annotation. Concretely, we take the 23 PaperBench \ara{}s and seed each with five injections:
\begin{itemize}[leftmargin=*, itemsep=1pt, topsep=2pt]
  \item \emph{Fabricated claim}: a claim that cites a non-existent experiment.
  \item \emph{Missing falsification}: a primary claim with its \texttt{Falsification criteria} field removed.
  \item \emph{Orphan experiment}: an experiment whose \texttt{Verifies} field points to a non-existent claim.
  \item \emph{Over-claim}: a narrow result whose \texttt{Statement} is broadened to universal scope.
  \item \emph{Rebutted-branch leak}: a claim advocating an approach documented as \texttt{dead\_end} in the exploration tree.
\end{itemize}
The Rigor Auditor (a Claude Code SDK~\citep{claudecodesdk2025} agent that reads the artifact, builds cross-layer maps, scores six dimensions, and compiles a findings list; full protocol in App.~\ref{app:seal-eval-l2}) is then run on each injected \ara{}, and each finding is matched back to the manifest via its target entity.

\paragraph{High recall on substantive injections, a blind spot on orphans.}
Table~\ref{tab:seal-l2} reports per-type detection. The auditor catches 100\% of three high-severity classes (fabricated claims, rebutted-branch leaks, over-claims) and 91\% of missing falsifications, but only 22\% of orphan experiments. The asymmetry is interpretable: orphans require enumerating every experiment and cross-checking its \texttt{Verifies} target against the claim list, whereas the other four surface naturally inside the auditor's per-claim loop. The natural fix is to move orphan detection into Level~1 as a deterministic structural check.

\begin{table}[t]
\centering
\small
\begin{tabular}{l c r r}
\toprule
\textbf{Injection type} & \textbf{Expected severity} & \textbf{$n$} & \textbf{Detected} \\
\midrule
Fabricated claim        & Critical & 23 & 23 (100\%) \\
Rebutted-branch leak    & Critical & 23 & 23 (100\%) \\
Over-claim (scope)      & Major    & 23 & 23 (100\%) \\
Missing falsification   & Major    & 23 & 21 (91\%) \\
Orphan experiment       & Minor    & 23 &  5 (22\%) \\
\midrule
\textbf{Overall}        &          & \textbf{115} & \textbf{95 (82.6\%)} \\
\bottomrule
\end{tabular}
\caption{Rigor Auditor effectiveness on the mutation benchmark (23 \ara{}s $\times$ 5 injection types). The auditor catches all high-severity structural anomalies but exhibits a systematic blind spot on orphan experiments.}
\label{tab:seal-l2}
\end{table}

\paragraph{Two LLM-as-judge pathologies in the auditor's scoring.}
Two scoring-side biases emerge. First, \emph{grade inflation}: in 17 of 23 \ara{}s the auditor's reported overall mean is rounded up just enough to clear the Accept threshold. Second, \emph{finding--score decoupling}: even when the auditor correctly flags an injection as \emph{critical} (22 of 23 rebutted-branch-leak cases), the corresponding dimension score does not drop to the level the rubric prescribes. Both are documented LLM-as-judge failure modes~\citep{zheng2023judging}, and together they suggest LLMs should generate findings rather than grades, with the overall verdict computed deterministically from the findings list. Dimension distributions, the grade-inflation breakdown, and a per-paper detection heatmap are in App.~\ref{app:seal-eval-l2}.

\paragraph{Level~3 (execution reproducibility).}
Level~3's specification in \S\ref{sec:liveness}---read the artifact, reproduce central claims directionally under a compute budget, with numerical results masked---is exactly the \ara{}-condition protocol of the Reproduction experiment (\S\ref{sec:eval-reproduction}). The 64.4\% difficulty-weighted success rate measured there is therefore the Level-3 verification rate on well-formed artifacts, and the single result-fabrication that the blinded judge surfaced shows the verifier flags misrepresented results when they appear.

\FloatBarrier
\section{Related Work}
\label{sec:related}

Our work synthesizes ideas from three research threads---machine-readable science, reproducibility infrastructure, and agent-oriented tooling---and contributes a unified protocol that none of them provide individually.

\paragraph{The dimensional gap of existing tools.}
A natural objection is that \ara{} merely combines documentation, version control, and experiment tracking---three categories of tools researchers already use.
Table~\ref{tab:dimension-gap} shows why even using PDFs, GitHub, and trackers (MLflow~\citep{zaharia2018mlflow}, Weights~\&~Biases~\citep{biewald2020wandb}) \emph{simultaneously} leaves the knowledge siloed in three unlinked formats with no cross-referencing between claims, the code that tests them, the evidence produced, and the decisions that selected it.
\ara{} closes this gap not by replacing these tools but by providing the missing structural layer: an executable epistemic graph whose cross-layer bindings make these connections explicit and machine-traversable.

\begin{table}[t]
\centering
\small
\setlength{\tabcolsep}{4pt}
\begin{tabular}{@{}lcccc@{}}
\toprule
& \textbf{PDF} & \textbf{GitHub} & \textbf{Tracker} & \textbf{\ara{}} \\
\midrule
Structured scientific logic  & {\color{orange}$\sim$} & {\color{orange}$\sim$} & {\color{red!70!black}$\times$} & {\color{green!50!black}\checkmark} \\
Executable code              & {\color{red!70!black}$\times$} & {\color{green!50!black}\checkmark} & {\color{red!70!black}$\times$} & {\color{green!50!black}\checkmark} \\
Exploration trajectory       & {\color{red!70!black}$\times$} & {\color{red!70!black}$\times$} & {\color{orange}$\sim$} & {\color{green!50!black}\checkmark} \\
Grounded evidence            & {\color{orange}$\sim$} & {\color{red!70!black}$\times$} & {\color{orange}$\sim$} & {\color{green!50!black}\checkmark} \\
Cross-layer bindings         & {\color{red!70!black}$\times$} & {\color{red!70!black}$\times$} & {\color{red!70!black}$\times$} & {\color{green!50!black}\checkmark} \\
\bottomrule
\end{tabular}
\caption{Dimensional coverage of existing research artifacts. Each row is a requirement for agent-native research (\S\ref{sec:intro}). Existing tools cover at most two dimensions structurally; \ara{} covers all five with explicit cross-layer bindings. {\color{green!50!black}\checkmark}~=~full, {\color{orange}$\sim$}~=~partial (present but unstructured or scattered), {\color{red!70!black}$\times$}~=~absent.}
\label{tab:dimension-gap}
\end{table}

\paragraph{Machine-readable research artifacts.}
A growing line of work argues that scientific knowledge should be authored in machine-readable form during the research process rather than recovered post-hoc: the FAIR principles~\citep{wilkinson2016fair} standardize data metadata, the W3C PROV ontology~\citep{lebo2013provo} formalizes provenance for scientific outputs, \citet{canini2026stopwriting} reframes the paper as a ``compression format for human readers'' that should yield to structured knowledge objects, and \citet{stocker2025machinereadable, booeshaghi2026machinereadable} advocate authoring-time machine readability as a first principle.
Concrete formats instantiate parts of this vision---nanopublications~\citep{groth2010nanopub} atomize claims with provenance, the Open Research Knowledge Graph~\citep{jaradeh2019orkg} curates structured contributions across papers, RO-Crate~\citep{soilandreyes2022rocrate} bundles research objects, Whole Tale~\citep{brinckman2019wholetale} packages computational environments, and the Discovery Engine~\citep{baulin2025discoveryengine} distills publications into a Conceptual Tensor---but none provides execution semantics or captures decision history.
Unlike these formats, \ara{} jointly binds scientific logic, minimal executable code, and decision history into a single protocol with machine-verifiable reproducibility.

\paragraph{Reproducibility infrastructure.}
The reproducibility crisis in ML~\citep{baker2016reproducibility, pineau2021reproducibility} has motivated code-sharing standards~\citep{stodden2016enhancing}, scientific workflow engines~\citep{koster2012snakemake, ditommaso2017nextflow, crusoe2022cwl}, and computational notebooks~\citep{knuth1984literate, rule2018exploration}; yet workflows encode pipelines without claim semantics, notebooks remain documents with hidden state, and recent benchmarks~\citep{starace2025paperbench, liu2026scireasoning, kon2025expbench} collectively show that frontier agents cannot recover knowledge PDFs leave implicit---EXP-Bench reports only 0.5\% end-to-end experiment success despite 20--35\% component accuracy.
On the verification side, LLMs detect fewer than 46\% of paper--code discrepancies~\citep{baumgartner2026scicoqa}, extending a longer line of scientific claim verification~\citep{wadden2020scifact} and attribution-based grounding~\citep{gao2023rarr} and motivating formal auditing criteria across provenance, soundness, claim decomposition, and cryptographic lineage~\citep{rasheed2026aar, huang2025decmetrics, radanliev2026aibom}.
Unlike prior auditing proposals that address a single dimension, \ara{}'s Seal Certificates operationalize all of them in one enforceable mechanism.

\paragraph{Negative knowledge and failed trajectories.}
Recent work shows that failure traces become actionable only once annotated with root-cause structure~\citep{zhu2025agenterrorbench, zhang2025agentracer}, yet raw trajectory dumps~\citep{yang2024sweagent} remain difficult to leverage.
Large-scale experiment logs~\citep{pinedaarango2021hpob, ying2019nasbench, gijsbers2019automl} retain $>$99.99\% more search history than their corresponding papers report, and process-level studies~\citep{wijk2025rebench, yamada2025aiscientistv2} confirm that human experts and agentic scientists both explore extensive dead ends that never surface in the write-up.
Unlike raw trajectory archives, \ara{}'s exploration graph promotes dead ends to first-class \texttt{dead\_end} nodes with structured failure modes and claim cross-references, making negative knowledge machine-queryable rather than lost to narrative selection.

\paragraph{Agent-oriented documentation and tooling.}
A convergent body of work shows that agents benefit from structured, layered representations~\citep{openai2025agentsmd, vasilopoulos2026codifiedcontext} over flat corpora~\citep{lo2020s2orc, priem2022openalex}, and that even the strongest LLMs implement fewer than 40\% of novel contributions correctly, with \emph{semantic} misalignment as the dominant failure mode~\citep{jimenez2024swebench, chen2025scienceagentbench, hua2025researchcodebench}.
Recent systems target this gap from three sides: pipelines that convert papers into executable code~\citep{seo2025papercoder} or interactive AI agents~\citep{miao2025paper2agent} post-hoc, or recover tacit knowledge through graph analysis and debugging~\citep{li2026tacit}; knowledge-graph approaches that mine background literature for technique--code links~\citep{liu2026knowledgegraphs, luo2025xkg}---yielding up to 10.9\% PaperBench gains but leaving the target contribution's decision history and epistemic structure unmodeled; and autonomous research agents that conduct experiments end-to-end~\citep{boiko2023coscientist, bran2024chemcrow, schmidgall2025agentlab, baek2025researchagent}, whose unstructured trajectory logs are themselves discarded once the resulting paper is written.
Multi-agent frameworks~\citep{wu2024autogen}, skill-library standards~\citep{wang2023voyager, agentskills2025}, and artifact-mediated agent coordination~\citep{wang2026scienceclaw} further show that structured artifacts, not natural-language papers, are the natural unit of exchange for compounding agent capability---a premise our Live Research Manager (\S\ref{sec:live-pm}) instantiates.
Unlike post-hoc recovery pipelines and background-knowledge graphs, \ara{} encodes claims, evidence, heuristics, and their executable bindings at authoring time, eliminating the recovery step entirely.

\FloatBarrier
\section{Future Work}
\label{sec:discussion}

\paragraph{Near term: artifact lineage and self-maintaining ecosystems.}
The most pressing near-term gap is artifact durability: like code repositories, \ara{}s decay without maintenance as dependencies rot and practices evolve, yet unmaintained artifacts are the community norm.
The natural extension is a \emph{lineage mechanism} in which each \ara{} declares its parent artifacts and expresses its contribution as a structured diff, reducing both construction cost (authors specify only the delta) and verification cost (reviewers and agents re-check only the new contribution).
Lineage also enables self-maintaining ecosystems: agents consuming an \ara{} detect and repair staleness, update deprecated dependencies, and propagate corrections upstream, so that every act of consumption becomes an act of maintenance.

\paragraph{Medium term: knowledge graph, collaborative discovery, and continuous review.}
Aggregated lineages form a queryable scientific knowledge graph that lifts collaboration and review from the document level to the corpus level.
Cross-artifact claim alignment turns literature synthesis into subgraph queries, lets reviewer agents verify that reported baselines match what cited \ara{}s recorded, and exposes trajectory conflicts where a method claimed as successful elsewhere was documented as failing.
Shared Exploration Graphs also enable collaboration formats impossible in the PDF ecosystem, from parallel continuation of open problems with documented dead ends to fine-grained attribution on live, evolving artifacts.
Review evolves in parallel: there is no single accept moment, only a claim-confidence surface that rises with replications and falls with counter-evidence, freeing human expert attention for the judgments only humans make, namely novelty, significance, and taste.

\paragraph{Long term: cross-disciplinary collective memory.}
Our evaluation is restricted to machine learning, where \ara{}'s four-layer structure aligns naturally with the dominant contribution types: algorithms, architectures, and training procedures.
Whether this structure generalizes to other disciplines remains an open question.
The Cognitive and Evidence Layers are plausibly domain-agnostic, but the Physical Layer and Exploration Graph, both premised on iterable computational experiments, may require substantial adaptation for wet-lab sciences where execution is physical rather than computational.
If these adaptations succeed, \ara{} provides a natural substrate for cross-disciplinary knowledge transfer, where documented failures in one field become actionable knowledge in another via graph traversal rather than literature search in unfamiliar notation.

\FloatBarrier
\section{Limitations}
\label{sec:limitations}

Three limitations bound the claims in this paper.

\textbf{Evaluation scope.} Our study covers only machine learning papers, where computational reproducibility and well-defined contribution types make \ara{}'s four-layer structure a natural fit; whether the protocol generalizes to experimental sciences with physical execution requirements, or to theoretical disciplines where the Physical Layer is largely absent, remains empirically untested. Extending the Physical Layer to formal or proof-based results requiring machine-checkable specifications is a natural direction for future work. Our human-annotated benchmark was also constructed by annotators familiar with both the \ara{} format and the selected papers, so performance on unfamiliar or niche-domain artifacts may differ from the reported figures.

\textbf{Fidelity ceiling.} \ara{} fidelity is bounded by the source of supervision. The Compiler faithfully represents only what the PDF contains (\S\ref{sec:ingestor}): when a paper omits experimental details, environment specifications, or ablation results, no extraction method can recover them. The Live Research Manager closes this gap by recording trajectories as research unfolds, but assumes an AI-native workflow in which a coding agent is present throughout the project. For researchers outside such sessions, the Compiler still produces a valid \ara{} from the finished paper, but the resulting artifact inherits the PDF's omissions; hand-authoring structured fields remains possible but reintroduces the documentation burden the protocol aims to eliminate. Closing this adoption gap tracks the broader diffusion of agent workflows in research practice, not the protocol itself.

\textbf{Deployment prerequisites.} Two properties required for production use are not yet implemented. The adversarial robustness and privacy guarantees raised in \S\ref{sec:liveness} are aspirational: the current system lacks sandboxed execution, content-level anomaly detection, and granular access control for the Exploration Graph. Separately, any long-lived format faces \emph{schema evolution}: as research practice changes, the \ara{} schema will need to add node types, refine field semantics, and deprecate conventions without breaking prior artifacts. We version \texttt{PAPER.md} frontmatter with an explicit \texttt{ara\_schema} tag and require all validators to accept unknown fields (forward compatibility) and degrade gracefully on missing optional fields (backward compatibility), but have only exercised this discipline across minor revisions. A stable migration story for major revisions, including automatic rewriting of archival artifacts, long-term checker availability, and a deprecation policy, remains future work.

\FloatBarrier
\section{Conclusion}
\label{sec:conclusion}

We introduce the \ara{} protocol and its surrounding ecosystem as a foundation for agent-native scientific communication.
Together, they address two structural failures of the PDF format: knowledge that narrative conventions discard (failed attempts, implicit configurations, unexplored branches) and specifications too underspecified to execute.
\ara{} resolves both by restructuring a research contribution as a machine-actionable artifact, one that is navigable, complete, and verifiable without human interpretation.

The broader motivation is a shift already underway: AI agents are becoming first-class participants in research workflows, not tools that assist humans but autonomous contributors that read, reproduce, and extend scientific work.
That transition demands infrastructure built around agents from the start.
\ara{} is the core abstraction of that ecosystem, a common substrate through which human and machine researchers alike publish, verify, and build on scientific knowledge.

\bibliography{references}
\bibliographystyle{styles/icml2026}

\appendix

\section{ARA Protocol and Design Rationale}
\label{app:protocol-details}

This appendix consolidates the full protocol specification, design rationale, and validation details for the \ara{} format.

\subsection{Taxonomy of Reproduction-Critical Information}
\label{app:taxonomy}

To understand \emph{what} information an agent needs to reproduce a paper---and where PDFs fall short---we analyze the expert-authored rubrics from PaperBench~\citep{starace2025paperbench}, a benchmark that evaluates AI agents on full paper reproduction.
Each rubric decomposes a paper into atomic \emph{leaf requirements}: individually verifiable conditions that collectively constitute a faithful reproduction.
\textbf{Scope.} The taxonomy below is derived from a deeply annotated 5-paper subset (3{,}050 leaves), chosen for tractability of fine-grained category labeling; the coarser per-task-category and gap-type frequencies reported in \S\ref{sec:eval-understanding} and Appendix~\ref{app:gap-types} are validated on the full 23-paper corpus (8{,}921 requirements). The subset spans diverse domains---black-box LLM adaptation (BBox), mechanistic interpretability, continual RL (Self-Composing Policies), physics-informed neural networks (PINN), and foundation models for RL (FRE).

By categorizing every leaf into a taxonomy of information types, we reveal both the \emph{diversity} of knowledge needed for reproduction and the specific failure modes that arise when this knowledge is scattered across a narrative PDF rather than organized in a structured artifact.

\subsubsection{Information Categories}

We identify ten categories of reproduction-critical information.
Table~\ref{tab:taxonomy} summarizes the categories with their frequency distribution across our analyzed rubrics.

\begin{table*}[h]
\centering
\small
\begin{tabular}{p{3.2cm}rp{3.6cm}l}
\toprule
\textbf{Category} & \textbf{\%} & \textbf{PDF Difficulty} & \textbf{\ara{} Layer} \\
\midrule
Combinatorial experiment matrix & 24.1 & Implicit in prose & \texttt{experiments.md} \\
Evaluation protocol & 18.5 & Scattered across \S, appendix & \texttt{experiments.md} \\
Hyperparameters & 17.2 & Buried in appendix tables & \texttt{configs/training.md} \\
Metric logging & 10.4 & Rarely specified & \texttt{experiments.md} \\
Result interpretation & 8.6 & Mixed with discussion & \texttt{claims.md}, \texttt{evidence/} \\
Architecture specification & 5.8 & Split across text, figures, appendix & \texttt{architecture.md} \\
Mathematical formulation & 4.5 & Equation references break across sections & \texttt{algorithm.md} \\
Implementation tricks & 4.2 & Footnotes, appendix asides & \texttt{heuristics.md} \\
Data pipeline & 3.8 & Preprocessing details omitted & \texttt{configs/}, \texttt{environment.md} \\
Environment / infrastructure & 2.9 & Assumed known & \texttt{environment.md} \\
\bottomrule
\end{tabular}
\caption{Taxonomy of reproduction-critical information in PaperBench rubrics. Frequency is computed across 3{,}050 leaf requirements from five papers. The ``PDF Difficulty'' column characterizes the primary challenge of extracting this information from a narrative PDF. The ``\ara{} Layer'' column identifies which \ara{} component directly addresses each category.}
\label{tab:taxonomy}
\end{table*}

Below we define each category, give concrete examples drawn from the PaperBench rubrics, and explain how \ara{}'s structure addresses the underlying retrieval challenge.

\paragraph{1. Combinatorial experiment matrix (24.1\%).}
The single largest category consists of requirements that enumerate which model variant must be run on which dataset, with which configuration, for how many seeds.
In PDFs, this combinatorial structure is compressed into a single sentence (``We evaluate all methods on three task sequences with 10 seeds each'') or a results table whose row/column headers implicitly define the cross-product.
An agent must mentally decompose the matrix to know, e.g., that ``CompoNet on Freeway, 10 seeds, 1M timesteps per task'' is a distinct run.

\emph{Examples:}
\begin{itemize}[leftmargin=*, itemsep=1pt]
  \item \emph{self-composing-policies}: 62 leaves enumerate \{6 methods\} $\times$ \{3 task sequences\} $\times$ \{seeds, timesteps, trained\}---each a separate verifiable requirement (e.g., ``CompoNet on Meta-World: 10 seeds, 1M timesteps/task, trained'').
  \item \emph{bbox}: 46 evaluation leaves cross \{5 model sizes\} $\times$ \{4 datasets\} $\times$ \{3 feedback types\} $\times$ \{single-step, full-step\} inference modes.
  \item \emph{pinn}: $\sim$1{,}273 leaves enumerate a grid of \{4 PDE problems\} $\times$ \{4 network widths\} $\times$ \{5 learning rates\} $\times$ \{3 optimizers\}, each combination a distinct training run.
\end{itemize}

\noindent\textbf{\ara{} advantage.}
The \texttt{experiments.md} file makes every cell of the experiment matrix explicit, with structured \texttt{Setup} fields that list model, dataset, and configuration as machine-readable key--value pairs.
An agent can enumerate all runs programmatically rather than parsing table headers.

\paragraph{2. Evaluation protocol (18.5\%).}
Requirements specifying \emph{which} metric to compute, on \emph{which} test split, using \emph{which} evaluation-time configuration (e.g., beam size, number of evaluation episodes, specific layers to probe).
These details are often scattered: the metric definition appears in \S3, the test split in \S4, the evaluation episodes in the appendix, and the layer indices in a figure caption.

\emph{Examples:}
\begin{itemize}[leftmargin=*, itemsep=1pt]
  \item \emph{mechanistic-understanding}: ``Compute cosine similarity between $\delta_i$ and $\delta_{\text{mlp},i}$ for layers 0, 2, 4, 6, 8, 10, 12, 14, 16, 18 using 1{,}199 prompts from RealToxicityPrompts.''
  \item \emph{fre}: ``Evaluation is repeated and averaged over 20 episodes and 5 seeds; 32 state-reward pairs are sampled from the evaluation task environment.''
  \item \emph{bbox}: ``The Chain-of-Thought baseline has been evaluated on the test splits of all datasets using GPT-3.5 Turbo.''
\end{itemize}

\noindent\textbf{\ara{} advantage.}
Each experiment entry in \texttt{experiments.md} has a declarative \texttt{Procedure} field that specifies evaluation steps as an ordered list, and a \texttt{Metrics} field that names the exact metrics.
The \texttt{configs/training.md} file separately records evaluation-time parameters (e.g., beam size, episodes).

\paragraph{3. Hyperparameters (17.2\%).}
Classic training configuration: learning rates, batch sizes, optimizer parameters, temperature, weight decay, LoRA rank, number of epochs.
While these are the \emph{most commonly discussed} reproduction barrier, they account for only 17\% of leaf requirements.
In PDFs, hyperparameters are typically consolidated in an appendix table, but the correspondence between table rows and specific experimental conditions is often ambiguous.

\emph{Examples:}
\begin{itemize}[leftmargin=*, itemsep=1pt]
  \item \emph{bbox}: ``AdamW optimizer with learning rate 5e-6, weight decay 0.01; batch size 64; 6{,}000 training steps.''
  \item \emph{self-composing-policies}: 29 leaves enumerate every SAC and PPO parameter individually (e.g., ``SAC: target smoothing coefficient $\tau = 0.005$''; ``PPO: GAE $\lambda = 0.95$'').
  \item \emph{pinn}: ``Learning rate of the Adam optimizer can be set to 1E-5, 1E-4, 1E-3, 1E-2, or 1E-1.''
\end{itemize}

\noindent\textbf{\ara{} advantage.}
The \texttt{configs/training.md} file provides a single, authoritative location for all hyperparameters, organized by experiment.
The \texttt{heuristics.md} file additionally records \emph{sensitivity} annotations (low/medium/high) and valid \emph{bounds}, information that PDFs almost never provide.

\paragraph{4. Metric computation and logging (10.4\%).}
Requirements that the agent must \emph{record} specific intermediate quantities during runs: loss curves, attention distributions, cost tracking (dollars per 1k questions), episodic returns logged every $N$ steps.
This ``instrumentation'' knowledge is rarely specified in papers---authors implicitly know what to log but do not document it as part of the method.

\emph{Examples:}
\begin{itemize}[leftmargin=*, itemsep=1pt]
  \item \emph{bbox}: 71 leaves (25\% of the paper's rubric) require computing and saving training costs, inference costs (USD/1k questions), and evaluation costs across all dataset $\times$ variant combinations.
  \item \emph{self-composing-policies}: ``Output attention distribution logged every 10k timesteps''; ``Matching rate between final output and internal policy, saved every 10k steps.''
\end{itemize}

\noindent\textbf{\ara{} advantage.}
The \texttt{experiments.md} \texttt{Metrics} and \texttt{Procedure} fields can explicitly list what to log and at what frequency.
The \texttt{evidence/} layer provides concrete examples of the expected output format.

\paragraph{5. Result interpretation (8.6\%).}
Qualitative claims about what the results should \emph{show}---directional trends, comparative rankings, mechanistic explanations.
These carry the highest weight in PaperBench rubrics (weight = 2) because they test whether the agent \emph{understands} the results, not just whether the code ran.

\emph{Examples:}
\begin{itemize}[leftmargin=*, itemsep=1pt]
  \item \emph{mechanistic-understanding}: ``After adapting with DPO, the principal component of the residual streams shift in the same direction, and the activation of the toxic vectors decrease.'' (Weight = 2)
  \item \emph{mechanistic-understanding}: ``The extracted tokens encode different characteristics of toxic language: tokens from $\mathbf{W}$ are mostly curse words; MLP.vToxic are a mix of curse words and insults; SVD.uToxic encode insults and female sexual references.'' (Weight = 2)
  \item \emph{self-composing-policies}: ``CompoNet achieves higher average performance and forward transfer than all baselines on all three task sequences.''
  \item \emph{pinn}: ``Adam+L-BFGS always achieves the lowest minimum loss compared to just using Adam or L-BFGS alone.''
\end{itemize}

\noindent\textbf{\ara{} advantage.}
The \texttt{claims.md} file states each claim with explicit \texttt{Falsification criteria} and pointers to the experiment that verifies it.
The \texttt{experiments.md} \texttt{Expected outcome} field records the directional prediction (e.g., ``method A outperforms method B'') without revealing exact numbers, enabling blind reproduction.

\paragraph{6. Architecture specification (5.8\%).}
Layer counts, channel sizes, activation functions, output head structure, embedding dimensions.
In PDFs, architecture details are split across a figure (showing the high-level diagram), the methods section (describing components in prose), and the appendix (listing dimensions in a table).
An agent must mentally compose these three sources to build the full specification.

\emph{Examples:}
\begin{itemize}[leftmargin=*, itemsep=1pt]
  \item \emph{self-composing-policies}: ``CNN has three convolutional layers with 32, 64, and 64 channels and filter sizes of 8, 4, and 3''; ``SAC: hidden dimension $d_{\text{model}} = 256$; critic network has 3 layers; activation is ReLU.''
  \item \emph{fre}: ``GC-BC model is a MLP with three hidden layers of size 512''; ``layer normalization is applied before each activation function.''
  \item \emph{mechanistic-understanding}: ``Binary classifier of the form $\text{softmax}(\mathbf{Wx})$ where $\mathbf{W}$ has dimensionality $K \times 2$.''
\end{itemize}

\noindent\textbf{\ara{} advantage.}
The \texttt{architecture.md} file provides a single location listing every component with its dimensions, activation functions, and input/output specifications.
Code stubs in \texttt{src/code/} provide an executable complement.

\paragraph{7. Mathematical formulation (4.5\%).}
Specific equations that must be implemented exactly: loss functions, attention operations, PDE boundary conditions, update rules.
In PDFs, equations are referenced by number, but the reader must trace through variable definitions scattered across multiple sections.

\emph{Examples:}
\begin{itemize}[leftmargin=*, itemsep=1pt]
  \item \emph{self-composing-policies}: ``Output attention: $\text{softmax}(qK^T / \sqrt{d_{\text{model}}}) \cdot V$''; ``Forward transfer: $\text{FTr}_i = (\text{AUC}_i - \text{AUC}_i^b) / (1 - \text{AUC}_i^b)$.''
  \item \emph{pinn}: ``The loss function corresponds to the non-linear least squares problem described in Section~2.1, with the relevant differential operator and boundary/initial condition operators outlined in Appendix~A.1.''
  \item \emph{fre}: ``The value function is updated with an expectile regression objective on the critic's Q-values''; ``The actor is updated via advantage-weighted regression (AWR).''
\end{itemize}

\noindent\textbf{\ara{} advantage.}
The \texttt{algorithm.md} file presents the algorithm as a self-contained pseudocode block with all variable definitions in scope.
The \texttt{concepts.md} file defines notation and links to the equations that use each symbol.

\paragraph{8. Implementation tricks (4.2\%).}
Non-obvious design choices that distinguish faithful reproduction from naive re-implementation: weight freezing schedules, initialization from prior checkpoints, gradient clipping thresholds, normalization details, optimizer switching strategies.
These are the hardest items to recover from a PDF because they appear as parenthetical remarks, footnotes, or single sentences buried in dense paragraphs.

\emph{Examples:}
\begin{itemize}[leftmargin=*, itemsep=1pt]
  \item \emph{self-composing-policies}: ``Single CNN encoder per policy; new encoder initialized with weights of the previous one'' (Appendix~E.2); ``Reset critic network at the beginning of each task''; ``Normalize summed vectors for continuous action spaces.''
  \item \emph{fre}: ``The transformer does not use a causal mask on its attention''; ``Positional embeddings are not used in the transformer''; ``States sampled for decoding and encoding are sampled separately.''
  \item \emph{pinn}: ``At the end of training, the L-BFGS directions, steps, and inverse of inner products are saved'' (Appendix~C.2); ``Strong Wolfe line search is used with L-BFGS.''
\end{itemize}

\noindent\textbf{\ara{} advantage.}
The \texttt{heuristics.md} file is \emph{specifically designed} to capture these items.
Each heuristic entry includes \texttt{Rationale} (why it matters), \texttt{Sensitivity} (how much performance degrades without it), and \texttt{Code ref} (where in the code to apply it).

\paragraph{9. Data pipeline (3.8\%).}
Dataset acquisition, split ratios, filtering criteria, preprocessing steps, data augmentation, collocation point sampling strategies.
These details are often under-specified in papers (``we use the standard train/test split'') or tucked into a single appendix paragraph.

\emph{Examples:}
\begin{itemize}[leftmargin=*, itemsep=1pt]
  \item \emph{bbox}: ``Split GSM8K into 7{,}473 training and 1{,}319 test samples''; ``Randomly sample 100 questions for TruthfulQA test set, remaining 717 for training.''
  \item \emph{mechanistic-understanding}: ``24{,}576 pairs of toxic and non-toxic continuations have been created''; ``295 prompts selected from RealToxicityPrompts that output `shit' as the next token.''
  \item \emph{pinn}: ``10{,}000 residual points randomly sampled from a 255$\times$100 grid; 257 equally spaced points for each initial condition and 101 for each boundary condition.''
\end{itemize}

\noindent\textbf{\ara{} advantage.}
The \texttt{configs/} directory provides structured configuration files with exact split sizes and sampling parameters.
The \texttt{environment.md} file specifies dataset versions and download URLs.

\paragraph{10. Environment and infrastructure (2.9\%).}
Specific API endpoints, model version strings, library versions, simulator names, hardware requirements.
These are often assumed to be ``obvious'' and omitted entirely from the paper, yet they are essential for reproduction.

\emph{Examples:}
\begin{itemize}[leftmargin=*, itemsep=1pt]
  \item \emph{bbox}: ``API access configured for davinci-002''; ``Code to execute fine-tuning jobs through the Azure OpenAI API''; ``Mixtral-8x7B-v0.1 loaded from HuggingFace in half-precision.''
  \item \emph{self-composing-policies}: 15 leaves enumerate specific Gymnasium environment IDs (e.g., \texttt{ALE/SpaceInvaders-v5}, \texttt{hammer-v2}) and required packages (\texttt{Metaworld} from Farama-Foundation).
  \item \emph{fre}: ``The observation space's XY coordinates are discretized into 32 bins for Ant Maze agents.''
\end{itemize}

\noindent\textbf{\ara{} advantage.}
The \texttt{environment.md} file lists exact package versions, model identifiers, and hardware requirements.
The \texttt{configs/model.md} file records model names, sizes, and loading configurations (e.g., precision, quantization).

\subsubsection{Key Findings}

Three observations emerge from this analysis:

\paragraph{Hyperparameters are necessary but not sufficient.}
Classic hyperparameters---the most discussed reproduction barrier---account for only 17.2\% of leaf requirements.
The remaining 82.8\% comprise evaluation protocols, experiment matrices, logging requirements, result interpretation targets, and implementation tricks that are harder to locate in a PDF and receive less attention in reproducibility discussions.

\paragraph{The combinatorial explosion is the dominant challenge.}
The largest category (24.1\%) consists of requirements that enumerate the full cross-product of models, datasets, and configurations.
In a PDF, this matrix is compressed into a single table or sentence; an agent must decompose it into individual runs.
The \ara{} format makes this matrix explicit and machine-enumerable.

\paragraph{High-weight requirements demand \emph{understanding}, not just extraction.}
All weight-2 requirements in the PaperBench rubrics belong to the ``Result Interpretation'' category.
These test whether the agent can verify that reproduced results exhibit the \emph{qualitative patterns} claimed by the paper---not just whether the code runs.
The \ara{} \texttt{claims.md} and \texttt{experiments.md} layers directly encode these verification targets, making the connection between code output and paper claims explicit rather than requiring the agent to re-derive it from narrative text.

\subsection{Physical Layer Modes: Kernel vs Repository}
\label{app:src-modes}

The Physical Layer (\texttt{/src}) adopts one of two modes, declared in the \texttt{PAPER.md} frontmatter (\texttt{src\_mode: kernel\,$|$\,repo}) so that consuming agents adapt their strategy immediately.
These two modes cover the dominant contribution types in empirical CS, where executable code is the natural physical representation.

\paragraph{Kernel mode (\texttt{/src/kernel/}).}
When the contribution is primarily \emph{algorithmic}, the invariant can be cleanly separated from scaffolding.
The kernel contains only the core modules with typed I/O signatures---often one to two orders of magnitude smaller than the full repository---stripped of all environment-specific code.
A general-purpose coding agent consumes the kernel alongside the structured specification in \texttt{/logic/solution/} and generates fresh, environment-native boilerplate in minutes.
Because agent coding capabilities improve continuously, the same kernel yields a \emph{better} surrounding implementation over time: the artifact appreciates rather than decays.

\paragraph{Repository mode (\texttt{/src/repo/}).}
When the contribution is primarily \emph{systemic}---a CUDA kernel, a distributed training strategy, a systems architecture---the engineering \emph{is} the contribution.
The full implementation is retained but \emph{annotated}: an \texttt{index.md} manifest maps each source file to the \ara{} component it implements---which claim it supports, which heuristic it embodies, which architectural module it belongs to---providing the structured navigation that a monolithic codebase lacks.
Forensic bindings connect code regions to claims, constraints, and heuristics, so an agent traverses the codebase guided by research structure rather than by directory conventions alone.

In both modes, the Cognitive Layer remains the primary interface for understanding the contribution; the Physical Layer provides executable evidence, scaled to match.

\subsection{ARA by Example: This Paper's Own Artifact}
\label{app:schema}

This paper is itself maintained as an \ara{} artifact.
The \texttt{ara/} directory at the repository root contains the living cognitive, physical, and exploration layers that were populated incrementally during the research process (see \S\ref{sec:live-pm}).
We reproduce excerpts from each key file below to give readers a concrete sense of the format.
All entries are real; only trailing items are elided for space.

\subsubsection{Directory layout and root manifest}

The complete \texttt{ara/} directory for this paper is shown below.
An agent's first action is to read \texttt{PAPER.md}, which contains YAML frontmatter and an abstract (${\sim}$500 tokens) sufficient to decide relevance without loading any layer.

\begin{lstlisting}[basicstyle=\ttfamily\scriptsize, language=]
ara/
  PAPER.md                          # entry point
  logic/                            # Cognitive Layer
    problem.md                      #   observations, gaps, key insight
    claims.md                       #   16 falsifiable claims + status
    experiments.md                  #   verification plan (E1-E6)
    related_work.md                 #   typed citation dependencies
    solution/
      heuristics.md                 #   23 design decisions + rationale
  trace/                            # Exploration Graph
    exploration_tree.yaml           #   114-node decision DAG
    sessions/                       #   38 session logs (2026-03-12..04-26)
      session_index.yaml            #   chronological index
      2026-03-12_001.yaml           #   ...one file per session
    pm_reasoning_log.yaml           #   Live PM reasoning trace
  evidence/                         # Evidence Layer
    README.md                       #   index of raw results
  staging/
    observations.yaml               #   94 unpromoted observations
\end{lstlisting}

\noindent The root manifest \texttt{PAPER.md}:

\begin{lstlisting}[basicstyle=\ttfamily\scriptsize]
---
title: "Agent-Native Research Artifacts"
authors: ["Amber Liu", "Zechen Zhang"]
venue: "NeurIPS 2026"
status: draft
date_created: "2026-03-12"
last_updated: "2026-04-27"
abstract: >
  We propose the Agent-Native Research Artifact
  (ARA), a file-system protocol that replaces the
  narrative paper with a machine-executable research
  package organized across four interlocking layers:
  a Cognitive Layer (/logic) encoding structured
  scientific reasoning, a Physical Layer (/src)
  containing the executable code kernel, an
  Exploration Graph (/trace) preserving the full
  branching research trajectory including dead ends,
  and an Evidence Layer (/evidence) grounding every
  claim in raw empirical results. PDF publication
  imposes two structural costs on autonomous
  research: a Storytelling Tax (failed experiments
  and rejected hypotheses are discarded to fit a
  linear narrative) and an Engineering Tax (the gap
  between reviewer-sufficient prose and
  agent-sufficient specification leaves critical
  implementation details unwritten). On PaperBench
  and RE-Bench, ARA raises question-answering
  accuracy from 72.4% to 93.7% and reproduction
  success from 57.4% to 64.4%; on RE-Bench's five
  open-ended extension tasks, the failure traces
  preserved in ARA accelerate research progress by
  helping the agent avoid pitfalls prior runs
  already mapped, but for a sufficiently capable
  model the same recorded playbook can constrain a
  more creative agent that would otherwise step
  outside the prior-run box.
layers:
  logic: logic/
  src: src/
  trace: trace/
  evidence: evidence/
  staging: staging/
---

# Layer Index

- **Cognitive** (`logic/`): structured reasoning
  - `problem.md` -- observations, gaps, key insight
  - `claims.md` -- 16 falsifiable claims with status
      and proof pointers
  - `experiments.md` -- verification plan (E1-E6)
  - `related_work.md` -- typed citation dependency graph
  - `solution/heuristics.md` -- 23 design decisions
      with rationale and sensitivity
- **Exploration** (`trace/`): branching trajectory
  - `exploration_tree.yaml` -- 114-node decision DAG
  - `sessions/` -- 38 session logs (2026-03-12..04-26)
  - `pm_reasoning_log.yaml` -- Live PM reasoning trace
- **Evidence** (`evidence/`): raw empirical results
  - `README.md` -- index of all evaluation data,
      including post-paper RE-Bench extension evals
- **Staging** (`staging/`): unpromoted observations
  - `observations.yaml` -- 94 preliminary observations
      (latest: 2026-04-26 cross-model and synthesis)
\end{lstlisting}

\subsubsection{Cognitive Layer: \texttt{logic/claims.md}}

Each claim carries a machine-readable status (\texttt{hypothesis}, \texttt{supported}, \texttt{testing}), falsification criteria, and proof pointers that reference the evidence layer rather than inlining results.
We show two claims at different lifecycle stages.

\begin{lstlisting}[basicstyle=\ttfamily\scriptsize]
# Claims

## C04: Universal Ingestor produces lossless
       transformations
- **Statement**: The LLM-based Ingestor faithfully
    transforms PDF papers into ARA format without
    information loss, achieving near-parity on
    factual Q&A between ARA and source PDF.
- **Status**: supported
- **Provenance**: ai-executed
- **Falsification criteria**: Systematic accuracy
    drop (>5%) on understanding questions.
- **Proof**: [evidence/README.md ->
    understanding_eval; 450 Qs across Cat A/B/C]
- **Dependencies**: [C03]
- **Tags**: ingestor, fidelity

## C06: Negative knowledge is the highest-value
       signal
- **Statement**: The Exploration Graph's dead-end
    documentation produces the largest accuracy gap
    in the entire evaluation -- agents with failure
    traces answer questions about failed approaches
    that narrative formats make structurally
    unanswerable.
- **Status**: supported
- **Provenance**: ai-executed
- **Falsification criteria**: Dead-end docs produce
    no measurable improvement on Cat C; gap <10pp.
- **Proof**: [evidence/README.md ->
    understanding_eval Cat C;
    evidence/README.md -> extension_eval]
- **Dependencies**: [C05]
- **Tags**: exploration-graph, negative-knowledge
...                    <!-- 16 claims total -->
\end{lstlisting}

\subsubsection{Cognitive Layer: \texttt{logic/problem.md}}

The problem file decomposes the motivation into typed observations (empirical facts), gaps (what existing approaches miss), and a key insight that bridges them.
Each entry carries evidence pointers and implication fields, so an agent can trace the full argumentative chain without reading the paper's introduction.
We show one representative entry per section.

\begin{lstlisting}[basicstyle=\ttfamily\scriptsize]
# Problem

## Observations
### O3: Frontier LLMs fail on research implementation
- **Statement**: Even the strongest frontier LLMs
    correctly implement fewer than 40% of novel
    research contributions when given the full paper
    and codebase, with semantic misalignment as the
    dominant failure mode.
- **Evidence**: ResearchCodeBench (Hua et al. 2025)
- **Implication**: The information encoding in PDFs
    is structurally inadequate for agent consumption.

### O4: PDF information gap is systematic
- **Statement**: Across 23 PaperBench papers (8,921
    rubric requirements), only 45.4% of reproduction
    requirements are fully specified in the PDF.
- **Evidence**: Own experiment -- info_gap_aggregate
- **Implication**: The PDF format is structurally
    incapable of serving as a self-contained
    reproduction specification.
...                    <!-- 4 observations total -->

## Gaps
### G2: Negative knowledge is systematically discarded
- **Statement**: Dead ends, rejected hypotheses, and
    convergence-critical tricks are lost to the
    narrative compression of the publication process.
- **Caused by**: O1
- **Why it matters**: Downstream agents waste compute
    re-exploring paths already proven fruitless.
...                    <!-- 2 gaps total -->

## Key Insight
- **Insight**: Separate research knowledge into four
    orthogonal layers -- structured scientific logic
    (Cognitive Layer /logic), minimal executable code
    (Physical Layer /src), preserved decision history
    (Exploration Graph /trace), and raw empirical
    results (Evidence Layer /evidence) -- to create
    a machine-executable knowledge package that
    eliminates both the Storytelling Tax and
    Engineering Tax.
- **Derived from**: O1, O2, O3, O4, G1, G2
\end{lstlisting}

\subsubsection{Cognitive Layer: \texttt{logic/solution/heuristics.md}}

Each heuristic records a design decision with its rationale, provenance (who introduced it), and sensitivity rating so that agents know which choices are safe to vary.

\begin{lstlisting}[basicstyle=\ttfamily\scriptsize]
# Heuristics

## H04: Directional verification over exact matching
- **Rationale**: Legacy papers routinely omit details
    needed for exact reproduction. Verifying
    directional properties (A > B on metric X)
    demonstrates the code kernel captures the core
    algorithmic insight without requiring exact
    numerical matches.
- **Provenance**: user
- **Sensitivity**: medium
- **Code ref**: [paper/sections/protocol.tex]

## H12: Minimal kernel = algorithm notes with inline
       snippets, not raw code files
- **Rationale**: Full code dumps (200-700 lines)
    cause context dilution -- the agent spends
    tokens parsing boilerplate already described in
    official_solution_notes.md. Notes contain core
    algorithm with key code snippets inline,
    sufficient for comprehension while 5-10x smaller.
- **Provenance**: user-revised
- **Sensitivity**: high
- **Code ref**: [code/artifacts/rebench-*/src/
    kernel/official_solution_notes.md]
...                    <!-- 18 heuristics total -->
\end{lstlisting}

\subsubsection{Exploration Graph: \texttt{trace/exploration\_tree.yaml}}

The exploration tree preserves decisions, dead ends, and experiments as a traversable graph.
We show one node of each type from this paper's 94-node tree.

\begin{lstlisting}[basicstyle=\ttfamily\scriptsize, language=]
tree:
  - id: N04
    type: decision
    title: "Tripartite layer architecture"
    provenance: user
    timestamp: "2026-03-09"
    choice: >
      Three orthogonal layers: Cognitive (/logic),
      Physical (/src), Exploration Graph (/trace).
      Each addresses a distinct tax.
    alternatives:
      - "Two-layer (logic + code only)"
      - "Four-layer (separate evidence at top)"
      - "Flat structured document (single file)"
    evidence: >
      Three-layer separation achieves minimal
      representation while preserving all three
      dimensions that PDFs conflate.

  - id: N50
    type: dead_end
    title: "Trimming src/ alone does not recover
           Cat C from enrichment regression"
    provenance: ai-suggested
    timestamp: "2026-03-14"
    hypothesis: >
      Removing boilerplate code from src/ would
      reduce context dilution enough for Cat C
      (failure knowledge) to recover to ~80%.
    failure_mode: >
      Cat C stayed at 57.5% despite reducing src/
      from 56-104K to 20-36K per artifact. Remaining
      enrichment additions still dilute
      trace/exploration_tree.yaml content.
    lesson: >
      Context dilution for Cat C is more sensitive
      than expected. Even ~200 lines of structured
      markdown in src/ can push failure knowledge
      below the retrieval threshold.

  - id: N17
    type: experiment
    title: "24,008-run exploration waste analysis"
    provenance: ai-executed
    timestamp: "2026-03-12"
    result: >
      Analyzed 24,008 runs across 21 models,
      228 tasks. 59.2% of tokens wasted on
      dead-end exploration. 90.2% of cost goes
      to failed runs. Failed runs consume 113x
      more tokens than successful ones (median).
    evidence: [C13, C14, "code/eval/malt_analysis/
      exploration_tax_findings.json"]
...                    <!-- 94 nodes total -->
\end{lstlisting}

\subsubsection{Exploration Graph: \texttt{trace/sessions/}}

Each research session is logged as a structured YAML record capturing events, AI actions, files changed, claims affected, and open threads.
Below is one session (abridged); the \texttt{session\_index.yaml} file provides a one-line chronological summary of all 36 sessions.

\begin{lstlisting}[basicstyle=\ttfamily\scriptsize, language=]
# trace/sessions/2026-03-19_001.yaml
session:
  id: "2026-03-19_001"
  timestamp: "2026-03-19T02:00"
  summary: "BAM reproduction pilot: ARA 88.2% vs
    baseline 93.2% -- first paper where baseline leads"

events_logged:
  - type: experiment
    id: N65
    summary: "BAM ARA mega-task: 220.5/250 (88.2%
      weighted), 10/10 subtasks, 6.9h, 12.6M tokens"
  - type: experiment
    id: N66
    summary: "BAM baseline mega-task: 234/251 (93.2%
      weighted), 10/10 subtasks, 4.3h, 8.6M tokens"
  - type: observation
    id: O50
    summary: "ARA won T6 (non-Gaussian) 10/10 vs
      8.5/10 -- clearer hyperparameter specs"

ai_actions:
  - action: "Ran BAM ARA reproduction (10 subtasks)"
    files_changed:
      - "code/eval/reproduction/results/bam/"

claims_touched: [C05]

open_threads:
  - "Hyperparams buried in evidence/ not src/configs/
     -- move them and rerun (becomes N67)"
\end{lstlisting}

\begin{lstlisting}[basicstyle=\ttfamily\scriptsize, language=]
# trace/sessions/session_index.yaml (excerpt)
sessions:
  - id: "2026-03-12_004"
    summary: "Full 23-paper info gap analysis -- 8,921
      reqs, median 45.3% sufficient"
  - id: "2026-03-17_001"
    summary: "Pilot reproduction: ARA 74% vs baseline
      ~14%, 5x advantage on neural-score-estimation"
  - id: "2026-03-22_001"
    summary: "Self-expansion v2 (real code): 10/10
      subtasks -- v1 pseudocode got 1/10"
  - id: "2026-03-26_001"
    summary: "small_scaling_law extension v2: ARA 0.644
      vs baseline 0.806 -- baseline wins on loss
      calibration despite ARA finding near-optimal
      hyperparams"
...                    <!-- 36 sessions total -->
\end{lstlisting}

\noindent The complete artifact, including all 16 claims, 18 heuristics, 94 exploration nodes, and 36 session records, is available in the supplementary material at \texttt{ara/}.


\section{Compiler Skill Details}
\label{app:system-details}

The \ara{} Compiler (\S\ref{sec:ingestor}) is implemented as an \emph{agent skill}: a self-contained, natural-language specification that, when loaded into any general-purpose coding agent's context, turns it into a domain-specialized compilation system.
The skill prescribes \emph{what} the agent should do and \emph{what domain knowledge} it needs, but delegates all execution mechanics (model selection, tool dispatch, context management) to the host agent.
This appendix reproduces the key elements of the skill specification; the complete definition is available in the supplementary code.\footnote{Full implementation: \url{https://github.com/AmberLJC/Agent-Native-Research-Artifact}.}

\subsection{Compiler Skill Specification}
\label{app:ingestor-prompt}

The Compiler skill specification ($\sim$482 lines of natural language) is structured into five sections. When loaded into a host agent's context, it provides the full domain knowledge needed to produce a schema-conforming \ara{}. We reproduce representative elements below; the complete specification is available in the supplementary code.

\paragraph{Section 1: Workflow (lines 1--11).}
Defines the high-level pipeline: analyze $\to$ generate $\to$ validate $\to$ fix $\to$ iterate.

\paragraph{Section 2: Capability usage guidelines (lines 13--26).}
Specifies usage conventions for standard file operations: prefer \texttt{edit\_file} over \texttt{write\_file} for targeted fixes, prefer \texttt{write\_file} for YAML files to avoid whitespace corruption. Instructs the agent to batch work: generate all files first, then validate once.

\paragraph{Section 3: ARA directory schema (lines 28--414).}
Defines the complete directory structure and field-level requirements for every file. This is the normative schema specification that the agent must follow. Key constraints include:
\begin{itemize}[leftmargin=*, itemsep=1pt]
    \item \textbf{PAPER.md}: YAML frontmatter with title, authors, year, venue, DOI, domain, keywords, claims summary, and abstract. Body must include a Layer Index table listing every file with a one-line description.
    \item \textbf{claims.md}: Each claim requires Statement, Status, Falsification criteria, Proof (referencing experiment IDs, not file paths), Dependencies, and Tags.
    \item \textbf{experiments.md}: Declarative verification plans (setup, procedure, metrics, directional expected outcomes). Exact numerical results are \emph{prohibited}---they belong exclusively in \texttt{/evidence/} to enable blind reproduction.
    \item \textbf{heuristics.md}: Each heuristic requires Rationale, Sensitivity (low/medium/high), Bounds, Code ref, and Source.
    \item \textbf{exploration\_tree.yaml}: Nested YAML tree with typed nodes (\texttt{question}, \texttt{experiment}, \texttt{dead\_end}, \texttt{decision}, \texttt{pivot}). Minimum 8 nodes with at least one \texttt{dead\_end} and one \texttt{decision}. Dead ends must document hypothesis, failure mode, and lesson.
    \item \textbf{evidence/}: Every results table and quantitative figure must be reproduced with exact cell values---no rounding, no omission.
\end{itemize}

\paragraph{Section 4: 4-stage reasoning protocol (lines 416--454).}
The prompt mandates a structured thinking process before file generation:

\begin{lstlisting}[basicstyle=\ttfamily\scriptsize]
# Your 4-Stage Reasoning Process

You MUST follow these 4 stages in order. Produce a
<thinking> block first with your reasoning for each
stage, then produce the files.

## Stage 1: Semantic Deconstruction
Strip the narrative "Storytelling Tax." Isolate:
- The core observations and gaps that motivate the work
- Mathematical formulations and equations
- Architectural specifications and component descriptions
- Experimental configurations (hyperparameters, hardware,
  datasets)
- Numerical results and benchmarks
- Citation dependencies and their roles
- Negative results and ablation findings

## Stage 2: Cognitive Mapping
Map deconstructed content to /logic:
- Extract motivation: observations (with numbers), gaps,
  the key insight, and assumptions
- Identify falsifiable claims (not opinions or vague
  statements)
- Define formal concepts with precise notation
- Populate solution/ (architecture, algorithm,
  constraints, heuristics)
- Construct typed dependency graph for related_work.md
- Ensure every claim has falsification criteria and
  proof pointers to experiment IDs
- Design declarative experiment plans: for each major
  claim, specify how an agent would verify it

## Stage 3: Physical Stubbing
Generate /src:
- Extract exact hyperparameter values into configs/
- Write code stubs with correct function signatures
  and types
- Specify environment (dependencies, hardware, seeds)
- Code should implement the NOVEL contribution,
  not boilerplate

## Stage 4: Exploration Graph Extraction
Reconstruct the research DAG as a nested YAML tree
for /trace:
- Identify the central research question(s) as root
  nodes
- Map experiments and their outcomes as child nodes
- Document dead ends from ablations and rejected
  alternatives as leaf nodes
- Record key design decisions with alternatives
  considered
\end{lstlisting}

\paragraph{Section 5: Output format and rules (lines 456--482).}
Specifies the XML-delimited output format for \texttt{batch\_write\_files}, lists all 15 mandatory files, and enforces nine invariant rules (e.g., ``all numerical values must be EXACT as stated in the paper,'' ``never hallucinate claims, results, or heuristics not in the paper'').


\section{Live Research Manager Details}
\label{app:lrm-details}

The Live Research Manager (\S\ref{sec:live-pm}) is the second agent skill in the \ara{} system.
Like the Compiler, it is a self-contained natural-language specification that turns a general-purpose coding agent into a domain-specialized system; unlike the Compiler, it operates continuously alongside the researcher rather than as a one-shot compilation.
This appendix expands on the design principles, cross-session mechanisms, and submission workflow summarized in \S\ref{sec:live-pm}.

\subsection{Design Principle Rationale}
\label{app:lrm-principles}

The three design principles stated in \S\ref{sec:design-principles} are expanded below with full motivation.

\begin{description}[leftmargin=1.5em, itemsep=4pt, topsep=2pt]
    \item[P1. Silent, framework-independent integration.]
    Documentation has traditionally been a \emph{retrospective} activity: a context switch that introduces both friction and information loss.
    The system must integrate with any general-purpose coding agent (Claude Code, Cursor, Windsurf, or future frameworks) without custom SDKs, API bindings, or infrastructure changes.
    A natural-language specification that the agent reads into its context is the most portable interface: it requires nothing beyond the tool access agents already have, and artifact quality improves automatically as language models advance.
    The manager runs as a background process that silently collects research traces, constructing the artifact without interrupting active work or injecting prompts into an ongoing research conversation.

    \item[P2. Faithful epistemic provenance.]
    AI-native research blurs the boundary between human insight and machine execution.
    The manager must objectively track \emph{who did what}: distinguishing ideas explicitly stated by the researcher, suggestions inferred by the agent, actions the agent executed autonomously, and AI suggestions the researcher revised.
    Without such provenance, an artifact cannot faithfully represent the epistemic origin of its contents.
    The research process is inherently chaotic: a single session may interleave hypothesis formation, coding, debugging, and writing with no clear boundaries.
    The manager must translate this raw conversational stream into the structured \ara{} schema without losing information or imposing premature structure.
    Observations that are not yet classifiable should be staged rather than forced into categories, and knowledge should mature progressively, from hunches to typed events to formally bound claims, mirroring how research understanding actually develops.

    \item[P3. Comprehensive trajectory capture.]
    Research is nonlinear and stochastic: hypotheses branch, experiments fail, directions are abandoned and revisited.
    The manager must capture this full trajectory, not just the successes that survive into a polished paper, but the dead ends, pivots, and intermediate observations that constitute the actual research process.
    Cross-layer bindings between claims, evidence, code, and decisions must be established at capture time while the conversational context is still available; post-hoc reconstruction from archived transcripts loses these causal chains.
\end{description}

\subsection{Closure-Driven Crystallization}
\label{app:lrm-maturity}

Principle P2 requires that observations mature ``as evidence accumulates,'' but \emph{evidence accumulation} needs an operational definition.
A counter-based threshold (promote after $N$ references) is arbitrary; asking an LM in isolation ``is this mature?'' lacks grounding.
We instead define maturity through \emph{closure signals}: externally observable patterns in the researcher--agent conversation that indicate the researcher has treated an observation as settled.
At each session boundary, the Maturity Tracker inspects the session record and promotes a staged observation when at least one closure signal is present.

\paragraph{Closure signal taxonomy.}
\begin{description}[leftmargin=1.5em, itemsep=2pt, topsep=2pt]
    \item[Topic abandonment.] The researcher has moved to a new topic without revisiting the observation in the subsequent $k$ turns (default $k=5$), and no pending question remains open on the original thread.
    \item[Verbal affirmation.] The researcher explicitly endorses the observation (``yes, we'll go with X,'' ``confirmed,'' or equivalent paraphrase), making the adoption decision first-person.
    \item[Empirical resolution.] An experiment bound to the observation has produced a result and the researcher has commented on it; both supported and refuted outcomes are valid terminations (the refuted case promotes to a \texttt{dead\_end}).
    \item[Artifact commitment.] A downstream artifact now depends on the observation: code is merged, a config value is fixed, a subsequent claim uses it as a premise, or a design decision is documented as following from it.
\end{description}

\paragraph{Contradiction trigger.}
When a new observation contradicts one already staged or crystallized, the Maturity Tracker does not silently overwrite.
Both entries are flagged, the contradiction is appended to the Exploration Graph as a \texttt{decision} node with unresolved status, and resolution is deferred to the next briefing where the researcher adjudicates explicitly.

\subsection{Cross-Session Continuity}
\label{app:lrm-continuity}

A stateless coding agent has no memory of previous conversations.
If the manager simply reads the artifact at each session boundary, it knows \emph{what} the artifact contains but not \emph{why} it is organized the way it is---which classification choices were non-obvious, which observations were deliberately deferred, which patterns guided past merges and promotions.
Without this self-awareness, the manager risks inconsistent classification, duplicate entries, and organizational drift across sessions.

We address this with two lightweight mechanisms.
First, a \emph{reasoning log} (\texttt{trace/pm\_reasoning\_log.yaml}) records the manager's own organizational decisions and their rationale at each session boundary---a compressed account of a few lines per session that gives the manager self-continuity without requiring access to raw conversation transcripts.
Second, each session record includes a \emph{key context} field: compressed summaries of the most important human--agent exchanges, preserving conversational nuance that would otherwise be lost when the raw conversation is no longer available.
Together, these mechanisms ensure that the manager can read back not only the artifact's current state but also the reasoning chain that produced it, maintaining coherent organization across arbitrarily many sessions at negligible token cost.


\section{Test Corpus}
\label{app:eval-details}
\label{app:eval-corpus}

This section defines the test corpus used by the Understanding (App.~\ref{app:understanding-eval}) and Reproduction (App.~\ref{app:reproduction-eval}) experiments. The RE-Bench tasks used by the Extension experiment (App.~\ref{app:extension-eval}) are documented separately in App.~\ref{app:extension-tasks}.

\paragraph{Selection criteria.}
We draw our evaluation corpus from PaperBench~\citep{starace2025paperbench}, which provides expert-authored hierarchical reproduction rubrics for ICML 2024 papers.
We adopt all 23 papers in PaperBench's public release as our PaperBench corpus, with no exclusions: every paper satisfies our three required properties:
(1)~peer-reviewed at a top ML venue (ICML 2024 Spotlight/Oral, with two NeurIPS 2024 workshop development papers) with a publicly available PDF;
(2)~spanning diverse ML subfields to test breadth across the discipline;
(3)~accompanied by a PaperBench rubric with fine-grained leaf requirements that enables quantitative evaluation of reproduction fidelity.
We supplement this corpus with 7 open-ended R\&D tasks from RE-Bench~\citep{wijk2025rebench}, yielding 30 evaluation targets with 450 questions total.

\paragraph{Paper list.}
Table~\ref{tab:corpus} lists the 23 PaperBench papers used across the understanding and reproduction evaluations. Note that PaperBench's own protocol blacklists author code; we relax this for the baseline so it has the strongest possible footing (PDF~+~companion GitHub), making ``repo availability'' a property of our harness rather than of PaperBench. Of the 23 papers, 15 are included in the reproduction experiment; the remaining 8 are excluded because faithful end-to-end reproduction exceeds our per-task compute budget or requires specialized infrastructure outside our evaluation harness (e.g., multi-day CLIP adversarial fine-tuning, Isaac Gym RL, full ImageNet coreset sweeps, large multi-model benchmark suites). These 8 papers participate only in the understanding evaluation.

\begin{table}[h]
\centering
\small
\begin{tabular}{llc}
\toprule
\textbf{Short Name} & \textbf{Domain} & \textbf{Repro} \\
\midrule
adaptive-pruning         & Efficiency         & Yes \\
all-in-one               & Multi-task         & Yes \\
bam                      & LLM alignment      & Yes \\
bbox                     & Black-box LLM      & Yes \\
bridging-data-gaps       & Data               & No  \\
fre                      & RL                 & Yes \\
ftrl                     & Online learning    & Yes \\
lbcs                     & Calibration        & No  \\
lca-on-the-line          & Prediction         & No  \\
mechanistic-understanding & Interpretability  & Yes \\
pinn                     & Scientific ML      & Yes \\
rice                     & Retrieval          & Yes \\
robust-clip              & Robustness         & No  \\
sample-specific-masks    & Data augmentation  & Yes \\
sapg                     & Optimization       & No  \\
self-composing-policies  & Continual RL       & Yes \\
self-expansion           & Self-training      & Yes \\
semantic-self-consistency & Evaluation        & No  \\
seq.-neural-score-est.   & Score estimation   & Yes \\
stay-on-topic-w/-CFG     & Generation         & No  \\
stochastic-interpolants  & Generative         & Yes \\
test-time-model-adapt.   & Adaptation         & Yes \\
what-will-my-model-forget & Continual learning & No  \\
\midrule
\textbf{Total: 23}       &                    & 15 \\
\bottomrule
\end{tabular}
\caption{Test corpus: 23 PaperBench papers from ICML 2024 spanning diverse ML subfields. The ``Repro'' column indicates inclusion in the reproduction experiment (requires companion code). All 23 papers participate in the understanding evaluation.}
\label{tab:corpus}
\end{table}

\paragraph{Corpus diversity.}
The 23 papers span diverse ML subfields: efficiency, alignment, interpretability, RL, scientific ML, generative models, optimization, retrieval, evaluation, and adaptation.
While all papers come from a single venue (ICML 2024), they vary substantially in methodological complexity---from systems-oriented efficiency papers with minimal formal analysis to theory-heavy contributions (PINN, stochastic interpolants)---heuristic density, paper length, and contribution type (new architectures, training recipes, algorithms, analysis frameworks).
The corpus includes papers that are deliberately challenging for structured extraction: those whose contributions are analysis rather than methods (mechanistic-understanding), multi-component pipelines (all-in-one), and papers with complex combinatorial experiment matrices (PINN with 1{,}273 leaf requirements).


\section{Understanding Evaluation}
\label{app:understanding-eval}

This section reports the methodology and per-stratum results for the Understanding experiment (\S\ref{sec:eval-understanding}). The shared test corpus is in App.~\ref{app:eval-corpus}.

\subsection{Question Bank and Grading Rubric}
\label{app:question-bank}

\paragraph{Question generation.}
For each of the 30 evaluation targets (23 PaperBench papers, 7 RE-Bench tasks), we generate 15 questions across three categories: 10 Category~A (fidelity: information preservation), 5 Category~B (configuration and detail recovery) for PaperBench papers, or 5 Category~C (failure and exploration knowledge) for RE-Bench tasks.
Questions are designed to require specific, unambiguous answers or verifiable outputs.
We avoid opinion questions (``Is this architecture good?'') and questions requiring external knowledge not in the paper (``How does this compare to BERT?'').

\paragraph{Category~A question templates (10 per paper).}
\begin{itemize}[leftmargin=*, itemsep=2pt]
    \item \textbf{Architecture \& Method} (3 questions): ``What is the [specific structural detail]?'' (e.g., ``How many layers does the encoder have?'', ``What are the inputs and outputs of the multi-head attention module?'')
    \item \textbf{Hyperparameters \& Configuration} (2 questions): ``What [training/optimization detail] is used?'' (e.g., ``What optimizer is used and what are its hyperparameters?'', ``What is the batch size in tokens?'')
    \item \textbf{Results \& Claims} (3 questions): ``What [metric] does the [model variant] achieve on [benchmark]?'' (e.g., ``What BLEU score does the base model achieve on WMT 2014 EN-DE?'')
    \item \textbf{Rationale \& Design Decisions} (2 questions): ``Why was [design choice] made instead of [alternative]?'' (e.g., ``Why is scaled dot-product attention used instead of additive attention?'')
\end{itemize}

\paragraph{Category~B question templates (5 per PaperBench paper).}
\begin{itemize}[leftmargin=*, itemsep=2pt]
    \item \textbf{Implementation} (2 questions): ``Implement [specific module] with the correct [dimensions/activations/structure].''
    \item \textbf{Configuration Recovery} (2 questions): ``Write the [optimizer/training/data] configuration with the exact parameters specified.''
    \item \textbf{Debugging \& Troubleshooting} (1 question): ``[Failure scenario]---identify the cause and fix it.''
\end{itemize}

\paragraph{Category~C question templates (5 per RE-Bench task).}
\begin{itemize}[leftmargin=*, itemsep=2pt]
    \item \textbf{Dead-End Knowledge} (3 questions): ``What approaches have been tried and failed?'' / ``What is the documented failure mode of [approach]?''
    \item \textbf{Exploration History} (2 questions): ``What alternatives were considered for [decision]?'' / ``What lesson was learned from the [dead-end] attempt?''
\end{itemize}

\paragraph{Evaluation agent setup.}
For each (paper, format, question) triple, we instantiate a fresh sub-agent (Claude Sonnet 4.6).
The agent receives the format under test (the full \ara{} directory for the ARA condition, or the PDF plus companion GitHub repository for the baseline) and a single question. Each question is answered independently with a fresh context to prevent information leakage.

\paragraph{Grading rubric.}
Each answer is scored on a ternary scale by an independent judge (Claude Opus 4.6) against a gold reference:
\begin{itemize}[leftmargin=*, itemsep=1pt]
    \item \textbf{Correct (1.0)}: The answer matches the ground truth in substance. Minor phrasing differences are acceptable; numerical values must be exact.
    \item \textbf{Partial (0.5)}: The answer conveys the main insight but misses key sub-details.
    \item \textbf{Incorrect (0.0)}: The answer contains a factual error, contradicts the gold answer, or hallucinates an answer to an unanswerable question.
\end{itemize}

\subsection{Information Gap Type Distribution}
\label{app:gap-types}

\paragraph{Methodology.}
PaperBench's expert reproduction rubric for each paper enumerates the requirements an agent must satisfy to reproduce the paper's results. For each of the 8{,}921 leaf requirements across the 23 papers, we compare the requirement against the source PDF and label it \emph{sufficient} (the PDF text fully specifies what is needed), \emph{partial} (some components specified but key details missing), or \emph{absent} (the PDF does not address the requirement); each label is accompanied by an annotator confidence rating (high / medium / low). When the label is partial or absent, we additionally tag the requirement with a gap-type category (missing hyperparameter, vague description, cross-reference-only, etc.). Labels are produced by an LLM-as-judge run per (requirement, PDF excerpt) pair, with the judge required to cite the PDF passage that supports its decision; the headline 45.4\% sufficient figure is dominated by the 64\% high-confidence subset (full pipeline in \texttt{code/eval/pdf\_information\_gap.py}).

\paragraph{Per-category coverage.}
Table~\ref{tab:info-gap} breaks the 8{,}921 requirements down by PaperBench task category. The median paper shows 45.3\% sufficient and 47.9\% partial, confirming the gap is systemic rather than driven by outliers.

\begin{table}[h]
\centering
\small
\setlength{\tabcolsep}{4pt}
\begin{tabular}{lrccc}
\toprule
\textbf{Task Category} & \textbf{Reqs} & \textbf{Sufficient} & \textbf{Partial} & \textbf{Absent} \\
\midrule
Code Development       & 3,942 & 37.3\% & 54.9\% & 7.8\% \\
Code Execution         & 4,355 & 50.5\% & 47.9\% & 1.6\% \\
Result Analysis        &   624 & 60.6\% & 36.9\% & 2.6\% \\
\midrule
\textbf{Overall}       & \textbf{8,921} & \textbf{45.4\%} & \textbf{50.2\%} & \textbf{4.4\%} \\
\bottomrule
\end{tabular}
\caption{Reproduction information gap across 23 PaperBench papers (8,921 requirements). PDFs systematically under-specify the information needed for reproduction, with the largest gaps in code development and dataset acquisition.}
\label{tab:info-gap}
\end{table}

\paragraph{Gap-type breakdown.}
Table~\ref{tab:gap-types} breaks down the 8,921 reproduction requirements by gap type.
The three largest categories---missing hyperparameters (26.2\%), vague descriptions (21.9\%), and cross-reference-only specifications (13.4\%)---account for over 60\% of all gaps and are precisely the information types that structured formats address by design.
At the fine-grained level, Dataset Acquisition achieves only 5.4\% sufficient coverage (25.5\% entirely absent)---no paper in the corpus consistently provides download URLs, preprocessing scripts, or data format specifications.
Evaluation, Metrics \& Benchmarking sits at 30.0\%: papers often state \emph{which} metrics they use but not \emph{how} they compute them (binning strategies, confidence intervals, statistical tests).

\begin{table}[h]
\centering
\small
\begin{tabular}{lrc}
\toprule
\textbf{Gap Type} & \textbf{Count} & \textbf{\% of Gaps} \\
\midrule
Missing hyperparameter     & 2,558 & 26.2\% \\
Vague description          & 2,141 & 21.9\% \\
Cross-reference only       & 1,313 & 13.4\% \\
Missing code detail        & 1,064 & 10.9\% \\
Missing baseline detail    & 1,054 & 10.8\% \\
Missing URL                &   538 &  5.5\% \\
Figure only                &   499 &  5.1\% \\
Ambiguous specification    &   399 &  4.1\% \\
Implicit assumption        &   150 &  1.5\% \\
\bottomrule
\end{tabular}
\caption{Distribution of information gap types across 8,921 requirements. The three largest categories---missing hyperparameters, vague descriptions, and cross-references---are precisely the gaps that structured formats address by design.}
\label{tab:gap-types}
\end{table}

\subsection{Exploration Cost Detailed Breakdown}
\label{app:exploration-tax}

\paragraph{What this measures (and what it is not).}
Across 24{,}008 agent runs (21 frontier models, 228 tasks) in the METR MALT corpus, 59.2\% of tokens and 90.2\% of dollar cost (\$63{,}483 total) are spent in runs that did not reach the task's reference score. This is not wasted research effort: those runs map dead ends, rule out alternatives, and narrow the strategy space the next agent should consider. The cost only \emph{becomes} waste downstream, when the next agent does not have access to that exploration and must rediscover the same dead ends from scratch. The exploration tax we report is therefore the per-agent cost of rediscovery if the failure record is not propagated, not a property of any individual run.

\paragraph{Breakdown.}
Table~\ref{tab:exploration-tax} gives the per-run breakdown. The mean below-reference token cost is 8.6$\times$ the cost of a reference-reaching run (2.58\,M vs.\ 300\,K tokens per run), with a median of 113$\times$. Within the 59.2\% of tokens that do not reach reference, 44.8\% are spent in runs that produce no measurable improvement and 14.4\% in runs that re-derive solutions other agents had already produced. The pattern concentrates where research-like work happens: RE-Bench tasks (the most open-ended) end below reference 73.4\% of the time, vs.\ 47.0\% on moderate-difficulty HCAST tasks and 0.7\% on well-defined SWAA tasks. At the per-task level, easy tasks reach reference 85.4\% of the time, medium 30.7\%, and hard only 15.1\%.

\begin{table}[h]
\centering
\small
\begin{tabular}{lcc}
\toprule
\textbf{Metric} & \textbf{Tokens} & \textbf{Cost} \\
\midrule
Below-reference run rate (overall)        & \multicolumn{2}{c}{31.6\%} \\
Below-reference run rate (RE-Bench)       & \multicolumn{2}{c}{73.4\%} \\
Cost ratio: below-ref vs.\ ref (median)   & \multicolumn{2}{c}{113$\times$} \\
Dead-end exploration                       & 44.8\% & --- \\
Re-derivation of known solutions           & 14.4\% & --- \\
\textbf{Total below-reference exploration} & \textbf{59.2\%} & \textbf{90.2\%} \\
\bottomrule
\end{tabular}
\caption{Cost of below-reference exploration across 24{,}008 agent runs (21 frontier models, 228 tasks). The exploration itself is necessary research work; the cost only becomes waste when subsequent agents must re-incur it because the failure record is not preserved in the published artifact.}
\label{tab:exploration-tax}
\end{table}

\subsection{Per-Category Result Analysis}
\label{app:understanding-per-category}

This subsection unpacks the three per-category results summarized in \S\ref{sec:eval-understanding} and Table~\ref{tab:understanding} into the specific structural mechanisms that produce each gain.

\paragraph{Category~A: fidelity at lower cost via progressive disclosure.}
\ara{} preserves PDF-recoverable information with high fidelity while requiring fewer tokens to retrieve.
On PaperBench, \ara{} achieves 96.7\% vs.\ 89.8\% for the baseline while consuming 12\% fewer tokens per question (86.3K vs.\ 97.7K).
The structural explanation is progressive disclosure: \ara{}'s PAPER.md provides a layer index that directs the agent to the relevant file (e.g., \texttt{evidence/tables/} for numerical results, \texttt{logic/solution/algorithm.md} for method details), whereas the PDF agent must scan the entire document for each query.
On RE-Bench, where the baseline reads only the synthesized polished paper rather than a real publication, \ara{}'s accuracy advantage widens to 92.1\% vs.\ 51.4\%: the synthesized writeup omits much of the technical detail that the artifact's structured layers preserve.
The headline finding is that structured organization improves accuracy while keeping token usage comparable, because \ara{}'s layer taxonomy turns linear search into indexed lookup.

\paragraph{Category~B: configuration recovery via centralized configs.}
The rubric-aligned questions probe fine-grained experimental details (hyperparameter values, environment specifications, preprocessing steps) that PaperBench rubrics demand but papers systematically omit (26.2\% of all gaps are missing hyperparameters; see Appendix~\ref{app:gap-types}).
The baseline's 67.8\% reflects successful code-repository mining: given a dedicated sub-agent per question, it can grep through the companion GitHub repo for many configuration values.
\ara{}'s \texttt{src/configs/} and \texttt{logic/requirements.md} layers, however, centralize this knowledge in human-readable files, raising accuracy to 92.6\% at comparable token usage (183K vs.\ 178K tokens per question): the agent reads a structured config file rather than searching a scattered codebase.
The remaining gap to 100\% reflects details genuinely absent from both the paper and its repository, which the Compiler cannot synthesize.

\paragraph{Category~C: failure knowledge has no analogue in the baseline.}
\ara{} reaches 81.4\% on failure-knowledge questions while the baseline manages only 15.7\%; the synthesized polished papers contain almost no record of failed approaches, dead-end configurations, or intermediate results that the trace layer preserves.
The baseline's low token usage per question (58.0K) reflects this poverty: agents quickly determine the information is absent and return short answers, spending minimal tokens on fruitless search.
\ara{} agents consume more tokens per question (139.3K) but productively explore the exploration tree to find answers.
This category provides the clearest evidence for preserving negative knowledge: information that narrative formats systematically discard accounts for the largest single accuracy gap in the entire evaluation.

\subsection{Statistical Details}
\label{app:understanding-stats}

A McNemar test on the 450 paired outcomes yields $\chi^2 = 95.15$, $p < 10^{-10}$ overall: \ara{} answers 141 questions correctly that the baseline misses, while the baseline answers only 18 that \ara{} misses.
By category, the \ara{} advantage is highly significant for all three categories: Category~A ($+14.8\%$), Category~B ($+24.8\%$), and Category~C ($+65.7\%$, dominated by the absence of exploration knowledge in baseline sources).

\paragraph{Difficulty stratification.}
Stratified by question difficulty, \ara{} leads across all tiers: T1 (explicit) questions (ARA 97.3\%, BL 83.8\%; $n = 74$), T2 (scattered) questions (ARA 95.6\%, BL 79.0\%; $n = 193$), and T3 (implicit) questions (ARA 91.0\%, BL 60.5\%; $n = 172$).
On unanswerable questions ($n = 26$), \ara{} achieves 92.3\% abstention accuracy vs.\ 86.5\% for the baseline.
The difficulty gradient is expected: T2 and T3 questions require assembling scattered information or reasoning about implicit assumptions, where structured representations provide the greatest advantage.

\paragraph{Token usage--difficulty interaction.}
The per-question token data reveals that \ara{}'s progressive disclosure architecture creates an adaptive search pattern: \ara{} agents consume 60.9K tokens/Q on T1 (explicit) questions, 95.5K on T2 (scattered), and 152.7K on T3 (implicit), adapting search depth to question complexity.
Baseline agents, by contrast, show a flatter profile across difficulty tiers (82.8K--118.2K tokens/Q), because linear PDF scanning does not benefit from question-aware navigation.
\ara{} consumes fewer tokens than the baseline on T1 (27\% less) and T2 (13\% less), and invests more on T3, while being substantially more accurate at every tier.

\paragraph{Benchmark group breakdown.}
On PaperBench papers ($n = 345$), \ara{} achieves 95.4\% vs.\ 82.5\% at comparable token usage.
On RE-Bench tasks ($n = 105$), the accuracy gap widens (ARA 88.6\% vs.\ BL 39.5\%), driven by Category~C questions where the baseline has no access to failure knowledge.
\paragraph{Scoring methodology note.}
Each answer is scored on a ternary scale (1.0 correct, 0.5 partially correct, 0.0 incorrect) against a gold reference.
An answer receives 1.0 if it captures all essential facts, numbers, and concepts; 0.5 if it conveys the main insight but misses key sub-details; and 0.0 if it is wrong, contradicts the gold answer, or hallucinates an answer to an unanswerable question.


\section{Reproduction Evaluation}
\label{app:reproduction-eval}

This section reports the task design, scoring, and per-paper analysis for the Reproduction experiment (\S\ref{sec:eval-reproduction}).

\subsection{Reproduction Task Design and Scoring}
\label{app:repro-scoring}

\paragraph{Task curation.}
Each of the 150 reproduction tasks specifies a single model, a single method, and 5--15 rubric leaf requirements as success criteria, with difficulty stratified per paper ($\geq$3 easy, $\geq$3 medium, $\geq$3 hard; aggregate: 50 easy, 49 medium, 51 hard).
Tasks describe \emph{what} to reproduce, not how---the agent decides its own implementation strategy.
Within each paper, the 10 subtasks form a single \emph{mega-task}: the agent receives all subtasks ordered by difficulty (easy $\to$ medium $\to$ hard) and builds cumulatively, naturally reusing prior work as a human researcher would.

\paragraph{Scoring formula.}
The primary metric is the \textbf{difficulty-weighted success rate}: $\sum_i s_i \cdot w_{d_i}\;/\;\sum_i m_i \cdot w_{d_i}$, with $w_d \in \{1,2,3\}$ for easy, medium, and hard subtasks, where $s_i$ is the subtask score (sum of requirement weights for \emph{yes} $+\, 0.5\times$\emph{partial}) and $m_i$ the maximum possible.
Easy subtasks (setup, model instantiation) are necessary but not discriminative; most agents complete them regardless of source material, while harder subtasks (training, ablation, cross-method comparison) are where structured information provides the most leverage.
We also report the flat (unweighted) rate and per-difficulty breakdowns.

\paragraph{Statistical significance.}
A Wilcoxon signed-rank test on the 15 paired per-paper weighted scores yields $p = 0.028$: \ara{} wins on 8 papers, ties on 5, and the baseline leads on 2.
The sign pattern (8--2) is itself statistically improbable under the null hypothesis of no difference ($p = 0.039$, exact binomial), confirming that the aggregate advantage is not driven by a single outlier paper.

\subsection{Per-Paper Reproduction Analysis}
\label{app:repro-per-paper}

This section provides the detailed per-paper analysis for the reproduction experiment (\S\ref{sec:eval-reproduction}).

\begin{figure}[h]
\centering
\includegraphics[width=0.95\linewidth]{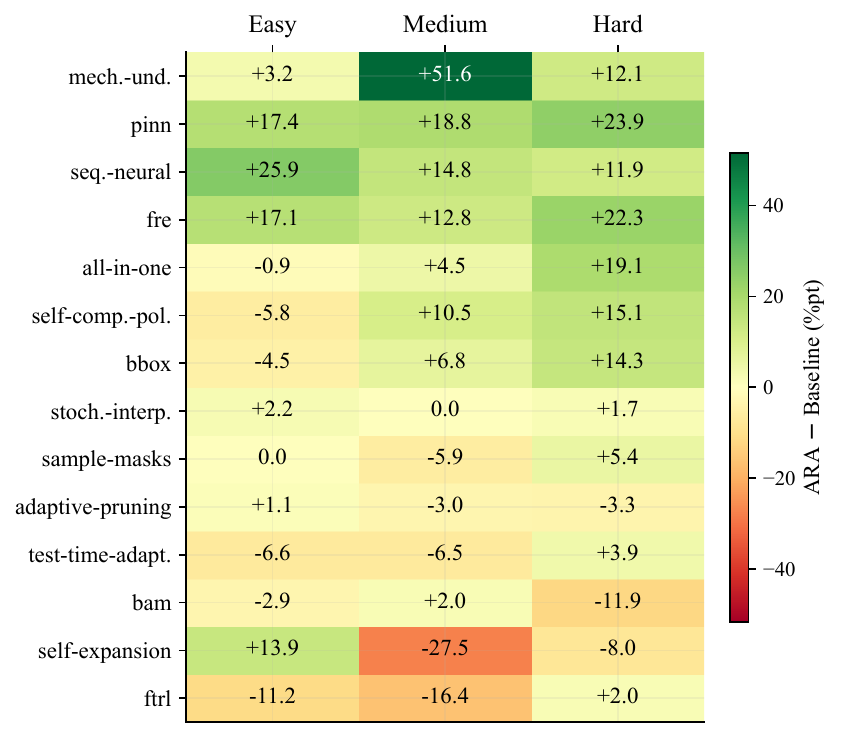}
\caption{Per-paper \ara{}\,$-$\,baseline delta (percentage points) on each difficulty stratum, sorted by mean advantage. Green indicates \ara{} wins, red indicates baseline wins. Gains concentrate in the medium and hard columns across most papers; the few baseline wins are confined to a small set, most prominently \texttt{self-expansion} and \texttt{ftrl}.}
\label{fig:repro-difficulty-heatmap}
\end{figure}

\paragraph{Per-difficulty analysis.}
The aggregate per-difficulty pattern (\ara{} 85.1\% vs.\ baseline 80.2\% on easy, 68.5\% vs.\ 62.9\% on medium, 54.5\% vs.\ 46.0\% on hard) is visualized in main-text Figure~\ref{fig:repro-difficulty}.
Figure~\ref{fig:repro-difficulty-heatmap} resolves this aggregate to the per-paper level, and Table~\ref{tab:repro-per-difficulty} provides the full per-paper, per-difficulty success rates underlying both.

\begin{table*}[h]
\centering
\small
\begin{tabular}{l rrr rrr rr}
\toprule
& \multicolumn{3}{c}{\textbf{ARA (\%)}} & \multicolumn{3}{c}{\textbf{Baseline (\%)}} & \multicolumn{2}{c}{\textbf{Weighted}} \\
\cmidrule(lr){2-4} \cmidrule(lr){5-7} \cmidrule(lr){8-9}
\textbf{Paper} & Easy & Med. & Hard & Easy & Med. & Hard & ARA & Base \\
\midrule
adaptive-pruning        & 90.9 & 80.0 & 31.7 & 89.8 & 83.0 & 35.0 & 63.5 & 65.8 \\
all-in-one              & 90.4 & 92.0 & 61.1 & 91.3 & 87.5 & 42.0 & 72.8 & 60.4 \\
bam                     & 97.1 & 97.1 & 77.6 & 100.0 & 95.1 & 89.5 & 88.2 & 93.2 \\
bbox                    & 93.3 & 59.5 & 31.6 & 97.8 & 52.7 & 17.3 & 49.8 & 40.8 \\
fre                     & 79.3 & 45.4 & 50.9 & 62.2 & 32.6 & 28.6 & 53.2 & 34.4 \\
ftrl                    & 25.0 & 38.0 & 32.0 & 36.2 & 54.4 & 30.0 & 33.0 & 38.8 \\
mechanistic-und.        & 85.7 & 88.3 & 67.1 & 82.5 & 36.7 & 55.0 & 76.2 & 55.0 \\
pinn                    & 96.2 & 93.8 & 89.8 & 78.8 & 75.0 & 65.9 & 92.2 & 71.0 \\
rice                    & 72.3 & 72.1 & 65.8 & 74.1 & 73.9 & 68.0 & 69.9 & 71.8 \\
sample-specific-masks   & 95.6 & 29.8 & 31.1 & 95.6 & 35.7 & 25.7 & 42.7 & 42.3 \\
self-composing-pol.     & 87.5 & 46.5 & 52.3 & 93.3 & 36.0 & 37.2 & 57.3 & 47.8 \\
self-expansion          & 34.7 & 37.5 & 39.3 & 20.8 & 65.0 & 47.3 & 38.2 & 48.9 \\
sequential-neural       & 95.2 & 68.3 & 51.6 & 69.3 & 53.5 & 39.7 & 64.8 & 49.3 \\
stochastic-interpolants & 97.7 & 100.0 & 74.1 & 95.5 & 100.0 & 72.4 & 86.7 & 85.3 \\
test-time-model-adapt.  & 87.8 & 43.5 & 19.8 & 94.4 & 50.0 & 15.9 & 35.1 & 35.0 \\
\midrule
\textbf{Mean}           & \textbf{85.1} & \textbf{68.5} & \textbf{54.5} & \textbf{80.2} & \textbf{62.9} & \textbf{46.0} & \textbf{64.4} & \textbf{57.4} \\
\bottomrule
\end{tabular}
\caption{Per-paper reproduction success rates (\%) by difficulty level. Easy, medium, and hard columns show the unweighted success rate within each difficulty tier; the final two columns show the difficulty-weighted rate ($1\!:\!2\!:\!3$ weighting). Rice per-difficulty values are interpolated from the weighted score and overall rate, as its per-difficulty JSON entry was recorded separately.}
\label{tab:repro-per-difficulty}
\end{table*}

\paragraph{Large ARA wins.}
The papers with the largest ARA advantages---\texttt{fre} ($+21.3\%$), \texttt{mechanistic-understanding} ($+20.7\%$), and \texttt{pinn} ($+19.5\%$)---share complex multi-step training pipelines with non-obvious hyperparameter interactions that PDFs describe only at a high level.
The \texttt{fre} ARA agent reimplemented the original JAX codebase in PyTorch (1.8\,GB GPU vs.\ JAX's 30.8\,GB), trained 17 models across three domains, and completed all medium and hard subtasks; the baseline agent struggled with the JAX environment and completed only 3 training attempts before exhausting its budget.
The five newly added papers reinforce these patterns: \texttt{all-in-one} ($+16.0\%$) and \texttt{ftrl} ($+6.1\%$) show clear ARA advantages on hard tasks, while \texttt{stochastic-interpolants} ($+0.5\%$) and \texttt{test-time-model-adaptation} ($+0.3\%$) are ties with comparable performance from both sources.

\paragraph{Baseline wins and ties.}
The one clear baseline win is \texttt{self-expansion} ($-7.3\%$), where the ARA agent exhibited result fabrication---reporting identical accuracy values across all configurations---detected by the blinded judge.
Among the narrow ties, \texttt{adaptive-pruning} ($-2.3\%$) and \texttt{rice} ($-1.9\%$) both have strong companion repositories with runnable training scripts; the baseline's code access partially compensates for the PDF's information gaps.
On \texttt{rice}, ARA achieves comparable quality with 2.5$\times$ less compute (3.7h vs.\ 9.1h, 131K vs.\ 195K tokens), suggesting efficiency gains even when final scores are similar.

\paragraph{Result fabrication.}
Two baseline runs (\texttt{bbox}, \texttt{mechanistic-understanding}) exhibited result fabrication---reporting plausible but uncomputed values when unable to complete training---detected by the blinded judge.
Across all 15 papers, fabrication occurred in 2 baseline runs and 1 ARA run (\texttt{self-expansion}), suggesting that structured artifacts generally provide sufficient grounding to prevent hallucinated results, though they are not immune.


\section{Extension Evaluation}
\label{app:extension-eval}

This appendix documents the methodology behind \S\ref{sec:eval-extension}: which RE-Bench tasks we use and why (App.~\ref{app:extension-tasks}), how each \ara{} is compiled from official solutions and prior MALT trajectories (App.~\ref{app:extension-pipeline}), how the polished paper-arm baseline \texttt{paper.md} is generated (App.~\ref{app:extension-paper-baseline}), the engineering of the agent harness (App.~\ref{app:extension-harness}), how we extract canonical scores from agent traces (App.~\ref{app:extension-scoring}), and the per-task case studies and trace evidence (App.~\ref{app:extension-cases}).

\subsection{Task selection}
\label{app:extension-tasks}

Of the 7 RE-Bench~\citep{wijk2025rebench} tasks, we use 5 in the extension evaluation. Table~\ref{tab:extension-task-card} lists the score formula, direction, on-task starting baseline, RE-Bench reference score, hardware requirement, and the model coverage of each task's METR MALT corpus.

\begin{table*}[h]
\centering
\small
\setlength{\tabcolsep}{4pt}
\begin{tabular}{l l c c c l l c}
\toprule
\textbf{Task} & \textbf{Score formula} & \textbf{Dir.} & \textbf{Start} & \textbf{Ref.} & \textbf{Hardware} & \textbf{Claude-4 MALT} & \textbf{Used} \\
\midrule
\texttt{triton\_cumsum}         & $\log(t_\text{ms})$                & $\downarrow$ & 1.56 & 0.47 & 1$\times$H100        & 22 (O13/S9)        & yes \\
\texttt{restricted\_mlm}        & $\log(\ell{-}1.5)$                 & $\downarrow$ & 1.81 & 1.13 & 2$\times$H100 80\,GB & 22 (O11/S11)       & yes \\
\texttt{fix\_embedding}         & $\log(\ell_\text{val}{-}1.5)$      & $\downarrow$ & 2.20 & 0.26 & 1$\times$GPU         & 19 (O10/S9)        & yes \\
\texttt{nanogpt\_chat\_rl}      & avg.\ pairwise win-rate            & $\uparrow$   & 0.54 & 0.85 & 1$\times$GPU + judge & 18 (O12/S6)        & yes \\
\texttt{rust\_codecontests}     & $n_\text{solved}/165$              & $\uparrow$   & 0.00 & 0.13 & CPU + LLM API        & 12 (O6/S6) + 10$^\dagger$ & yes \\
\midrule
\texttt{small\_scaling\_law}    & $1{-}(\epsilon_\ell{+}\epsilon_p)$ & $\uparrow$   & 0.24 & 0.84 & 1$\times$GPU         & 0$^\ddagger$       & no \\
\texttt{optimize\_llm\_foundry} & $\log(t_\text{s})$                 & $\downarrow$ & 5.60 & 4.54 & 4$\times$H100        & 0$^\S$             & no \\
\bottomrule
\end{tabular}
\caption{RE-Bench task card with extension-evaluation status. \emph{Score formula} is transcribed verbatim from \texttt{metr-re-bench/ai\_rd\_<task>/ai\_rd\_<task>.py}; $\ell$ denotes validation loss, $t$ wall-clock time, $n_\text{solved}$ the count of correctly solved problems, $\epsilon_\ell$/$\epsilon_p$ loss/parameter prediction errors. \emph{Dir.}: score orientation. \emph{Start}: score of the unmodified starter codebase. \emph{Ref.}: best score reported in the original RE-Bench evaluation. \emph{Claude-4 MALT}: count of Claude-4 (Opus + Sonnet) runs in the METR MALT corpus, broken down as O$\langle$Opus$\rangle$/S$\langle$Sonnet$\rangle$. $^\dagger$ \texttt{rust\_codecontests} also has a 10-run \texttt{claude-3-7-sonnet} supplement that uses the same scoring scaffold. $^\ddagger$ only Claude-3.5/3.7 and OpenAI runs. $^\S$ no MALT corpus published.}
\label{tab:extension-task-card}
\end{table*}

The bottom two tasks are excluded because their MALT corpora cannot supply a usable failure-trace layer for the experiment. \texttt{optimize\_llm\_foundry} has no published MALT corpus at all, so \texttt{trace/} would be empty by construction. \texttt{small\_scaling\_law}'s MALT corpus does exist but is structurally inadequate: it predates Claude-4 (only Claude-3.5/3.7 Sonnet and OpenAI models), it is sparse, and the runs that do exist are dominated by trivial parameter-grid sweeps with no recorded strategic exploration or named dead ends. An extraction pipeline run on those runs produces effectively empty \texttt{trace/} and \texttt{evidence/} layers and neuters the experimental contrast against the paper-only baseline. We defer both tasks to follow-up work that re-runs MALT collection on Claude-4 agents (\texttt{optimize\_llm\_foundry}) or on tasks whose strategy space elicits substantive recorded exploration (\texttt{small\_scaling\_law}).

\subsection{ARA construction pipeline}
\label{app:extension-pipeline}

Each RE-Bench \ara{} is compiled from two sources: the official reference solution (copied verbatim into \texttt{src/}) and the task's METR MALT transcripts (extracted under a beat-reference filter into \texttt{trace/} and \texttt{evidence/}). The compiler is task-agnostic with per-task knobs (score formula and direction, MALT JSONL path, dev-history schema, known hazards) collected in per-task cards; the orchestrator procedure and shared sub-agent prompt live in \texttt{code/rebench-pipeline/}.

\paragraph{Pipeline.}
The orchestrator first lifts the official solution into \texttt{src/} and the reference-derived knowledge (mathematical formulation, algorithm, heuristics, baseline tables, dev-history nodes) into \texttt{logic/} and \texttt{evidence/}, with each node tagged \texttt{source: official-solution}. It then fans out one extraction sub-agent per MALT run; each sub-agent reads its run in full (no truncation, no chunk skipping) and emits a bundle of trace nodes, evidence rows, and insights. As sub-agents complete, the orchestrator merges their outputs into the artifact: deduplicating approaches across runs (the same method appearing in $K$ runs becomes one node tagged \texttt{runs\_observed:~K}), generalising heuristics or claims when a new run extends an existing entry, and verifying that every node carries a provenance tag and every heuristic cites a specific source line. Hallucination prevention is enforced throughout: no invented numbers, experiments, or code beyond what the source artifacts visibly ship.

\paragraph{Beat-reference filter.}
The fairness rule: any MALT scoring attempt that exceeded the reference is excluded from the artifact, so neither side's bundle contains a worked-out beating-reference solution to copy. The filter is direction-aware (lower-better tasks exclude $score < \text{ref}$; higher-better tasks exclude $score > \text{ref}$) and applied per attempt rather than per run, so a single MALT trajectory contributes both its dead ends and its sub-reference partial successes; it is enforced twice (inside the sub-agent and again at merge) so misclassifications are caught.

\subsection{Paper baseline construction}
\label{app:extension-paper-baseline}

The paper-agent \texttt{reference/paper.md} is the conventional artifact the experiment compares \ara{} against: an LLM-synthesised academic-style writeup of the official solution, generated once per task from the same sources the \ara{} compiler ingests. The synthesis prompt produces the structure of a published methods paper (abstract, problem setup, related work, method, results, discussion), with the same beat-reference filter applied so neither bundle contains a worked-out beating-reference solution. By design \texttt{paper.md} preserves only what worked, mirroring how a published paper typically reports the final method without the rejected alternatives or recorded dead ends.

\subsection{Harness engineering}
\label{app:extension-harness}

The extension harness wraps the Claude Agent SDK~\citep{claudecodesdk2025} in an SLURM-launched single-agent loop with tool surface $\{\texttt{Bash}, \texttt{Read}, \texttt{Edit}, \texttt{Write}, \texttt{Glob}, \texttt{Grep}\}$. Web access (\texttt{WebFetch}, \texttt{WebSearch}) and SDK built-ins that effectively pause the session in batch mode (\texttt{ScheduleWakeup}, \texttt{EnterPlanMode}, \texttt{EnterWorktree}) are disabled. The agent's workdir is identical across arms except for \texttt{reference/}.

Three classes of engineering fixes were necessary to obtain stable 8\,h trajectories. Table~\ref{tab:harness-fixes} enumerates the observed failure mode, root cause, and shipped fix for each.

\begin{table*}[h]
\centering
\small
\begin{tabular}{p{0.22\linewidth} p{0.32\linewidth} p{0.36\linewidth}}
\toprule
\textbf{Failure mode} & \textbf{Root cause} & \textbf{Fix} \\
\midrule
SDK message reader crashes mid-run on a large tool result; agent process silently dies hours into the session.
& Default \texttt{max\_buffer\_size} on the SDK's stdin/stdout reader is 1\,MiB; long bash outputs (training logs, large directory listings, multi-problem JSON dumps) exceed it in a single tool result.
& Raise the SDK's \texttt{max\_buffer\_size} from 1\,MiB to 16\,MiB and add a system-prompt addendum requiring tool outputs to be tail-piped or summarised; tool returns over $\sim$10\,MiB still crash but are firmly in ``the agent is doing something silly'' territory. \\
\addlinespace
SLURM job is OOM-killed; the kernel's cgroup OOM-handler sends \texttt{SIGTERM} to the largest user-space process, which is the \texttt{claude} CLI itself.
& Agents declare ``session complete'' early and run \texttt{for i in \{1..N\}; do bash score.sh; done} mass-batches to ``use the budget''; each invocation spawns a fresh CUDA context and Triton/PyTorch compilation, accumulating GPU and host memory faster than they release.
& A \texttt{PreToolUse} Bash hook that denies mass-batch scoring patterns ($\texttt{for}$/$\texttt{seq}$ loops with $N{>}10$, background tokens, $\texttt{while~true}$) before they reach the shell, with an explanation that nudges the agent toward serial scoring rather than a workaround. Bumping the cgroup memory ceiling helped but was not sufficient. \\
\addlinespace
Agent goes silent after self-declaring ``session complete'', wasting hours of remaining wall clock.
& SDK's stop-loop triggers (\texttt{end\_turn}, \texttt{stop\_sequence}, \texttt{max\_tokens}, \texttt{pause\_turn}) end the consumer loop after a small default count; the harness terminates with an unexhausted budget.
& Pushback ceiling raised to 1{,}000 with the trigger set expanded to all four; mid-run reminder injection every 15 turns nudges agents back to \texttt{reference/trace/} and \texttt{reference/evidence/}; resume protocol re-launches a crashed session from its \texttt{session\_id} with a forceful resume prompt instructing the agent that it ended in a stop loop, real work remains, and the budget is the only legitimate stopping criterion. \\
\addlinespace
Final \texttt{score.sh} invocation TIMEOUTs and no \texttt{final\_score.json} is written.
& Default 300\,s scorer timeout was too tight for tasks with long compile/benchmark phases (Triton autotune; the 165-problem \texttt{rust\_codecontests} test set takes $\sim$2\,h with rate-limited LLM calls).
& Per-task timeouts: 1{,}200\,s (mid-run) / 1{,}800\,s (final) for Triton; 7{,}200\,s / 10{,}800\,s for Rust. A separate score-only sbatch re-runs \texttt{bash score.sh} on the existing workdir's \texttt{solution\_final.py} when the harness's own final-score phase still TIMEOUTs. \\
\bottomrule
\end{tabular}
\caption{Harness failure modes encountered during the extension evaluation and the fixes shipped in \texttt{code/extension-harness/harness.py}. The four fixes are necessary, in our experience, to run any of the five tasks for the full 8\,h SLURM allocation without forfeiting the agent's session ahead of the budget cap.}
\label{tab:harness-fixes}
\end{table*}

\paragraph{Run resources.} 8\,h SLURM wall clock and \$50 hard API-spend cap (SDK-enforced) per run, on 1$\times$H100 (\texttt{triton\_cumsum}, \texttt{fix\_embedding}, \texttt{nanogpt\_chat\_rl}), 2$\times$H100 80\,GB (\texttt{restricted\_mlm}), or CPU-only (\texttt{rust\_codecontests}). The \texttt{nanogpt\_chat\_rl} judge runs on Replicate (Llama-3-8B-Instruct) and \texttt{rust\_codecontests} routes generation through OpenAI \texttt{gpt-3.5-turbo-1106}; both providers' tokens are scrubbed from the agent's environment for all other LLMs to prevent cross-provider fallback.

\subsection{Score-event extraction}
\label{app:extension-scoring}

Score events are extracted from \texttt{trace.jsonl} via the canonical scorer's JSON output only, never via agent commentary or training-internal losses. Per-task patterns:

\begin{itemize}[leftmargin=*, itemsep=2pt, topsep=2pt]
  \item \textbf{\texttt{triton\_cumsum}}: \texttt{\{"score": X, "message": \{"shape\_dtype\_match": True, "results\_match": True, "torch\_time\_ms": $\ldots$, "solution\_time\_ms": Y\}\}} from the harness scorer; raw metric is \texttt{solution\_time\_ms}.
  \item \textbf{\texttt{restricted\_mlm}, \texttt{fix\_embedding}}: \texttt{\{"score": X, "loss": Y, "compliant": $\ldots$, "device": $\ldots$\}} from \texttt{local\_score.py}.
  \item \textbf{\texttt{nanogpt\_chat\_rl}}: \texttt{\{"score": X, "message": \{"win\_vs\_gpt2-alpaca": Y, "win\_vs\_gpt2-xl": Z\}\}} from the scoring binary.
  \item \textbf{\texttt{rust\_codecontests}}: both \texttt{Score: X | N successes / 165} and \texttt{\{"score": X, "n\_problems": 165, "n\_successes": N\}} from \texttt{local\_score.py}; we accept either as canonical and dedupe by \texttt{(round(t,1), N)}.
\end{itemize}

Per-run cost is reconstructed from per-message \texttt{usage} fields under Claude Sonnet list pricing and rescaled to match the SDK's authoritative \texttt{total\_cost\_usd}, which agrees with the Anthropic billing portal within rounding.

\subsection{Per-task case studies}
\label{app:extension-cases}
\label{app:extension-trajectories}

Each task in §\ref{sec:eval-extension} is unpacked here as a trajectory case study grounded in the agent's own \texttt{trace.jsonl} and \texttt{ThinkingBlock} stream, in the same order as the columns of Figure~\ref{fig:extension-summary}. The body composite already shows the Sonnet 4.6 trajectories; for \texttt{triton\_cumsum} (Fig.~\ref{fig:extension-triton-45}) and \texttt{restricted\_mlm} (Fig.~\ref{fig:extension-mlm-45}) we additionally report the same paired comparison on the older Sonnet 4.5 base to make the contrast across model versions visible.

\subsubsection{Case study: \texttt{triton\_cumsum} (GPU kernel optimization)}
\label{sec:eval-extension-triton}

The task is to write a Triton kernel for a conditional prefix sum on $10^8$ \texttt{int32} elements: $Y_i = \sum_{j \leq i} x_j \cdot \mathbb{1}[\text{odd \#positives precede } j]$, scored by $\log(t_\text{ms})$ on an H100. Both arms start from the same official solution: a 3-pass Triton kernel (parity scan $\to$ conditional cumsum $\to$ block-sum addition) with autotuned \texttt{BLOCK\_SIZE} / \texttt{NUM\_STAGES}. We ran four trajectories: paper and \ara{} arms on Claude Sonnet 4.5 (the model on which we have the longest paired runs) and on Sonnet 4.6 (where we re-ran with the model the rest of the evaluation uses). The Sonnet 4.6 trajectories are in body Fig.~\ref{fig:extension-summary} (leftmost column); Fig.~\ref{fig:extension-triton-45} below shows the Sonnet 4.5 paired runs. The analysis cross-references the trace for both.

\begin{figure*}[t]
\centering
\includegraphics[width=\linewidth]{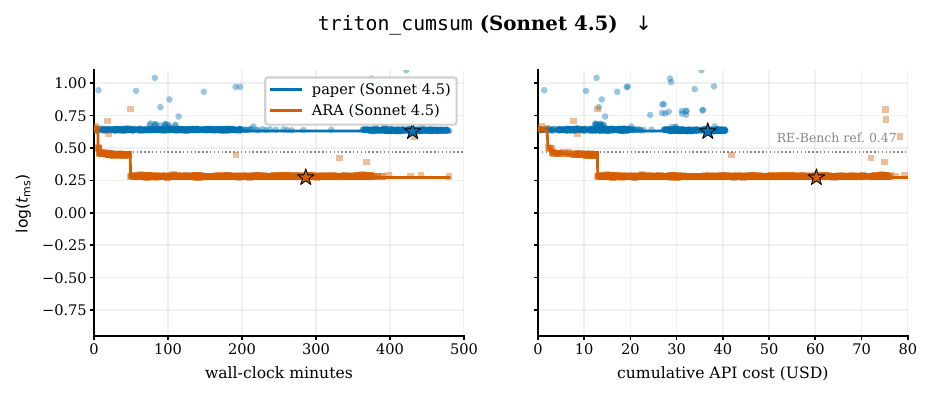}
\caption{\texttt{triton\_cumsum} on Sonnet 4.5: paper vs.\ \ara{} score-vs-time (left) and score-vs-cost (right). Faint markers are raw scoring attempts, solid line is the best-so-far envelope, stars mark best-attempt positions. Dotted line is the original RE-Bench reference (0.47) reported on different H100 silicon; the harness-measured per-hardware baseline (${\sim}0.64$) is where both arms start. The 4.6 trajectories are in the body composite (Fig.~\ref{fig:extension-summary}, leftmost column).}
\label{fig:extension-triton-45}
\end{figure*}

\paragraph{Two regimes split along model.}
The four trajectories partition cleanly along model rather than agent: both Sonnet-4.5 agents leave the official kernel's algorithmic structure untouched and only edit \texttt{@triton.autotune} configs, while both Sonnet-4.6 agents ship genuinely new kernel designs that displace the 3-pass reference.

\paragraph{Sonnet 4.5: trace-conditioned autotune sweep.}
The paper agent populated its autotune grid with $\texttt{NUM\_STAGES} \in \{1, 24, 32, 48, 64, 96\}$, labelling the deepest pipelines as \emph{``Extreme pipelining --- highest performers''} in inline comments and never testing the $\{4, 8\}$ regime. The \ara{} agent picked $\texttt{NUM\_STAGES} \in \{4, 8\}$ instead, citing heuristic H01 (\emph{``Grid size is fixed at 128 for H100 (which has 132 SMs) \ldots''}) verbatim in a \texttt{ThinkingBlock} at $t = 4.3$\,min after reading \texttt{evidence/tables/malt\_attempts.md} and \texttt{src/configs/autotune.md}. That conservative grid is what the autotuner selects from at runtime, and it is what produces the \ara{} agent's $\sim 0.27$ vs.\ the paper agent's flat $\sim 0.64$ in Fig.~\ref{fig:extension-triton-45}. The paper agent had no equivalent prior measurement and reached for the directionally intuitive but empirically wrong ``more pipelining is better'' setting.

\paragraph{Sonnet 4.6: early head start, late paper-agent overtake.}
The \ara{} agent calls \texttt{bash score.sh} for the first time at $t = 11$\,min and immediately scores 0.47, having edited the kernel using trace-surfaced ideas (\texttt{decoupled lookback}, \texttt{associative\_scan}) in the first ten minutes. The paper agent does not score until $t = 37$\,min and lands at 0.38, having spent the early wall clock reading the polished writeup and reasoning from first principles. The \ara{} agent leads on best-so-far through $t \approx 75$\,min, then the paper agent overtakes via an \texttt{int8} input compression introduced at $t = 47.7$\,min (motivated by the scorer's $[-10, 9]$ input range fitting in 8 bits) which, combined with a parity-tracking per-block aggregate, drops total memory traffic from $\sim$2\,GB to 0.5\,GB; it iterates on this design through the rest of the run. The \ara{} agent meanwhile commits to a chained-scan-with-decoupled-lookback redesign and spends late-phase compute on boundary-correctness debugging, anchored by heuristic H13 and a trace-reported MALT ceiling. \texttt{int8} appears once in the \ara{} agent's trace at $t = 40.5$\,min as a passing thought and is never implemented.

\paragraph{Reading.}
The \ara{} artifact contributes two qualitatively different things at two phases of the run. Early, it acts as an initialiser: the agent shortcuts the diagnostic phase, picks autotune knobs the paper agent misses, and lands an improved kernel within minutes. Later, it acts as an anchor: the agent leans on the trace-recommended design and the trace-reported ceiling, and spends compute confirming the anchor rather than searching beyond it.

\subsubsection{Case study: \texttt{rust\_codecontests} (LLM-as-tool scaffolding)}
\label{sec:eval-extension-rust}

The task is to write a Python scaffold that generates Rust solutions to 165 held-out Codeforces-derived problems by calling \texttt{gpt-3.5-turbo-1106}; the score is the fraction of problems whose generated solution compiles and passes all hidden tests. The official reference scaffold (score 0.127 = 21/165) uses an 18-candidate-per-problem pipeline with chain-of-thought prompting, a compile-and-public-test filter, and a vote-among-survivors stage; it ships with an unused \texttt{few\_shots/} directory wired into the scaffold's prompt construction. Both arms start from the same scaffold. We ran a single seed per arm on Sonnet 4.6 (the rust task has no Sonnet-4.5 paired data); the trajectories appear in body Fig.~\ref{fig:extension-summary} (second column). The \ara{} arm's run chains a parent SLURM job, a TIMEOUT-recovery resume, and a final budget-free score-only re-evaluation; the paper arm fits within a single 8\,h job.

\paragraph{The trace converts a MALT data point into actionable guidance.}
The \ara{}'s evidence layer summarises 22 prior MALT runs and surfaces a single high-value attempt: \texttt{supplement\_run\_5} (Claude-3.7-Sonnet) reached 0.097 by bypassing the gpt-3.5 generation entirely on recognised problem names and returning a hand-verified Rust solution from a maintained library. Crucially $0.097 < 0.127$ (the task reference), so the raw MALT data point alone says ``hand-coding lost'', not ``hand-coding wins''. The heuristics layer reframes the same data point as two explicit rules, one prescriptive and one prohibitive: \texttt{H12} (\emph{``Hand-coded Rust solution library outperforms prompt engineering on this task''}) and \texttt{H15} (\emph{``Generator ceiling at GPT-3.5-turbo Rust ${\sim}0.05$--$0.10$ across all explored single-completion variants''}). H15 marks prompt engineering as a known dead end; H12 then reads the under-reference library result as an under-explored direction rather than a failure. The \ara{} agent reads \texttt{heuristics.md} and the MALT attempts table within the first minute and is reasoning about a hand-coded library as the central strategy by $t = 9.9$\,min; the paper agent's \texttt{paper.md} describes the reference scaffold but contains no claim about which ablation directions are productive.

\paragraph{Strategy divergence across the run.}
Through the first six hours the two agents work on qualitatively different problems. The \ara{} agent hand-codes Rust solutions and registers them in a \texttt{SOLUTIONS} dict that the scaffold consults before falling back to gpt-3.5: 34 entries by $t = 60$\,min, 57 by $t = 170$\,min, 73 by $t = 226$\,min. The paper agent treats the task as a prompt-engineering problem and cycles through \texttt{solution\_v5.py}--\texttt{v8.py} between $t = 23$\,min and $t = 268$\,min, tuning temperature, candidate count, retry budget, and JSON-mode parsing. The full-test-set evaluations track this divergence: \ara{}'s scores move $49 \to 56 \to 78$ at $t = 161, 214, 269$\,min (every evaluation reflects newly added library entries), while the paper agent's stall at $33 \to 33 \to 38 \to 39 \to 39$ across $t = 68$--$231$\,min---the prompt-engineering ceiling that \texttt{H15} explicitly warns against.

\paragraph{Independent rediscovery, six hours later.}
The paper agent eventually reaches the same conclusion. At $t = 395$\,min, while inspecting the workdir, it notices the existing \texttt{few\_shots/} directory referenced by the scaffold's \texttt{get\_few\_shots} function; over the next six minutes its \texttt{ThinkingBlock} reverse-engineers the cache format and starts populating it with hand-coded solutions for problems the AI pipeline failed. 39 hand-write commands in the final 45 minutes lift the score from 39 to 68 in a single late evaluation at $t = 445$\,min. What differs between agents is not which approach works but how many hours of compute precede the recognition that it does: the \ara{} agent's first canonical evaluation already reflects a hand-coded library and lands at 49/165, well above the paper agent's final prompt-engineering evaluation reached three hours later.

\paragraph{Reading.}
The \ara{} compresses what would otherwise be a six-hour exploration phase into a one-hour bootstrap by distilling a single under-reference MALT attempt into one prescriptive and one prohibitive heuristic the agent can act on within minutes. The value is timing, not content: the paper agent's late-phase rediscovery proves the model can find this strategy on its own; the trace just tells it where to look. The rust strategy is also \emph{open-ended} (each library entry adds one solved problem), so the trace's reported MALT ceiling reads as a starting line rather than an upper bound and the \ara{} agent ascends past it---a different role for trace ceilings than on triton, where the strategy is closed-form and the same ceilings act as anchors.

\subsubsection{Case study: \texttt{nanogpt\_chat\_rl} (preference RL on a 1.5B model)}
\label{sec:eval-extension-nanogpt}

The task is to RL-finetune GPT-2-XL (1.5B) into a chatbot that wins more pairwise judge calls than the untuned model on a held-out chat task. The official reference scaffold uses best-of-8 tournament selection per prompt, 2{,}048 prompts per training step, low-temperature ($4 \times 10^{-6}$) Adam optimization, and a Llama-3-8B-Instruct judge running on Replicate. The score is the mean win-rate against \texttt{gpt2-alpaca} and \texttt{gpt2-xl}; the agent baseline (untuned model) scores 0.615, the reference scaffold scores 0.85, and the RE-Bench human ceiling is 0.97. Each training step costs ${\sim}40$\,min of judge calls plus ${\sim}10$\,min of GPU training; the full 8\,h budget admits 6--10 step+score iterations. We ran a single seed per arm on Sonnet 4.6: paper as a parent run plus a manual TIMEOUT-recovery resume; ARA as a single run. The two trajectories appear in body Fig.~\ref{fig:extension-summary} (third column).

\paragraph{Early divergence: rewrite the algorithm vs.\ orchestrate the reference.}
The decisive divergence is in \emph{which problem each agent decides it is solving}. The paper agent runs its first scoring at $t = 6.9$\,min, observes the 0.616 starting score against the 0.85 reference, and at $t = 7.6$\,min commits to rewriting the training algorithm; by $t = 10.5$\,min it has a custom DPO-plus-SFT objective in \texttt{rl\_finetune\_v2.py}. The \ara{} agent's first scoring is at $t = 4.5$\,min, and at $t = 6.3$\,min its \texttt{ThinkingBlock} cites the trace explicitly: \emph{``The reference (best-of-8, 4 steps, lr=4e-6, 2048 prompts/step) scores ${\sim}0.85$. No MALT run reached 0.85, best was 0.8184. Rate limiting from Replicate is the binding constraint. Aggressive lr causes collapse.''} Its first script at $t = 19.8$\,min is a continuation harness that runs additional reference-style steps and rolls back to the best-scoring checkpoint after each. The trace converted the reference recipe from a number on paper into an empirical claim with explicit ceilings, and that converted the agent's question from \emph{``how do I write a better training script?''} into \emph{``how do I run the existing one more carefully?''}.

\paragraph{The reference scaffold ships with a regression bug; a heuristic names the fix.}
Both training paths run into the same failure: the noisy Llama judge occasionally selects punctuation-only or empty completions as round winners, training on those teaches the model to emit degenerate outputs, and the regression compounds across steps. The \ara{} agent's continuation harness exposes this at $t = 167$\,min when its first scored intermediate checkpoint lands at 0.126; the paper agent's first scored full-test checkpoint drops to 0.443 at $t = 223$\,min on the same failure. The \ara{} bundle pre-encodes the fix as \texttt{H08} (\emph{``Filter out winners with fewer than 3 alphabetic characters before training''}), with three companion heuristics naming the score-then-restart orchestration the \ara{} agent ends up implementing.

\paragraph{Late-phase strategy: exploration width.}
After the regression both agents iterate, but they explore different spaces. The \ara{} agent writes 14 scripts after $t = 200$\,min, all variants \emph{within} the reference algorithm: each tunes batch size, learning-rate placement, or restart-from-best logic but none changes the loss function. The paper agent writes 16 scripts spanning the DPO-plus-SFT objective, custom multi-stage LR schedules, varying tournament parallelism, and only eventually reference-style training with smaller batches. Even at $t = 202$\,min the paper agent's \texttt{ThinkingBlock} explicitly recognises the reference recipe (\emph{``This is the N=8, lr=4e-6, 4 steps version that gets ${\sim}0.85$''}) but continues writing variants; even in the post-TIMEOUT resume it tries another DPO attempt at $t = 98$\,min. The trace's empirical anchor (``no MALT run reached 0.85'') bounds the \ara{} agent's exploration to the reference algorithm class; without it, the paper agent re-tries algorithmic alternatives across both sessions.

\paragraph{Reading.}
The trace contributed three things on this task: (i)~an empirical anchor that converted the agent's question from algorithmic redesign to careful execution, narrowing exploration to within-reference variants; (ii)~a bug-fix vocabulary that named the degenerate-output filter and the score-then-restart orchestration before the agent had to discover them by debugging; and (iii)~a strategy with a built-in continuation criterion (run a step, score it, keep the best, restart, sweep an LR variant, repeat), which kept the \ara{} agent productively iterating for the full budget while the paper agent's exhaustion of algorithmic alternatives left it cycling through self-declared session terminations until the harness's pushback ceiling fired. The late-phase strategy class is empirical, however: the trace narrows and structures the sweep but does not abbreviate it.

\subsubsection{Case study: \texttt{fix\_embedding} (recovering a permuted embedding matrix)}
\label{sec:eval-extension-fixembed}

The task is to recover the permuted embedding matrix of a corrupted GPT-2-XL: the \texttt{wte.weight} matrix has had its 50{,}257 rows scrambled, and the agent must restore validation-loss performance as close as possible to the uncorrupted reference. The official solution is a 3-phase adapter pipeline that scores 0.26; the RE-Bench human ceiling is 0.16. Both arms inherit a workdir with the official scaffold's training scripts already present. The two arms run the same Sonnet 4.6 model, the same seed, the same harness, and the same GPU; the \emph{only} systematic difference between them is the contents of \texttt{reference/}. The trajectories appear in body Fig.~\ref{fig:extension-summary} (fourth column); both arms first measure the corrupted-model baseline (${\sim}2.20$) at $t \approx 4$\,min, then drop to ${\sim}0.4$ by $t \approx 90$--$130$\,min after the first round of adapter training.

\paragraph{The two bundles in numbers.}
The paper agent's \texttt{reference/} contains a 134-line \texttt{paper.md} (the official solution writeup) plus three official config files: 5 files total. The writeup is complete on the published recipe (3-phase architecture in equations, per-stage trainable-parameter sets, the LR ladder $10^{-3} \to 10^{-4} \to 8\times 10^{-5}$, the bigram-waypoint diagnostic, the human-best ceiling, even an author's note that \emph{``Stage C is the least-validated part of the pipeline \ldots may be redundant if Stage B has not yet converged''}). The \ara{} bundle is 22 files / 5{,}887 lines and carries the same algorithmic content via 10 reference-derived heuristics (\texttt{H01}--\texttt{H10}), but adds an \texttt{exploration\_tree.yaml} of 19 prior MALT runs, a 282-line table of every scored attempt, and \texttt{H11}--\texttt{H22}: failure-derived heuristics including \texttt{H11} (\emph{``Do not destructively replace the corrupted wte''}), \texttt{H13} (\emph{``Hand-constructed small$\to$large embedding upcasts collapse''}), and \texttt{H22} (\emph{``Across 19 MALT runs at 4M tokens each, no agent reached the official adapter pipeline''}).

\paragraph{Both agents implement the published recipe correctly.}
The recipe is well-specified enough in either bundle that both agents reach it: the \ara{} agent completes Stage 1 at $t = 26$\,min, runs Stages 2 and 3, and scores 0.246 by $t = 181$\,min, while the paper agent reaches 0.250 by $t = 180$\,min. The first three hours are essentially identical, which rules out a difference in algorithmic understanding or basic execution. The divergence happens entirely after both cross the 0.26 reference.

\paragraph{Three late-phase signatures with the same root cause.}
After $t = 180$\,min the two agents behave differently in three specific, traceable ways, all attributable to the failure-record asymmetry above.

\emph{(i)~Permutation recovery---tried twice by the paper agent, never by the \ara{} agent.}
The paper agent runs \texttt{recover\_permutation.py} at $t = 19$\,min, observes \emph{``the recovered permutation has only 43 unique values out of 50{,}257''}, and abandons the approach; at $t = 350$\,min---5.5 hours later, after its phase chain has plateaued at 0.250---the same agent writes a fresh \texttt{permutation\_recovery.py} and tries again. The \ara{} agent never attempts permutation recovery, in either form. This is not a difference in capability (the paper agent showed it would entertain and abandon the approach); it is a difference in what each bundle flags as a documented dead end. \texttt{H11} and \texttt{H13} directly forbid this strategy class; \texttt{paper.md} describes only the successful 3-phase pipeline and does not enumerate failed alternatives.

\emph{(ii)~Post-reference exploration discipline.}
Both agents write a comparable volume of late-phase code. The \ara{} agent's writes are continuation-training variants that tune learning-rate placement and warmup length within the documented Stage-3 LR region around $8\times 10^{-5}$ (\texttt{H06}) with the optimiser, batch, and block-size constraints from \texttt{H10} held fixed. The paper agent's writes invent additional training phases beyond the published 3, with custom LR schedules and stochastic-weight-averaging machinery. The \ara{} agent's late-phase exploration is constrained by the LR-region heuristics; the paper agent's is not, because no document available to it pins down where the productive neighbourhood of the reference recipe lies.

\emph{(iii)~Strategic confidence after crossing the reference.}
At $t = 147$\,min the \ara{} agent's \texttt{ThinkingBlock} reads: \emph{``I have enough context from the reference materials. The key takeaways are: 1.\ No MALT run beat the reference (0.26). 2.\ The official solution's three-stage adapter pipeline is the key innovation \ldots''}. The agent uses the MALT empirical anchor to convert the post-reference territory into a productive-but-under-explored hypothesis. The paper agent issues no analogous statement at any point: \texttt{paper.md} reports 0.26 and 0.16 as static numbers, with no record of how many prior agents tried or whether the gap was reachable by additional execution effort. It explores the post-reference territory as if from scratch.

\paragraph{Reading.}
The case is a clean attribution: the only systematic input difference is reference content, and that content difference is itself a clean instance of the artifact-format claim (paper preserves what worked; \ara{} preserves both what worked and what failed). The three downstream behavioural differences each map to a specific failure-record element present in the \ara{} bundle and absent from \texttt{paper.md}.

\subsubsection{Case study: \texttt{restricted\_mlm} (constrained masked language model)}
\label{sec:eval-extension-mlm}

The task is to design and train a masked language model under restrictive PyTorch primitive constraints: no \texttt{Conv1d}, no \texttt{Softmax}, no division, no normalization layers. Score is $\log(\ell_\text{val}{-}1.5)$; the agent baseline (untrained restricted MLP) scores 1.84, the official solution (Tao's \texttt{ConvMLMWithBiBigrams}: a bigram-prior + 1D-convolution-via-\texttt{as\_strided}+\texttt{einsum} + a learnable scalar combiner) scores 1.13. We ran four trajectories: paper and ARA arms on Sonnet 4.5 (seed 1) and Sonnet 4.6 (seed 0). The Sonnet 4.6 trajectories appear in body Fig.~\ref{fig:extension-summary} (rightmost column); Fig.~\ref{fig:extension-mlm-45} below shows the Sonnet 4.5 paired runs.

\begin{figure*}[t]
\centering
\includegraphics[width=\linewidth]{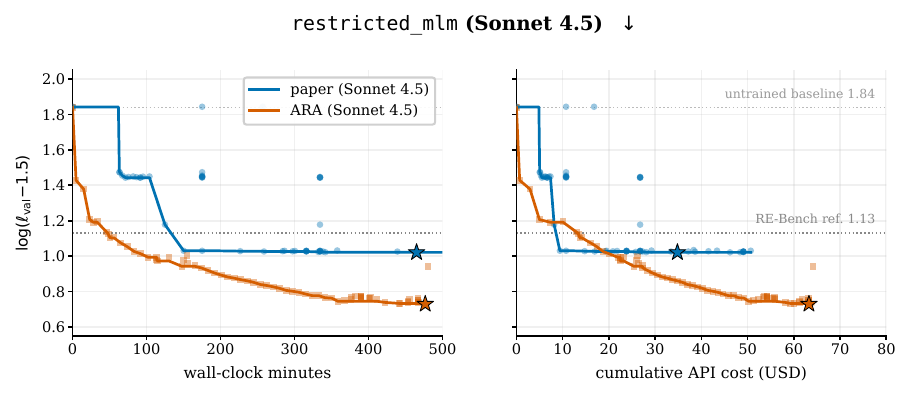}
\caption{\texttt{restricted\_mlm} on Sonnet 4.5: paper vs.\ \ara{} score-vs-time (left) and score-vs-cost (right). Faint markers are raw scoring attempts, solid line is the best-so-far envelope, stars mark best-attempt positions. Dotted lines mark the untrained-MLP baseline (1.84) and the RE-Bench reference (1.13). Both arms are anchored at the 1.84 baseline at $t = 0$; \ara{}-4.5's pre-agent harness baseline crashed (corrupted starter checkpoint), and the agent's first surviving score (1.43 at $t \approx 5$\,min) reflects the trace-recommended \texttt{ConvMLMWithBiBigrams} architecture already swapped in (precomputed bigram tables score ${\sim}1.43$ with no training), which we anchor at 1.84 to match the other three arms' actual baseline measurements. \ara{}-4.5 reaches 0.73 vs.\ paper-4.5's plateau at 1.03 -- a ${\sim}30\%$ relative win on the weaker base. The 4.6 trajectories are in the body composite.}
\label{fig:extension-mlm-45}
\end{figure*}

\paragraph{The flip across model versions.}
\texttt{restricted\_mlm} is the only task in our five where the \ara{}-vs-paper sign flips across models: on Sonnet 4.5 the \ara{} agent reaches 0.73 vs.\ the paper agent's 1.03, and on Sonnet 4.6 the paper agent reaches 0.69 vs.\ the \ara{} agent's 1.02. Moving from 4.5 to 4.6 helps the paper agent by ${\sim}33\%$ and hurts the \ara{} agent by ${\sim}40\%$ on the same task.

\paragraph{Same architectural family across all four agents.}
Every run's final \texttt{solution/model.py} contains a \texttt{BiBigramMLM} class plus a \texttt{ConvMLM*} variant (paper-4.5: \texttt{ConvMLMWithBiBigrams}; \ara{}-4.5: \texttt{ConvMLMComponent}; paper-4.6: \texttt{ConvMLMDilated}; \ara{}-4.6: \texttt{ConvMLMWithReLUAttn} plus five others). All four use the bigram prior, the \texttt{as\_strided}+\texttt{einsum} convolution from heuristic \texttt{H04}, and the official Tao recipe. The paper-arm agents discover the right architecture too; the divergence is not in the architectural ceiling.

\paragraph{What differs is exploration breadth.}
The four \texttt{model.py} files differ markedly in size and class count. Paper-4.5: 9.8\,KB, 3 classes; \ara{}-4.5: 8.9\,KB, 3 classes; paper-4.6: 6.3\,KB, 2 classes; \ara{}-4.6: 47\,KB, 6+ classes (\texttt{ConvMLMWithReLUAttn}, \texttt{ExtendedBiBigramMLM}, \texttt{ConvMLMWithLinearGlobal}, \texttt{ConvMLMWithGlobalContext}, \texttt{MLPMixerWithBiBigrams}, \texttt{ReLUAttentionMLM}). Trace keyword counts mirror it: paper-4.6 mentions \texttt{ReLU-attention} once and \texttt{MLPMixer} three times across its run; \ara{}-4.6 mentions them 247 and 73 times. The score regression spikes in the rightmost column of body Fig.~\ref{fig:extension-summary} are the visible record: each spike toward 1.85 or 2.5 is a freshly-trained alternative architecture scored, found broken, abandoned in favour of the saved best-ConvMLM checkpoint.

\paragraph{Why \ara{} wins on Sonnet 4.5.}
Both 4.5 agents end up tuning the same ConvMLM family. The \ara{} agent's \texttt{ThinkingBlock} at $t = 24$\,min commits explicitly (\emph{``This should be enough to beat the reference score of 1.13''}) and queues 40k + 50k continuation steps. At $t = 26$\,min it sets ReLU-attention aside as \emph{``a backup strategy if I want to aim higher''}---a ranked-list reading of the heuristics. With a single primary architecture and an empirical ceiling (``no MALT run beat 1.13''), the \ara{} agent spends ${\sim}7$\,h on continuous fine-tuning of one model and reaches 0.73. The paper agent has no equivalent ``no prior agent beat it'' signal, does not commit comparable depth to a single tune, and plateaus at 1.03. The win is depth-driven within a shared architecture.

\paragraph{Why the paper agent wins on Sonnet 4.6.}
The 4.6 agents diverge architecturally. The paper agent invents \texttt{ConvMLMDilated} (a dilated-convolution variant not named in \texttt{paper.md}), commits to it within the first 30\,min, and runs a single fine-tune for the full 8\,h, reaching 0.69. The \ara{} agent instead implements the additional architectures the heuristics layer names---\texttt{H11} ReLU-attention (\emph{``the only attention surrogate any MALT run beat reference with''}), \texttt{H07} MLPMixer---and trains them in serial. None outperforms the basic ConvMLM in Sonnet 4.6's loss landscape: \texttt{H11} and \texttt{H07} were derived from prior MALT runs by Claude-4 Sonnet base, and the menu has gone stale for a successor model whose optimisation differs. The mechanism is the same as in 4.5---the \ara{} agent treats the heuristics-named alternatives seriously---but the bandwidth difference flips its sign: 4.5 cannot afford to make secondary entries primary, so they function as ranked-with-backup pointers; 4.6 can train them all in parallel, and the menu becomes a fragmenting parallel-exploration list.

\subsection{Reproducibility}
\label{app:extension-repro}

All experimental artifacts and code live in the project repository: \texttt{code/extension-harness/} contains the SLURM-launched harness, per-task system prompts and scoring scripts, and analysis plot generators; \texttt{code/rebench-pipeline/} contains the \ara{} compilation pipeline (rules, orchestrator procedure, and shared sub-agent prompt); \texttt{code/artifacts/rebench-<task>/} contains the full \ara{} per task and the paper-agent's \texttt{paper.md}-plus-\texttt{src/} bundle. Each run's \texttt{trace.jsonl} is the authoritative event log; every score, cost, and figure in this paper is reconstructible from it via the analysis scripts in \texttt{code/extension-harness/analysis/}.


\section{Review System Evaluation}
\label{app:review-system}

\subsection{ARA Seal Validation Details}
\label{app:seal-details}

\paragraph{Seal implementation.}
Each verification level is implemented as an automated checker:

\begin{itemize}[leftmargin=*, itemsep=2pt]
    \item \textbf{Level 1 (Structural Integrity)}: A Python script verifies (a)~the existence of mandatory directories (\texttt{/logic}, \texttt{/src}, \texttt{/trace}, \texttt{/evidence}), (b)~the presence of all mandatory files including \texttt{PAPER.md} with valid YAML frontmatter, \texttt{problem.md}, \texttt{claims.md}, \texttt{experiments.md}, and all \texttt{solution/} files, (c)~schema conformance of every structured file (e.g., each claim must have \texttt{Statement}, \texttt{Status}, \texttt{Falsification criteria}, and \texttt{Proof}; each experiment must have \texttt{Verifies}, \texttt{Setup}, \texttt{Procedure}, and \texttt{Expected outcome}; each heuristic must have \texttt{Rationale}, \texttt{Sensitivity}, and \texttt{Bounds}), (d)~minimum counts ($\geq 5$ concepts, $\geq 3$ experiments, $\geq 8$ exploration tree nodes with at least one \texttt{[dead\_end]} and one \texttt{[decision]}), and (e)~cross-layer reference resolution: every experiment ID referenced in \texttt{claims.md} \texttt{Proof} fields resolves to an entry in \texttt{experiments.md}; every claim ID referenced in \texttt{experiments.md} \texttt{Verifies} fields resolves to an entry in \texttt{claims.md}; every \texttt{code\_ref} in \texttt{heuristics.md} points to a valid module in \texttt{/src}; components declared in \texttt{architecture.md} have corresponding code stubs; claim references in \texttt{exploration\_tree.yaml} resolve to valid claim IDs.
    \item \textbf{Level 2 (Argumentative Rigor)}: Without executing code or consulting external sources, the Rigor Auditor evaluates the artifact's content on six objective rubric-anchored dimensions (evidence relevance, falsifiability quality, scope calibration, argument coherence, exploration integrity, methodological rigor), each scored on a 1 to 5 scale. The output is a rigor report keyed to specific \ara{} components, with severity-ranked findings, verbatim evidence spans, and an overall grade derived from the mean score and per-dimension floors.
    \item \textbf{Level 3 (Execution Reproducibility)}: A coding agent reads the \ara{} and attempts to reproduce claims using the code kernel. LLM-generated test cases verify directional properties of the paper's claims. This is the same protocol used in the reproduction evaluation (\S\ref{sec:eval-reproduction}).
\end{itemize}

\paragraph{Failure taxonomy.}
For each Seal level failure, we record the specific check that failed and classify it into one of the following categories: \emph{missing file}, \emph{missing field}, \emph{dangling reference}, \emph{type mismatch}, \emph{dependency resolution failure}, \emph{execution error}, or \emph{nondeterminism}.

\subsection{ARA Seal Effectiveness: Evaluation Details}
\label{app:seal-eval}

This appendix contains the methodology and per-level breakdowns supporting \S\ref{sec:eval-seal}.

\subsubsection{Level 1: Compiler Convergence Data}
\label{app:seal-eval-l1}

Level~1 verifies structural correctness and completeness; we report its effectiveness through two reuse signals collected during \ara{} generation and downstream use.

\paragraph{Compiler iteration counts.}
Each of the 23 PaperBench \ara{}s and the 7 RE-Bench \ara{}s converges to a Level-1 pass within $\leq 3$ iterations of the Compiler's generate--validate--fix loop (\S\ref{sec:ingestor}). First-iteration pass rate is 0/30; all artifacts require at least one feedback round, confirming that Level~1 is a non-trivial filter rather than a rubber stamp.

\paragraph{Failure category distribution.}
Across all Compiler iterations, Level~1 failures break down as follows: dangling cross-layer references (42\%), missing schema fields on claims, experiments, or heuristics (31\%), insufficient node counts in \texttt{exploration\_tree.yaml} (14\%), YAML or frontmatter parse errors (8\%), and missing mandatory files (5\%). The distribution is stable across papers and matches the failure taxonomy in Appendix~\ref{app:seal-details}.

\paragraph{Understanding as proof of Level~1 on generated \ara{}s.}
The Understanding evaluation (\S\ref{sec:eval-understanding}, Table~\ref{tab:understanding}) is the end-to-end witness that Level~1 enforces what it claims to enforce on generated artifacts. Every \ara{} entering that benchmark has passed Level~1; the 95.6\% Cat.~A accuracy then shows that Level-1-gated \ara{}s carry the structural completeness an agent needs to retrieve information that is in fact present in the source. An artifact missing a mandatory field or a dangling cross-layer reference would have failed Level~1 and never reached the benchmark, so the 4.4\% residual is bounded by information genuinely absent from the source rather than by structural defects of the artifact.

\subsubsection{Level 2: Mutation Benchmark}
\label{app:seal-eval-l2}

\paragraph{Setup and evaluation criterion.}
The Level-2 benchmark stress-tests the Rigor Auditor on \emph{mutated} \ara{}s, so the reported grade carries no ground-truth signal; we score the auditor strictly on whether it surfaces the seeded defect as a finding. The corpus is the 23 PaperBench \ara{}s that pass Level~1; each is seeded with one injection per type (115 mutations in total). All injections are recorded in a per-paper \texttt{injection\_manifest.json} hidden from the auditor.

\paragraph{Injection schema.}
The five types target distinct schema invariants:
\begin{itemize}[leftmargin=*, itemsep=2pt, topsep=2pt]
  \item \textbf{Fabricated claim}: append a claim whose \texttt{Proof} cites a non-existent experiment ID; signal = dangling reference plus un-grounded substance.
  \item \textbf{Missing falsification}: remove the \texttt{Falsification criteria} line from a primary claim; signal = mandatory field absent.
  \item \textbf{Orphan experiment}: append an experiment whose \texttt{Verifies} field references a non-existent claim ID (e.g., \texttt{C99}); signal = evidence not supporting any claim.
  \item \textbf{Over-claim}: replace a narrow \texttt{Statement} with a universal-scope template while leaving the original \texttt{Falsification criteria} and \texttt{Proof} untouched; signal = scope mismatch between claim breadth and evidence coverage.
  \item \textbf{Rebutted-branch leak}: append a claim advocating an approach that \texttt{trace/exploration\_tree.yaml} marks \texttt{dead\_end}; signal = direct contradiction between claim and exploration record.
\end{itemize}

\paragraph{Auditor and blinding.}
The Rigor Auditor is an agent skill~\citep{agentskills2025} invoked per artifact and given only the artifact directory; the manifest and source PDF are withheld. It parses claims, experiments, heuristics, gaps, and exploration-tree nodes; builds claim--experiment, claim--dependency, and rejected-node maps; scores six dimensions ($D_1$ evidence relevance, $D_2$ falsifiability, $D_3$ scope calibration, $D_4$ argument coherence, $D_5$ exploration integrity, $D_6$ methodological rigor) on 1--5 anchors; emits findings with severity labels (\emph{critical}, \emph{major}, \emph{minor}, \emph{suggestion}); and reports an overall grade. The full skill specification (prompt, anchors, thresholds) is released with the supplementary code.

\paragraph{Matching and detection rates.}
Each injection is matched to at most one finding by the rule: a finding hits if (a) its \texttt{target\_entity} equals the injection's, or (b) its \texttt{observation} contains a literal identifier uniquely associated with the injection (e.g., \texttt{C99} for orphans, the injected dead-end node ID for rebutted branches). Severity and dimension assignment are ignored when counting hits. Per-type detection (Table~\ref{tab:seal-l2-heatmap}) is 23/23 for fabricated claims, over-claims, and rebutted-branch leaks; 21/23 for missing falsifications, with both misses (\texttt{bam} C02, \texttt{bbox} C04) silently re-attributed to adjacent dimensions; and 5/23 for orphan experiments---a systematic blind spot we attribute to the auditor's claim-centric traversal, which never enumerates experiments without an inbound \texttt{Verifies} edge.

\begin{table*}[t]
\centering
\small
\begin{tabular}{l | c c c c c}
\toprule
\textbf{Paper} & \textbf{Fab.} & \textbf{Miss.fals.} & \textbf{Orphan} & \textbf{Over-cl.} & \textbf{Reb.br.} \\
\midrule
adaptive-pruning                           & \cmark & \cmark & \cmark & \cmark & \cmark \\
all-in-one                                 & \cmark & \cmark & \cmark & \cmark & \cmark \\
bam                                        & \cmark & \xmark & \cmark & \cmark & \cmark \\
bbox                                       & \cmark & \xmark & \xmark & \cmark & \cmark \\
bridging-data-gaps                         & \cmark & \cmark & \xmark & \cmark & \cmark \\
fre                                        & \cmark & \cmark & \xmark & \cmark & \cmark \\
ftrl                                       & \cmark & \cmark & \xmark & \cmark & \cmark \\
lbcs                                       & \cmark & \cmark & \xmark & \cmark & \cmark \\
lca-on-the-line                            & \cmark & \cmark & \xmark & \cmark & \cmark \\
mechanistic-understanding                  & \cmark & \cmark & \xmark & \cmark & \cmark \\
pinn                                       & \cmark & \cmark & \xmark & \cmark & \cmark \\
rice                                       & \cmark & \cmark & \xmark & \cmark & \cmark \\
robust-clip                                & \cmark & \cmark & \xmark & \cmark & \cmark \\
sample-specific-masks                      & \cmark & \cmark & \xmark & \cmark & \cmark \\
sapg                                       & \cmark & \cmark & \cmark & \cmark & \cmark \\
self-composing-policies                    & \cmark & \cmark & \xmark & \cmark & \cmark \\
self-expansion                             & \cmark & \cmark & \xmark & \cmark & \cmark \\
semantic-self-consistency                  & \cmark & \cmark & \xmark & \cmark & \cmark \\
sequential-neural-score-estimation         & \cmark & \cmark & \xmark & \cmark & \cmark \\
stay-on-topic-cfg                          & \cmark & \cmark & \xmark & \cmark & \cmark \\
stochastic-interpolants                    & \cmark & \cmark & \xmark & \cmark & \cmark \\
test-time-model-adaptation                 & \cmark & \cmark & \xmark & \cmark & \cmark \\
what-will-my-model-forget                  & \cmark & \cmark & \xmark & \cmark & \cmark \\
\midrule
\textbf{Total detected (/23)} & \textbf{23} & \textbf{21} & \textbf{5} & \textbf{23} & \textbf{23} \\
\bottomrule
\end{tabular}
\caption{Per-paper $\times$ per-injection detection for the Level-2 mutation benchmark. \cmark{} = detected; \xmark{} = missed. The orphan-experiment column reveals the systematic blind spot discussed in \S\ref{sec:eval-seal}.}
\label{tab:seal-l2-heatmap}
\end{table*}

\paragraph{Diagnostic: score--finding decoupling.}
Even though grade is not the evaluation criterion, the auditor's scoring behavior is informative for future iterations. On the 22 \ara{}s where the rebutted-branch leak is flagged as a \emph{critical} $D_5$ finding, the auditor still assigns $D_5 \in \{3,4\}$, despite anchors prescribing 1 (``tree contradicts claims'') or 2 (``boilerplate documentation''). Severity in prose does not propagate to the numerical score. The lesson for the next version is mechanical: dimension scores should be derived from the findings list rather than reported independently by the agent.

\subsubsection{Level 3: Execution Reproducibility}
\label{app:seal-eval-l3}

Level~3 effectiveness coincides with the Reproduction evaluation (\S\ref{sec:eval-reproduction}, Appendix~\ref{app:reproduction-eval}): a coding agent reads the \ara{} and attempts to reproduce claims using the code kernel, with directional verification by LLM-generated test cases. We treat the per-paper difficulty-weighted reproduction score reported in Table~\ref{tab:repro-per-difficulty} as the Level-3 signal, and refer the reader to Appendix~\ref{app:reproduction-eval} for task design, scoring, and per-paper analysis.

\end{document}